\documentclass[11pt]{article}

\usepackage{jmlr2e}

\usepackage[labelfont=bf]{caption}
\usepackage[noend]{algpseudocode}
\usepackage[nottoc]{tocbibind}
\usepackage[subrefformat=parens]{subcaption}
\usepackage[T1]{fontenc}
\usepackage[usenames,dvipsnames,svgnames]{xcolor}

\usepackage{algorithmicx}
\usepackage{algorithm}
\usepackage{amsfonts}
\usepackage{amsmath}
\usepackage{amssymb}
\usepackage{bm}
\usepackage{booktabs}
\usepackage{chngcntr}
\usepackage{enumitem}
\usepackage{fancyvrb}
\usepackage{graphicx}
\usepackage{mdframed}
\usepackage{multicol}
\usepackage{txfonts}
\usepackage{placeins}
\usepackage{tikz}
\usepackage{todonotes}
\usepackage{url}

\DeclareCaptionFormat{algor}{%
  \hrulefill\par\offinterlineskip\vskip1pt%
  \textbf{#1#2}#3\offinterlineskip\hrulefill}
\DeclareCaptionStyle{algori}{singlelinecheck=off,format=algor,labelsep=space}
\newcounter{parentalgorithm}

\makeatletter
\newenvironment{subalgorithms}{%
  \refstepcounter{algorithm}%
  \protected@edef\theparentalgorithm{\thealgorithm}%
  \setcounter{parentalgorithm}{\value{algorithm}}%
  \setcounter{algorithm}{0}%
  \def\thealgorithm{\theparentalgorithm\alph{algorithm}}%
  \ignorespaces
}{%
  \setcounter{algorithm}{\value{parentalgorithm}}%
  \ignorespacesafterend
}
\makeatother

\usetikzlibrary{bayesnet}
\usetikzlibrary{arrows}

\newcommand*\Let[2]{\State {#1} $\gets$ {#2}}
\newcommand*\Sample[2]{\State {#1} $\sim$ {#2}}

\newcommand{\balpha}{\bm\alpha}

\newcommand{\bgamma}{\bm\gamma}
\newcommand{\btheta}{\bm\theta}
\newcommand{\D}{\mathcal{D}}
\newcommand{\G}{\mathcal{G}}
\newcommand{\M}{\mathcal{M}}
\newcommand{\R}{\mathbb{R}}
\newcommand{\s}{\bm{s}}

\newcommand{\w}{\bm{w}}
\newcommand{\x}{\bm{x}}
\newcommand{\X}{\bm{X}}
\newcommand{\y}{\bm{y}}
\newcommand{\Y}{\bm{Y}}
\newcommand{\z}{\bm{z}}
\newcommand{\Z}{\bm{Z}}

\newcommand{\indep}{\perp\!\!\!\perp}

\newcommand{\set}[1]{\{{#1}\}}

\newcommand{\pG}{p_{\G}}

\newcommand{\bcaption}[2]{\caption{\textbf{#1} {#2}}}

\newcommand{\red}[1]{\textcolor{BrickRed}{#1}}
\newcommand{\blue}[1]{\textcolor{MidnightBlue}{#1}}
\newcommand{\oliveg}[1]{\textcolor{OliveGreen}{\textbf{#1}}}

\title{Probabilistic Data Analysis with Probabilistic Programming}

\author{%
    \name Feras Saad \email fsaad@mit.edu \\
    \addr Computer Science \& Artificial Intelligence Laboratory\\
    Massachusetts Institute of Technology\\
    Cambridge, MA 02139, USA
    \AND
    \name Vikash Mansinghka \email vkm@mit.edu \\
    \addr Department of Brain \& Cognitive Sciences\\
    Massachusetts Institute of Technology\\
    Cambridge, MA 02139, USA
}

\begin{document}

\maketitle


\begin{abstract}%
Probabilistic techniques are central to data analysis, but different approaches can be difficult to apply, combine, and compare.
This paper introduces composable generative population models (CGPMs), a computational abstraction that extends directed graphical models and can be used to describe and compose a broad class of probabilistic data analysis techniques.
Examples include hierarchical Bayesian models, multivariate kernel methods, discriminative machine learning, clustering algorithms, dimensionality reduction, and arbitrary probabilistic programs.
We also demonstrate the integration of CGPMs into BayesDB, a probabilistic programming platform that can express data analysis tasks using a modeling language and a structured query language. The practical value is illustrated in two ways.
First, CGPMs are used in an analysis that identifies satellite data records which probably violate Kepler's Third Law, by composing causal probabilistic programs with non-parametric Bayes in under 50 lines of probabilistic code.
Second, for several representative data analysis tasks, we report on lines of code and accuracy measurements of various CGPMs, plus comparisons with standard baseline solutions from Python and MATLAB libraries.
\end{abstract}




\acks{%
The authors thank Taylor Campbell, Gregory Marton, and Alexey Radul for engineering support, and Anthony Lu, Richard Tibbetts, and Marco Cusumano-Towner for helpful contributions, feedback and discussions.
This research was supported by DARPA (PPAML program, contract number \textrm{FA8750-14-2-0004}), IARPA (under research contract \textrm{2015-15061000003}), the Office of Naval Research (under research contract N000141310333), the Army Research Office (under agreement number \textrm{W911NF-13-1-0212}), and gifts from Analog Devices and Google.}


\section{Introduction}
\label{sec:introduction}

Probabilistic techniques are central to data analysis, but can be difficult to apply, combine, and compare.
Families of approaches such as parametric statistical modeling, machine learning and probabilistic programming are each associated with different formalisms and assumptions.
This paper shows how to address these challenges by defining a new family of probabilistic models and integrating them into BayesDB, a probabilistic programming platform for data analysis.
It also gives empirical illustrations of the efficacy of the framework on multiple synthetic and real-world tasks in probabilistic data analysis.

This paper introduces composable generative population models (CGPMs), a computational formalism that extends graphical models for use with probabilistic programming.
CGPMs specify a table of observable random variables with a finite number of columns and a countably infinite number of rows.
They support complex intra-row dependencies among the observables, as well as inter-row dependencies among a field of latent variables.
CGPMs are described by a computational interface for generating samples and evaluating densities for random variables, including the (random) entries in the table as well as a broad class of random variables derived from these via conditioning.
We show how to implement CGPMs for several model families such as the outputs of standard discriminative learning methods, kernel density estimators, nearest neighbors, non-parametric Bayesian methods, and arbitrary probabilistic programs.
We also describe algorithms and new syntaxes in the probabilistic Metamodeling Language for building compositions of CGPMs that can interoperate with BayesDB.

The practical value is illustrated in two ways.
First, the paper outlines a collection of data analysis tasks with CGPMs on a high-dimensional, real-world dataset with heterogeneous types and sparse observations.
The BayesDB script builds models which combine non-parametric Bayes, principal component analysis, random forest classification, ordinary least squares, and a causal probabilistic program that implements a stochastic variant of Kepler's Third Law.
Second, we illustrate coverage and conciseness of the CGPM abstraction by quantifying the lines of code and accuracy achieved on several representative data analysis tasks.
Estimates are given for models expressed as CGPMs in BayesDB, as well as for baseline methods implemented in Python and MATLAB.
Savings in lines of code of \textasciitilde10x at no cost or improvement in accuracy are typical.

The remainder of the paper is structured as follows.
Section~\ref{sec:related-work} reviews related work to CGPMs in both graphical statistics and probabilistic programming.
Section~\ref{sec:cgpms} describes the conceptual, computational, and statistical formalism for CGPMs.
Section~\ref{sec:implementation} formulates a wide range of probabilistic models as CGPMs, and provides both algorithmic implementations of the interface as well as examples of their invocations through the Metamodeling Language and Bayesian Query Language.
Section~\ref{sec:bayesdb} outlines an architecture of BayesDB for use with CGPMs.
We show how CGPMs can be composed to form a generalized directed acyclic graph, constructing hybrid models from simpler primitives.
We also present new syntaxes in MML and BQL for building and querying CGPMs in BayesDB.
Section~\ref{sec:applications} applies CGPMs to several probabilistic data analysis tasks in a complex real-world dataset, and reports on lines of code and accuracy measurements.
Section~\ref{sec:discussion} concludes with a discussion and directions for future work.


\section{Related Work}
\label{sec:related-work}

Directed graphical models from statistics provide a compact, general-purpose modeling language to describe both the factorization structure and conditional distributions of a high-dimensional joint distribution \citep{koller2007}.
Each node is a random variable which is conditionally independent of its non-descendants given its parents, and its conditional distribution given all its parents is specified by a conditional probability table or density \citep[Sec 2.3]{nielsen2009}.
CGPMs extend this mathematical description to a computational one, where nodes are not only random variables with conditional densities but also computational units (CGPMs) with an interface that allows them to be composed directly as software.
A CGPM node typically encapsulates a more complex statistical object than a single variable in a graphical model.
Each node has a set of required input variables and output variables, and all variables are associated with statistical data types.
Nodes are required to both simulate and evaluate the density of a subset of their outputs by conditioning on all their inputs, as well as either conditioning or marginalizing over another subset of their outputs.
Internally, the joint distribution of output variables for a single CGPM node can itself be specified by a general model which is either directed or undirected.

CGPMs combine ideas from the vast literature on modeling and inference in graphical models with ideas from probabilistic programming.
This paper illustrates CGPMs by integrating them into BayesDB \citep{mansinghka2015}, a probabilistic programming platform for data analysis.
BayesDB demonstrated that the Bayesian Query Language (BQL) can express several tasks from multivariate statistics and probabilistic machine learning in a model-independent way.
However this idea was illustrated by emphasizing that a domain-general baseline model builder based on CrossCat \citep{mansinghka2015-2}, with limited support for plug-in models called ``foreign predictors'', provides good enough performance for common statistical tasks.
Due to limitations in the underlying formalism of generative population models (GPMs), which do not accept inputs and only learn joint distributions over observable variables, the paper did not provide an expressive modeling language for constructing a wide class of models applicable to different data analysis tasks, or for integrating domain-specific models built by experts into BayesDB.
By both accepting input variables and exposing latent variables as queryable outputs, CGPMs provide a concrete proposal for mediating between automated and custom modeling using the Metamodeling Language, and model-independent querying using the Bayesian Query Language.
The CGPM abstraction thus exposes the generality of BQL to a much broader model class than originally presented, which includes hybrids models with generative and discriminative components.

It is helpful to contrast CGPMs in BayesDB with other probabilistic programming formalisms such as Stan \citep{carpenter2015}.
Stan is a probabilistic programming language for specifying hierarchical Bayesian models, with built-in algorithms for automated, highly efficient posterior inference.
However, it is not straightforward to (i) integrate models from different formalisms such as discriminative machine learning as sub-parts of the overall model, (ii) directly query the outputs of the model for downstream data analysis tasks, which needs to be done on a per-program basis, and (iii) build composite programs out of smaller Stan programs, since each program is an independent unit without an interface.
CGPMs provide an interface for addressing these limitations and makes it possible to wrap Stan programs as CGPMs that can then interact, through BayesDB, with CGPMs implemented in other systems.

Tabular \citep{gordon2014} is a schema-driven probabilistic programming language which shares some similarity to composable generative population models.
For instance, both the statistical representation of a CGPM (Section~\ref{subsec:cgpms-statistical}), and a probabilistic schema in Tabular, characterize a data generating process in terms of input variables, output variables, latent variables, parameters and hyper-parameters.
However, unlike Tabular schemas, CGPMs explicitly provide a computational interface, which is more general than the description of their internal structure, and facilitates their composition (Section~\ref{subsec:composition}).
In Tabular, probabilistic programs are centered around parametric statistical modeling in factor graphs, where the user manually constructs variable nodes, factor nodes, and the quantitative relationships between them.
On the other hand, CGPMs express a broad range of model classes which do not necessarily naturally admit natural representations as factor graphs, and combine higher-level automatic model discovery (using baseline generative CGPMs) with user-specified overrides for hybrid modeling.


\section{Composable Generative Population Models}
\label{sec:cgpms}

In this section we describe composable generative population models (CGPMs), a computational abstraction that provides a uniform treatment of a broad class of models and methods in probabilistic data analysis.
This section is divided into three parts.
The first part formalizes the notion of a statistical population in terms of a random tabular data structure with a finite number of columns and a countably infinite number of rows, and establishes notation used throughout the paper.
The second part outlines the computational interface that defines CGPMs.
The third part describes a class of statistical graphical models which can be naturally expressed using the CGPM framework.


\subsection{Populations}
\label{subsec:cgpms-populations}

In our framework, a population $\mathcal{P}$ is defined in terms of a finite set of variables $(v_1,\dots,v_T)$, where each variable $v_t$ takes values in a general observation space $\mathcal{V}_t$.
Each variable has a qualitative interpretation as a particular property or attribute of the members of the population.
The $r\textsuperscript{th}$ member of the population, denoted $\x_r$, is a $T$-dimensional vector $(x_{[r,1]}, \dots, x_{[r,T]})$, and the element $x_{[r,t]}$ is a variable corresponding to the variable $v_t$ of member $r$.
The entire population is then organized as an infinite exchangeable sequence $(\x_1, \x_2, \dots)$ of members.

The population can be conceptualized as a tabular data structure with a finite number of columns and an infinite number of rows.
Column $t$ corresponds to variable $v_t$, row $r$ to member $\x_r$, and cell $(r,t)$ to element $x_{[r,t]}$.
The table is further associated with the observation spaces $\set{\mathcal{V}_t: t \in [T]}$.
The exchangeability assumption translates into the requirement that $\mathcal{P}$ is unchanged by permuting the member ids.
Finally, a measurement is defined as an observed value for cell $(r,t)$ in the data structure.
In general, we use $x_{[r,t]}$ to indicate the element as a variable as well as its measured value (if one exists); the meaning is typically clear from context.
A collection of measurements recorded in the infinite table is referred to as a dataset $\D$.

It is helpful to compare the standard notion of a statistical population with the formalism described above.
In classical multivariate statistics, a data analysis tasks starts with a ``data matrix'', a finite array containing the measurements from some experiment, and additional modeling assumptions then specify that these measurements are a ``random sample'' from a statistical population.
The members of the population are generated by a distribution (often a multivariate normal) whose unknown parameters (population mean, population covariance, etc) we wish to discover \citep{timm2002,khattree2000,gelman2006}.
This usage of the term ``statistical population'' thus combines domain knowledge (in defining the schema), observed data, and quantitative modeling assumptions (in terms of the random variables) under one umbrella idea.

By contrast, our framing characterizes a population only in terms of a set of population variables and their observation spaces.
This framing does not commit to a probabilistic description of the data generating process, and is intended to invite questions about populations without reference to an underlying statistical model.
Moreover, every member in our definition of a population is associated with a unique identifier -- while this paper only focuses on modeling measurements conditioned on the member ids, in principle the member ids themselves could be modeled by a process that is more complex than random sampling.

Moreover, our mathematical specification of a population attempts to be more granular than the standard formalism from multivariate statistics.
We explicitly differentiate between a variable $v_t$, and the set of elements $\set{x_{[r,t]}:r=1,2,\dots}$ which are versions of that variable $v_t$ for each member.
By separating a variable (a ``column'' in the infinite table) from its related element-level variables (``cells'' in that column), and carefully accounting for all elements in the data structure, we can discuss precisely the mathematical and algorithmic operations performed by CGPMs.
This level of analysis would not be possible had we coarsely specified a population as a single random vector $\x = (x_1,\dots,x_T)$, and viewed measurements collected in a ``data matrix'' as independent realizations of $\x$.
Moreover, specifying measurements at the cell level deals with arbitrary/sparse patterns of observations in the infinite table, in contrast with the standard notion of data matrices which are often treated as objects from linear algebra.
Similarly, explicitly notating the observation spaces $\set{\mathcal{V}_t:t\in[T]}$ allows us to capture heterogeneity in population variables, rather than assume the universe is $T$-dimensional Euclidean space.
These characteristics are common in real-world populations that arise in probabilistic data analysis.


\subsection{%
  Computational description of composable generative population models}
\label{subsec:cgpms-computational}

Having established populations, we now introduce composable generative population models in terms of the computational interface they provide.
A composable generative population model (CGPM) $\G$ characterizes the data generating process for a population $\mathcal{P}$.
The CGPM selects from the population variables $(v_1,v_2,\dots,v_T)$ a set of output variables $(v^{out}_1,\dots,v^{out}_O)$ and a set of input variables $(v^{in}_1,\dots,v^{in}_I)$.
For each member $r$, $\G$ is responsible for modeling the full joint distribution of all the output variables conditioned on all the input variables.
CGPMs differ from the mathematical definition of a probability density in that they are defined directly in terms of a computational interface, as shown in Listing~\ref{lst:cgpm-interface}.
This interface explicitly differentiates between the \textit{sampler} of a random variable from its conditional distribution, and the \textit{assessor} of its conditional density.

\begin{algorithm}[ht]
\floatname{algorithm}{Listing}
\caption{Computational interface for composable generative population models.}
\label{lst:cgpm-interface}
\begin{itemize}[leftmargin=*]
\item $\G \leftarrow$ \texttt{create}(%
    \texttt{population}: $\mathcal{P}$,
    \texttt{outputs}: $\set{v^{out}_i}_{i \in [O]}$,
    \texttt{inputs}: $\set{v^{in}_j}_{j \in [I]}$,
    \texttt{binary}: $\mathcal{B}$,
    \texttt{seed}: $s$)

Create a CGPM for the population, with the specified inputs and outputs.

\item $\mathbf{s} \leftarrow$ \texttt{simulate}
  ($\G$, \texttt{member}: $r$, \texttt{query}: $Q=\set{q_k}$,
  \texttt{evidence} : $E=\set{x_{[r,e_j]}}\cup \y_r$)

    Generate a sample from the distribution \hfill
    $\mathbf{s} \sim^\G \x_{[r,Q]}|\set{\x_{[r,E]},\y_r,\D}.$

\item $c \leftarrow$ \texttt{logpdf}
  ($\G$, \texttt{member}: $r$, \texttt{query} : $Q=\set{x_{[r,q_k]}}$,
    \texttt{evidence} : $E=\set{x_{[r,e_j]}}\cup \y_r$)

    Evaluate the log density \hfill
    $\log \pG(\x_{[r,Q]}|\set{\x_{[r,E]},\y_r,\D}).$

\item $\G' \leftarrow $ \texttt{incorporate}
  ($\G$, \texttt{measurement} : $x_{[r,k]}$)

    Record a measurement $x_{[r,k]}\in \mathcal{V}_k$ into the dataset $\D$.

\item $\G' \leftarrow $ \texttt{unincorporate}
  ($\G$, \texttt{member} : $r$)

    Eliminate all measurements of input and output variables for member $r$.

\item $\G' \leftarrow $ \texttt{infer}
  ($\G$, \texttt{program} : $\mathcal{T}$)

    Adjust internal state in accordance with the learning procedure
    specified by program $\mathcal{T}$.
\end{itemize}
\end{algorithm}

There are several key ideas to draw from the interface.
In \texttt{create}, $\mathcal{P}$ contains the set of all population variables and their observation spaces.
The \texttt{binary} is an opaque probabilistic program containing implementations of the interface, and \texttt{seed} is the entropy source from which the CGPM draws random bits.
The \texttt{outputs} requires at least one entry, the \texttt{inputs} may be an empty set, and any variable which is neither an input nor an output is unmodeled by the CGPM.
For simplicity, we use the symbol $x_{[r,t]}$ to denote the output variable $x_{[r,v^{out}_t]}$ and similarly $y_{[r,t]}$ for input variable $y_{[r,v^{in}_t]}$ of member $r$.
These elements are often collected into vectors $\x_r$ and $\y_r$, respectively

In \texttt{incorporate}, measurements are recorded at the cell-level, allowing only a sparse subset of observations for member $r$ to exist. The \texttt{measurement} may be either an output element from $\x_r$ or input element from $\y_r$.

Both \texttt{simulate} and \texttt{logpdf} are computed for single member $r$ of the population.
The \texttt{query} parameter differs between the two methods: in \texttt{simulate}, $Q=\set{q_k}$ is a set of indices of output variables that are to be simulated jointly; in \texttt{logpdf}, $Q=\set{x_{[r,q_k]}}$ is a set of values for the output variables whose density is to be assessed jointly.
The \texttt{evidence} parameter is the same for both \texttt{simulate} and \texttt{logpdf}, which contains additional information about $r$, possibly including the values of a set of output variables that are disjoint from the query variables.
In particular, if $x_{[r,E]}$ is empty, the CGPM is asked to marginalize over all its output variables that are not in the query $Q$; if $x_{[r,E]}$ is not empty, the CGPM is required to condition on those output values.

The target distributions in \texttt{simulate} and \texttt{logpdf} are also conditioned on all previously incorporated measurements in the dataset $\D$.
Because CGPMs generally model populations with inter-row dependencies, measurements of other members $s\ne r$ are relevant to a \texttt{simulate} or \texttt{logpdf} query about $r$.
The CGPM interface allows the user to override a previous measurement of $r$ in $\D$ on a per-query basis; this occurs when an element $x_{[r,e_j]}$ or $\y_r$ in the \texttt{evidence} contradicts an existing measurement $x'_{[r,e_j]}$ or $\y'_r$ in $\D$.
Asking such hypothetical queries addresses several tasks of interest in probabilistic data analysis, such as simulating ``what-if'' scenarios and detecting outliers in high-dimensional populations.

Finally, the \texttt{infer} procedure evolves the CGPM's internal state in response to the inflow of measurements.
The inference program $\mathcal{T}$ can be based on any learning strategy applicable to the CGPM, such as Markov Chain Monte Carlo transitions, variational inference, maximum-likelihood, least-squares estimation, or no learning.


\subsection{Statistical description of composable generative population models}
\label{subsec:cgpms-statistical}

\begin{figure}[ht]
\centering

\begin{tikzpicture}

\node[latent]                  (x)           {$\x_r$};
\node[latent, above=of x]      (z)           {$\z_r$};
\node[obs, left=of z, fill=]   (y)           {$\y_r$};
\node[latent, right=of z]      (theta)       {$\btheta$};
\node[const, below=of theta]   (alpha)       {$\balpha$};
\node[const, above=1 of y]     (input)       {\tt inputs};
\node[const, below=1.75 of x]   (output)      {\tt outputs};

\edge[>=stealth] {alpha} {theta} ; %
\edge[>=stealth] {input} {y} ; %
\edge[>=stealth] {x} {output} ; %
\edge[>=stealth] {theta} {x,z} ; %
\edge[>=stealth] {z} {x} ; %
\edge[>=stealth] {y} {z,x} ; %

\draw[very thick, dashed] (z) to [out=110,in=70,looseness=7] (z);

\plate[yshift=.15cm] {xzy-pl} {(x)(z)(y)} {$r=1,2,\dots$};

\plate[inner sep=.15cm, yshift=.2cm]
  {all-pl} {(xzy-pl)(x)(y)(z)(theta)(alpha)} {$\G$};

\end{tikzpicture}

\bcaption{%
Internal independence constraints for a broad class of composable generative population models.}{%
All nodes in the diagram are multidimensional.
Internally, the hyperparameters $\balpha$ are fixed and known quantities.
The global latents $\btheta$ are shared by all members of the population.
Member-specific latents $\z_r$ interact only with their corresponding observations $\x_r$, as well as other member-latents $\set{\z_{s}:s\ne r}$ as indicated by the dashed loop around the plate.
Nodes $\x_r$ and $\x_s$ across different members $r$ and $s$ are independent conditioned on their member-latents.
However, general dependencies are permitted within elements $\set{x_{[r,i]}:i\in[O]}$ of node $\x_r$.
The input variables $\y_r$ are ambient conditioning variables in the population and are always observed; in general, $\y_r$ may be the output of another CGPM (Section \ref{subsec:composition}).
Externally, $\G$ is specified by an opaque \texttt{binary}, e.g. a probabilistic program, describing the data generating process, and \texttt{outputs} and \texttt{inputs} that specify the variable names for \texttt{simulate} and \texttt{logpdf}.
}
\label{fig:cgpm-graphical}
\end{figure}

The previous section outlined the external interface that defines a CGPM without specifying its internal structure.
In practice, many CGPMs can be described using a general graphical model with both directed and undirected edges.
The data generating process is characterized by a collection of variables in the graph,
\begin{equation*}
\G=(\balpha, \btheta, \Z=\set{\z_r}_{r=1}^{\infty}, \X=\set{\x_r}_{r=1}^{\infty},
    \Y=\set{\y_r}_{r=1}^{\infty}).
\end{equation*}
\begin{itemize}
\item $\balpha$:
Fixed quantities such as input and output dimensionalities, observation spaces, dependence structures and statistical hyperparameters.

\item $\btheta$:
Population-level, or global, latent variables relevant to all members.

\item $\z_r = (z_{[r,1]},\dots,z_{[r,L]})$:
Member-specific latent variables governing only member $r$ directly. A subset of these variables may be exposed, and treated as queryable output variables.

\item $\x_r = (x_{[r,1]},\dots,x_{[r,O]})$:
Output variables representing observable attributes of member $r$.

\item $\y_r = (y_{[r,1]},\dots y_{[r,I]})$:
Input variables that must be present for any query about $\x_r$, such as the ``feature vectors'' in a discriminative model.
\end{itemize}

The notion of global and local latent variables is a common motif in the hierarchical modeling literature \citep{blei2016}.
They are useful in specifying the set of constraints governing the dependence between observable variables in terms of some latent structure.
From this lens, CGPMs satisfy the following conditional independence constraint,
\begin{align}
\forall r\ne s \in \mathbb{N}, \forall j,k \in [O]: x_{[r,j]}
    \indep x_{[s,k]} \mid \set{\balpha, \btheta, \z_r, \z_s}.
\label{eq:indep-gpm}
\end{align}
Equation \eqref{eq:indep-gpm} formalizes the notion that all dependencies across members $r\in\mathbb{N}$ are fully mediated by the global parameters $\btheta$ and member-specific variables $\set{\z_r}$.
However, elements $x_{[r,j]}$ and $x_{[r,i]}$ within a member are free to assume any dependence structure, allowing for arbitrary inter-row dependencies.
This feature allows CGPMs to express undirected models where the output variables are not exchangeably-coupled, such as Gaussian Markov random fields \citep{rue2005}.

A common specialization of constraint \eqref{eq:indep-gpm} further requires that the member-specific latent variables $\set{\z_r}$ are conditionally independent given $\btheta$; a comprehensive list of models in machine learning and statistics satisfying this additional constraint is given in \citep[Section 2.1]{hoffman2013}.
However, CGPMs permit more general dependencies in that member latents may be coupled conditioned $\btheta$, thus allowing for complex intra-row dependencies.
CGPMs can thus be used for models such as Gaussian process regression with noisy observations \citep{rasmussen2006}, where the member-specific latent variables (i.e. the noiseless observations) across different members in the population are jointly Gaussian \citep[Figure 1]{damianou2013}.

Figure~\ref{fig:cgpm-graphical} summarizes these ideas by showing a CGPM as a graphical model.
Finally, we note it is also possible for a CGPM to fully implement the interface without admitting a ``natural'' representation in terms of the graphical structure from Figure~\ref{fig:cgpm-graphical}, as shown by several examples in Section~\ref{sec:implementation}.


\subsection{Composable generative population models are an abstraction for probabilistic processes}
\label{sec:cgpms-processes}

By providing a computational interface, the CGPM interface provides a layer of abstraction which separates the internal implementation of a probabilistic model from the generative process it represents.
In this section we will explore how the computational (external) description of a CGPM provides a fundamentally different understanding than its statistical (internal) description.

As an example, consider a Dirichlet process mixture model \citep{antoniak1974} expressed as a CGPM.
The hyperparameters $\balpha=(H,\gamma,F)$ are the base measure $H$, concentration parameter $\gamma$, and parametric distribution $F$ of the observable variables $\set{\x_r}$.
The member latent variable $\z_r=(z_r)$ is the cluster assignment of $r$.
Consider now two different representations of the underlying DP, each leading to a different notion of (i) population parameters $\btheta$, and (ii) conditional independence constraints.

\begin{itemize}
\item
In the stick breaking representation \citep{sethuraman1994}, the population parameters $\btheta=\set{(\phi_i,\pi_i): i\in\mathbb{N}}$, where $\phi_i$ are the atoms that parameterize the likelihood $F(\cdot|\phi_i)$ (drawn i.i.d from $H$) and $\pi_i$ their weights (drawn jointly from GEM($\gamma$)).
Conditioned on $\set{\balpha, \btheta}$, the member latents are independent, $z_r \sim^{iid} \textsc{Categorical}(\set{\pi_1,\pi_2,\dots})$.

\item
In the Chinese restaurant process representation \citep{aldous1985}, the population parameters $\btheta=\set{\phi_i:i\in\mathbb{N}}$ are now only the atoms, and the weights are fully collapsed out.
Conditioned on $\set{\balpha, \btheta}$, the member latents are exchangeably coupled $\set{z_1,z_2,\dots} \sim \textsc{Crp}(\gamma)$.
\end{itemize}

These internal representation choices are not exposed by the CGPM interface and may be interchanged without altering the queries it can answer.%
\footnote{However, it is important to note that interchanging representations may result in different performance characteristics, such as compute time or approximateness of \texttt{simulate} and \texttt{logpdf}.}
It follows that the computational description of CGPMs provides an abstraction boundary between a particular implementation of a probabilistic model and the generative process for the population that it represents.
Two implementations of a CGPM may encapsulate the same process by inducing an identical marginal distribution over their observable variables, while maintaining different auxiliary-variable representations internally.

The encapsulation of a CGPM's internal state can be relaxed by asking the CGPM to expose member-specific latent variables as outputs.
In terms of the infinite table metaphor from Section~\ref{subsec:cgpms-populations}, this operation may be conceptualized as the CGPM ``fantasizing'' the existence of new columns in the underlying population.
Providing a gateway into the internal state of a CGPM trades-off the model independence of the interface with the ability to query the hidden structure of a particular probabilistic process.
Section~\ref{sec:bayesdb} describes surface-level syntaxes for exposing latent variables, and Section~\ref{subsec:experiment-satellites} illustrates its utility for inferring latent cluster assignments in an infinite mixture model, as well simulating projections of high-dimensional data onto low-dimensional latent subspaces.


\section{Algorithmic Implementations of Composable Generative Population Models}
\label{sec:implementation}

In this section, we illustrate that the computational abstraction of CGPMs is applicable to broad classes of modeling approaches and philosophies.
Table~\ref{tab:cgpm-examples} shows the collection of models whose internal structure we will develop from the perspective of CGPMs.
Section~\ref{sec:applications} shows both comparisons of these CGPMs and their practical application to data analysis tasks.

\begin{table}[ht]
\centering
\small
\begin{tabular*}{\textwidth}{@{}lll@{}}
  \toprule
    & \textbf{Composable Generative Population Model}
    & \textbf{Modeling Approach} \\ \midrule

  Section~\ref{subsec:implementation-crosscat}
  & Cross Categorization
  & non-parametric Bayesian generative modeling \\

  Section~\ref{subsec:implementation-ensemble}
  & Ensemble Classifiers and Regressors
  & discriminative machine learning \\

  Section~\ref{subsec:implementation-pca}
  & Factor Analysis \& Probabilistic PCA
  & dimensionality reduction \\

  Section~\ref{subsec:implementation-experts}
  & Parametric Mixture of Experts
  & discriminative statistical modeling \\

  Section~\ref{subsec:implementation-kde}
  & Multivariate Kernel Density Estimation
  & classical multivariate statistics \\

  Section~\ref{subsec:implementation-knn}
  & Generative Nearest Neighbors
  & clustering based generative modeling \\

  Section~\ref{subsec:implementation-venturescript}
  & Probabilistic Programs in VentureScript
  & probabilistic programming \\ \bottomrule

\end{tabular*}
\bcaption{%
Examples of composable generative population models, and a modeling framework for data analysis to which they belong.}{}
\label{tab:cgpm-examples}
\end{table}

The two methods from the interface in Listing~\ref{lst:cgpm-interface} whose algorithmic implementations we outline for each CGPM are

\begin{itemize}
\item $\mathbf{s} \leftarrow$ \texttt{simulate}
  ($\G$, \texttt{member}: $r$, \texttt{query}: $Q=\set{q_k}$,
  \texttt{evidence} : $E=\set{x_{[r,e_j]}}\cup \y_r$)

    Generate a sample from the distribution \hfill
    $\mathbf{s} \sim^\G \x_{[r,Q]}|\set{\x_{[r,E]},\y_r,\D}.$

\item $c \leftarrow$ \texttt{logpdf}
  ($\G$, \texttt{member}: $r$, \texttt{query} : $Q=\set{x_{[r,q_k]}}$,
    \texttt{evidence} : $E=\set{x_{[r,e_j]}}\cup \y_r$)

    Evaluate the log density \hfill
    $\log \pG(\x_{[r,Q]}|\set{\x_{[r,E]},\y_r,\D}).$
\end{itemize}

In both \texttt{simulate} and \texttt{logpdf}, the target distributions for the query variables $\x_{[r,Q]}$ require an implementation of two operations:
\begin{itemize}
\item
Conditioning on the evidence variables $\x_{[r,E]}$, in addition to the input variables $\y_r$ and entire measurement set $\D$.

\item
Marginalizing over all output variables $\set{x_{[r,i]}: i\in[O]\backslash(E\cup{Q})}$ not in the query or evidence.
\end{itemize}
Both conditioning and marginalizing over joint distributions allow users of CGPMs to pose non-trivial queries about populations that arise in multivariate probabilistic data analysis.
All our algorithms generally assume that the information known about member $r$ in \texttt{simulate} and \texttt{logpdf} is only what is provided for the \texttt{evidence} parameter.
Extending the implementations to deal with observed members $r'\in\D$ is mostly straightforward and often implementation-specific.
We also note that the figures in these subsections contain excerpts of probabilistic code in the Bayesian Query Language, Metamodeling Language, and VentureScript; most of their syntaxes are outlined in Section~\ref{sec:bayesdb}.
Finally, we leave the many possible implementations of \texttt{infer} for each CGPM, which learns the latent state using observed data, primarily to external references.


\subsection{Primitive univariate distributions and statistical data types}
\label{subsec:implement-primitive}

The statistical data type of a population variable $v_t$ provides a more refined taxonomy than the ``observation space'' $\mathcal{V}_t$ described in Section~\ref{subsec:cgpms-populations}.
Table~\ref{tab:stattypes} shows the collection of statistical data types available in the Metamodeling Language (Section~\ref{subsec:bayesdb-mml}), out of which more complex CGPMs are built.
The (parameterized) support of a statistical type defines the set in which samples from \texttt{simulate} take values.
Each statistical type is also associated with a base measure which ensures \texttt{logpdf} is well-defined.
In \textrm{high-dimensional} populations with heterogeneous types, \texttt{logpdf} is taken against the product measure of these univariate base measures.
The statistical type also identifies invariants that the variable maintains.
For instance, the values of a \texttt{NOMINAL} variable are permutation-invariant; the distance between two values for a \texttt{CYCLIC} variable is defined circularly (modulo the period), etc.
The final column in Table~\ref{tab:stattypes} shows the primitive univariate CGPMs that are compatible with each statistical type.
For these simple CGPMs, \texttt{logpdf} is implemented directly from their probability density functions, and algorithms for \texttt{simulate} are well-known \citep{devroye1986}.
For \texttt{infer}, the CGPMs may have fixed parameters, or learn from data using i.e. maximum likelihood \citep[Ch. 7]{casella2002} or Bayesian priors \citep{fink1997}.


\begin{table}[ht]
\centering
\footnotesize
    \begin{tabular}{@{}llllll@{}}
    \toprule
    \textbf{Statistical Data Type}
        & \textbf{Parameters}
        & \textbf{Support}
        & \textbf{Measure/$\sigma$-Algebra}
        & \textbf{Primitive Univariate CGPM}
        \\ \midrule
    \texttt{BINARY}
        & -
        & $\set{0,1}$
        & $(\#, 2^{\set{0,1}})$
        & \texttt{BERNOULLI}
        \\
    \texttt{NOMINAL}
        & symbols: $S$
        & $\set{0,1,\dots,S-1}$
        & $(\#, 2^{[S]})$
        & \texttt{CATEGORICAL}
        \\
    \texttt{COUNT/RATE}
        & base: $b$
        & $\set{0,\frac{1}{b},\frac{2}{b},\dots}$
        & $(\#, 2^{\mathbb{N}})$
        & \texttt{POISSON}, \texttt{GEOMETRIC}
        \\
    \texttt{CYCLIC}
        & period: $p$
        & $(0,p)$
        & $(\lambda, \mathcal{B}(\R))$
        & \texttt{VON-MISES}
        \\
    \texttt{MAGNITUDE}
        & --
        & $(0,\infty)$
        & $(\lambda, \mathcal{B}(\R))$
        & \texttt{LOGNORMAL}, \texttt{EXPONENTIAL}
        \\
    \texttt{NUMERICAL}
        & --
        & $(-\infty,\infty)$
        & $(\lambda, \mathcal{B}(\R))$
        & \texttt{NORMAL}
        \\
    \texttt{NUMERICAL-RANGED}
        & low: $l$, high:$h$
        & $(l, h) \subset \R$
        & $(\lambda, \mathcal{B}(\R))$
        & \texttt{BETA}, \texttt{NORMAL-TRUNC}
        \\
    \bottomrule
    \end{tabular}
    \captionsetup{skip=2pt}
    \bcaption{%
    Statistical data types, and their supports, base measures, and primitive CGPMs.}{}
    \label{tab:stattypes}
\end{table}%
\vspace{-.75cm}
\begin{figure}[ht]
    \centering
    \begin{subfigure}{\textwidth}
        \centering
        \includegraphics[width=.32\textwidth]
            {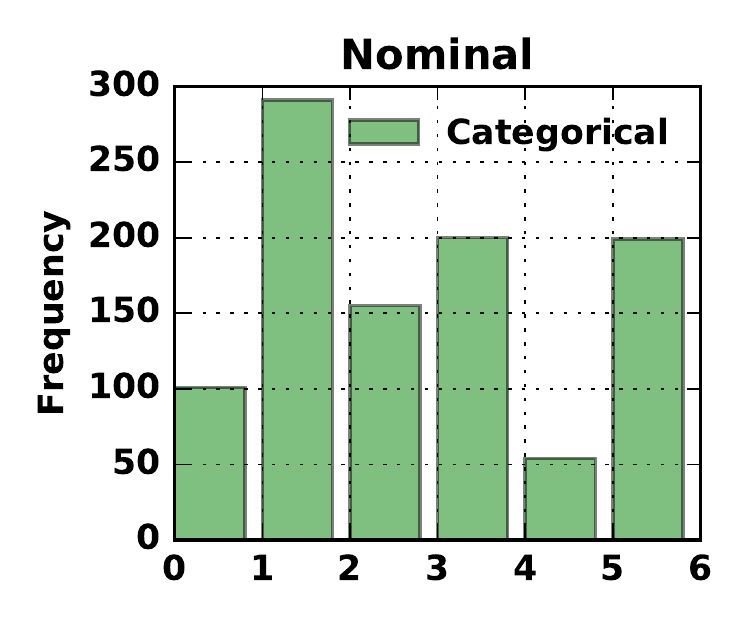}
        \includegraphics[width=.32\textwidth]
            {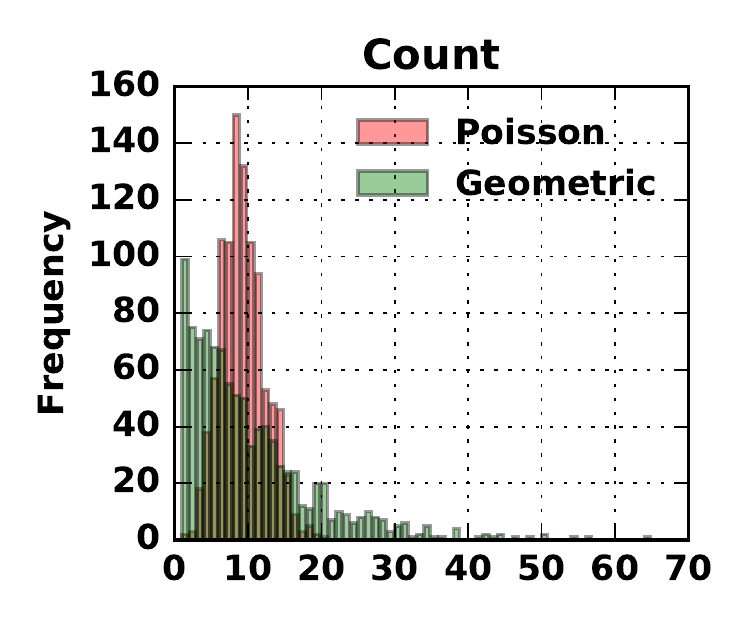}
        \includegraphics[width=.32\textwidth]
            {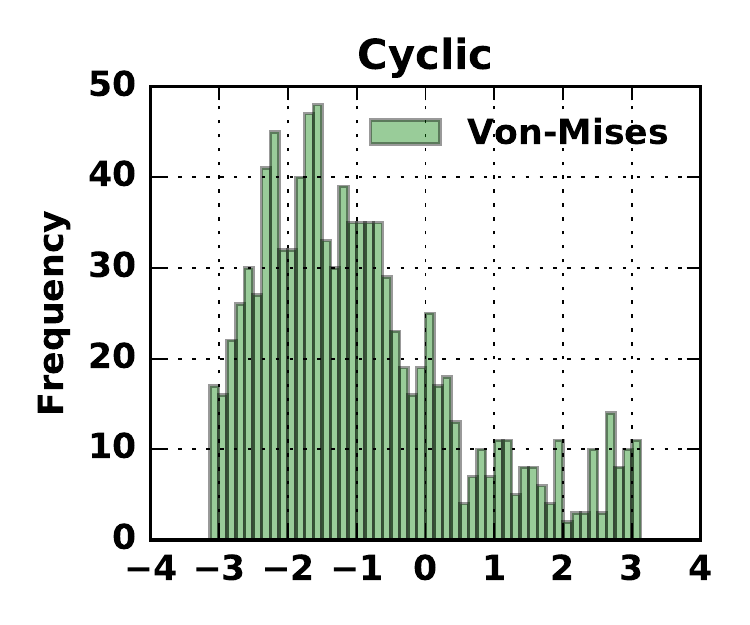}
    \end{subfigure}
    \begin{subfigure}{\textwidth}
        \centering
        \includegraphics[width=.32\textwidth]
            {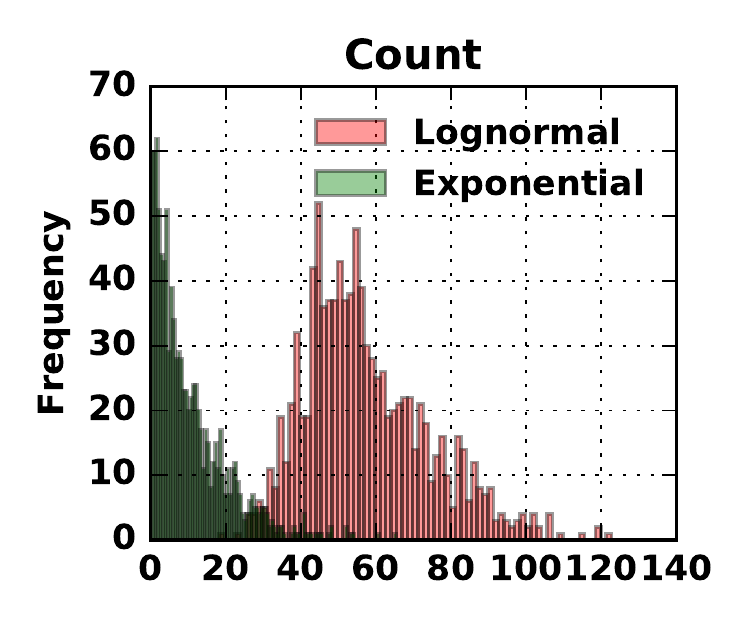}
        \includegraphics[width=.32\textwidth]
            {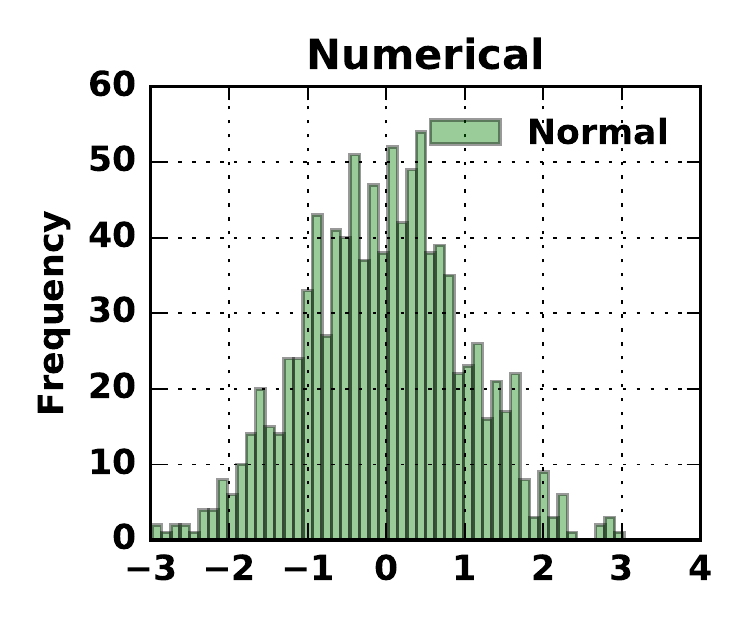}
        \includegraphics[width=.32\textwidth]
            {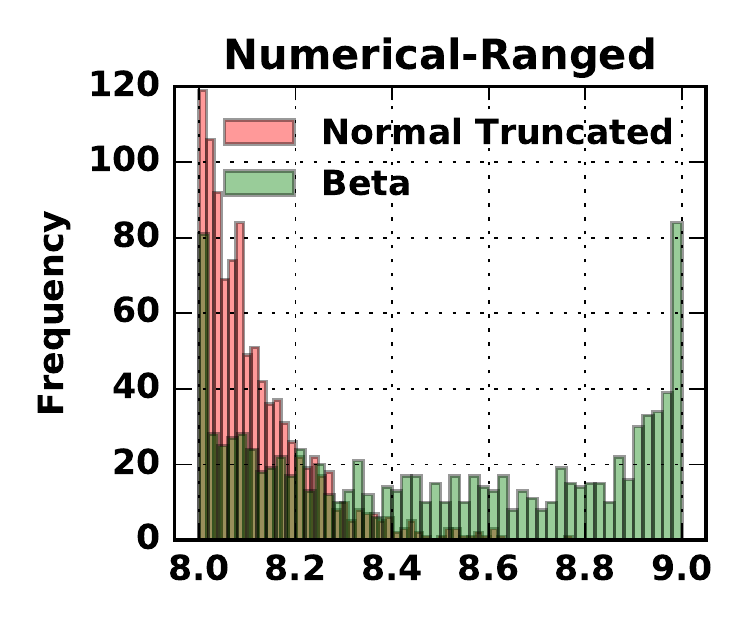}
    \end{subfigure}
    \captionsetup{skip=2pt}
    \bcaption{%
    Samples from the primitive CGPMs of each statistical data type.}{}
\end{figure}


\subsection{Cross-Categorization}
\label{subsec:implementation-crosscat}

Cross-Categorization (CrossCat) is a Bayesian non-parametric method for learning the joint distribution over all variables in a heterogeneous, high-dimensional population \citep{mansinghka2015-2}.
The generative model begins by first partitioning the set of variables $(v_1,\dots,v_T)$ into blocks.
This step is CrossCat's ``outer'' clustering, since it partitions the ``columns''(when viewing the population in terms of its infinite table representation from Section~\ref{subsec:cgpms-populations}).
Let $\pi$ denote the variable partition, and $\set{B_i: i \in |\pi|}$ denote its blocks.
$\pi$ is a global latent variable which dictates the structural dependencies between variables; any collection of variables in different blocks are mutually independent, and all variables in the same block are mutually dependent.
It follows that for each member $r$, the joint distribution for $\x_r$ factorizes,
\begin{align*}
\pG(\x_r|\btheta) = \prod_{B\in\pi}\pG(\x_{[r,B]}|\btheta_B).
\end{align*}
The bundle of global parameters $\btheta$ includes $\pi$ as well as a set of block-specific latent variables $\set{\btheta_B}_{B\in\pi}$.
Within each block $B$ of dependent variables, the elements $\set{x_{[r,i]}, i \in B}$ are conditionally independent given a member-specific latent variable $z_{[r,B]} \in \mathbb{N}$.
This variable is an ``inner'' clustering assignment in CrossCat, since it specifies the cluster identity of row $r$ with respect to the variables in block $B$.
The joint distribution over elements then factorizes,
\begin{align}
\pG(\x_{[r,B]}|\btheta_B)
    = \sum_{k}
        \left[
            \left(
                \prod_{i\in{B}} \pG(x_{[r,i]}|\phi_{[i,k]})
            \right)
            \pG(z_{[r,B]}=k|\bm\omega_B)
        \right].
    \label{eq:crosscat-naive-bayes}
\end{align}
The global parameter $\phi_{[i,k]}$ parameterizes the primitive univariate CGPM (of the appropriate statistical type) for $v_i$ in cluster $k$, and $\bm\omega_B$ is a parameter governing the distribution of the latent variable $z_{[r,B]}$.
This description fully specifies the CrossCat factorization of the joint distribution $\pG(\x_r|\btheta)$.
This generative template is encoded into a hierarchical Bayesian model by specifying priors over the partition $\pi$, mixture weights $\bm\omega_B$ in each block $B\in\pi$, and distributional parameters $\phi_{[i,k]}$.
In contrast to \citep{mansinghka2015-2}, Algorithm~\ref{alg:crosscat-sample} presents (for simplicity) a fully uncollapsed representation of the CrossCat prior, using a GEM distribution \citep{pitman2002} for the inner DP.

Having described the generative process and established notation, we now outline algorithms for \texttt{logpdf} and \texttt{simulate}.
Since CrossCat is a Bayesian CGPM, the distribution of interest $\pG(\x_{[r,Q]}|\x_{[r,E]},\D)$ requires us to marginalize out the latent variables $(\btheta,\Z)$.
Sampling from the posterior is covered in \citep[Section 2.4]{mansinghka2015-2}, so we only focus on implementing \texttt{simulate} and \texttt{logpdf} assuming posterior samples of latents are available.%
\footnote{Section~\ref{subsec:implementation-venturescript} outlines the Monte Carlo estimator for aggregating the samples in a general probabilistic programming setting.}
These implementations are summarized in Algorithms \ref{alg:crosscat-simulate} and \ref{alg:crosscat-logpdf}, where all routines have access to a posterior sample of the latent variables in Algorithm~\ref{alg:crosscat-sample}.
While our algorithms are based on an uncollapsed CrossCat, in practice, the \textsc{Parameter-Prior} and primitive CGPMs from lines \ref{alg-line:crosscat-parameter-prior} and \ref{alg-line:crosscat-parameter-likelihood} in Algorithm~\ref{alg:crosscat-sample} form a conjugate pair.
The density terms $\pG(x_{[r,c]}|\phi_{[c,k]})$ are computed by marginalizing $\phi_{[c,k]}$, and using the sufficient statistics in cluster $k$ along with the column hyperparameters $\bm\lambda_i$, i.e. $\pG(x_{[r,c]}|\set{x_{[r',c]}: z_{[r',B]}=k}, \bm\lambda_i)$.
This Rao-Blackwellization enhances the inferential quality and predictive performance of CrossCat, and the one sample approximation on line \ref{alg-line:crosscat-approximate} of Algorithm~\ref{alg:crosscat-weights}, an instance of Algorithm 8 from \citep{neal2000}, becomes exact for evaluating \texttt{logpdf}.
Section~\ref{sec:cgpms-processes} contains a discussion on the implications of different internal representations of a generative process (such as collapsed or uncollapsed) from the perspective of CGPMs.

\clearpage
\begin{subalgorithms}

\captionof{algorithm}{Forward sampling a population in the CrossCat CGPM.}
\label{alg:crosscat-sample}
\begin{algorithmic}[1]
    \small
    \State $\alpha \sim \textsc{Crp-Concentration-Prior}$
        \Comment{sample a concentration for the outer CRP}
    \State $\pi \sim \textsc{Crp}(\alpha|[T])$
        \Comment{sample partition of variables $\set{v_1,\dots,v_T}$}
    \For{$B \in \pi$}
        \Comment{for each block $B$ in the variable partition}
        \State $\alpha_B \sim \textsc{Crp-Concentration-Prior}$
        \Comment{sample a concentration for the inner CRP at $B$}
        \State $(\omega_{[B,1]}, \omega_{[B,2]}, \dots)
                \sim \text{GEM}(\alpha_B)$
            \Comment{sample stick-breaking weights of its clusters}
    \EndFor
    \For{$i \in [T]$}
            \Comment{for each variable $v_i$ in the population}
        \State $\bm\lambda_i \sim \textsc{Parameter-Hyper-Prior}$
            \Comment{sample hyperparams from a hyperprior}
        \State $(\phi_{[i,1]},\phi_{[i,2]},\dots)
            \overset{iid}{\sim} \textsc{Parameter-Prior($\bm\lambda_i$)}$
            \label{alg-line:crosscat-parameter-prior}
            \Comment{sample component distribution params}
    \EndFor
    \For{$r = 1,2,\dots$}
        \Comment{for each member $r$ in the population}
        \For{$B \in \pi$}
            \Comment{for each block $B$ in the variable partition}
            \State $z_{[r,B]} \sim \textsc{Categorical}(\bm\omega_B)$
            \Comment{sample the cluster assignment of $r$ in $B$}
            \For{$i \in B$}
                \Comment{for each variable $v_i$ in the block}
                \State $x_{[r,i]} \sim \pG(\cdot | \phi_{[i,z_{[r,B]}]})$
                \label{alg-line:crosscat-parameter-likelihood}
                \Comment{sample observable element $v_i$ for $r$}
            \EndFor
        \EndFor
    \EndFor
\end{algorithmic}
\captionof{algorithm}{\texttt{simulate} for the CrossCat CGPM.}
\label{alg:crosscat-simulate}
\begin{algorithmic}[1]
  \small
  \Function{Simulate}{}
  \State $\x_{[r,Q]} \gets \varnothing$
    \Comment{initialize empty sample}
  \For{$B \in \pi$}
        \Comment{for each block $B$ in the variable partition}
    \State $\bm{l} \gets \textsc{Compute-Cluster-Probabilities}(B)$
        \Comment{retrieve posterior probabilities of proposal clusters}
    \State $z_{[r,B]} \sim \textsc{Categorical}(\bm{l})$
        \Comment{sample a cluster}
    \For{$q \in (Q \cap B)$}
        \Comment{for each query variable in the block}
        \State $x_{[r,q]} \sim \pG(\cdot|\phi_{[q,z_{[r,B]}]})$
            \Comment{sample an observation element}
    \EndFor
  \EndFor
  \State \Return $\x_{[r,Q]}$
        \Comment{overall sample of query variables}
  \EndFunction
\end{algorithmic}
\captionof{algorithm}{\texttt{logpdf} for the CrossCat CGPM.}
\label{alg:crosscat-logpdf}
  \small
  \begin{algorithmic}[1]
  \Function{LogPdf}{}
  \For{$B \in \pi$}
    \Comment{for each block $B$ in the variable partition}
    \State $\bm{l} \gets \textsc{Compute-Cluster-Probabilities}(B)$
        \Comment{retrieve posterior probabilities of proposal clusters}
    \State $K \gets |\bm{l}|$
        \Comment{compute number of proposed clusters}
    \State $t_B \gets
            \sum_{k=1}^{K}
            \left[
                \left(
                    \prod\limits_{q\in{(Q\cap B)}}\pG(x_{[r,q]}|\phi_{[r,k]})
                \right)
                \frac{l_k}{\sum_{k'=1}^{K}l_{k'}}
            \right]$
        \Comment{compute density for query variables in $B$}
  \EndFor
  \State \Return $\sum_{B\in\pi}\log(t_B)$
        \Comment{overall log density of query}
  \EndFunction
  \end{algorithmic}
\captionof{algorithm}
  {Computing the cluster probabilities in a block of the CrossCat partition.}
\label{alg:crosscat-weights}
\begin{algorithmic}[1]
  \small
  \Function{Compute-Cluster-Probabilities}{} (\texttt{block}: $B$)
    \State $K \gets \underset{r'\in \D}{\max}\set{z_{[r',B]}}$
        \Comment{compute number of occupied clusters}
    \For{$k=1,2,\dots,K$}
        $c_k = |\set{r'\in \D: z_{[r',B]} = k}|$
        \Comment{compute number of members in each cluster}
    \EndFor
    \For{$k=1,2,\dots,K$}
        \Comment{for each cluster $k$}
        \State $l_{k} \gets
            \left(\frac{c_{k}}{\sum_{j}c_j+\alpha_B}\right)
            \prod\limits_{e\in{(E \cap B)}}\pG(x_{[r,e]}|\phi_{[e,k]})$
        \Comment{compute probability of $r$ joining $k$}
    \EndFor
    \State $l_{K+1} \gets \left(\frac{\alpha_B}{\sum_{j}c_j+\alpha_B}\right)
            \prod\limits_{e\in{(E \cap B)}}\pG(x_{[r,e]}|\phi_{[e,K+1]})$
            \Comment{compute probability of $r$ in singleton cluster}
            \label{alg-line:crosscat-approximate}
    \State \Return{ $(l_{1},\dots,l_{K}, l_{K+1})$}
        \Comment{normalized probabilities of proposed clusters}
  \EndFunction
\end{algorithmic}

\end{subalgorithms}

\clearpage

\begin{figure}[ht]
\begin{subfigure}{\textwidth}
    \begin{subfigure}{0.125\textwidth}
        \includegraphics[width=\textwidth]{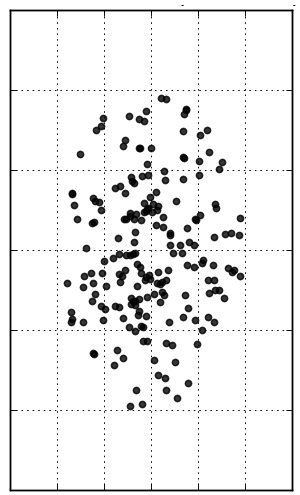}
    \end{subfigure}%
    \begin{subfigure}{0.125\textwidth}
        \includegraphics[width=\textwidth]{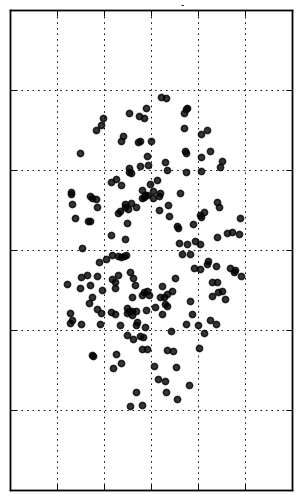}
    \end{subfigure}%
    \begin{subfigure}{0.125\textwidth}
        \includegraphics[width=\textwidth]{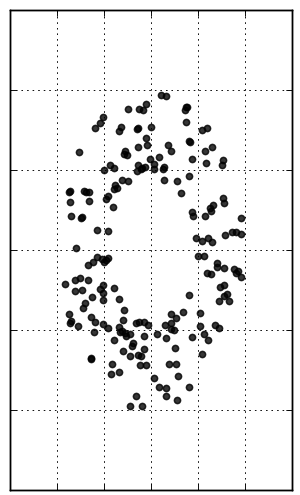}
    \end{subfigure}%
    \begin{subfigure}{0.125\textwidth}
        \includegraphics[width=\textwidth]{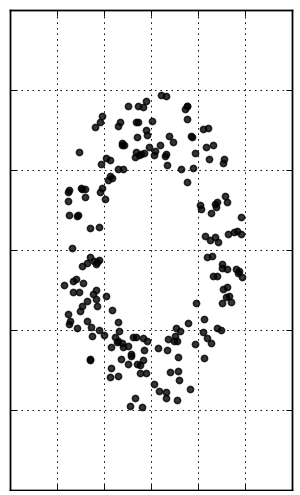}
    \end{subfigure}%
    \begin{subfigure}{0.125\textwidth}
        \includegraphics[width=\textwidth]{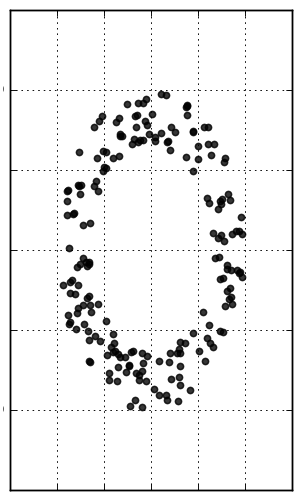}
    \end{subfigure}%
    \begin{subfigure}{0.125\textwidth}
        \includegraphics[width=\textwidth]{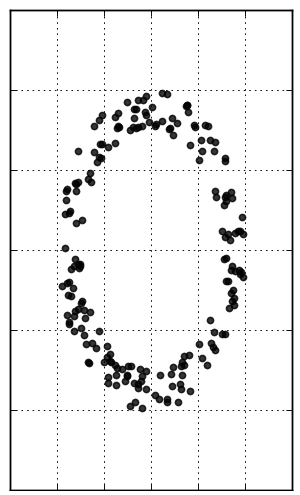}
    \end{subfigure}%
    \begin{subfigure}{0.125\textwidth}
        \includegraphics[width=\textwidth]{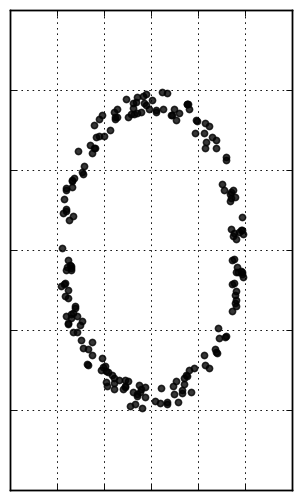}
    \end{subfigure}%
    \begin{subfigure}{0.125\textwidth}
        \includegraphics[width=\textwidth]{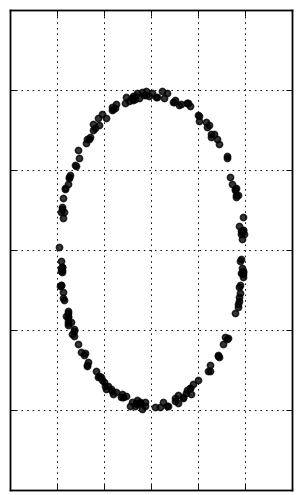}
    \end{subfigure}
    \begin{subfigure}{0.125\textwidth}
        \includegraphics[width=\textwidth]{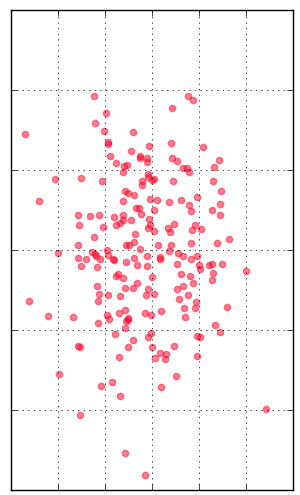}
    \end{subfigure}%
    \begin{subfigure}{0.125\textwidth}
        \includegraphics[width=\textwidth]{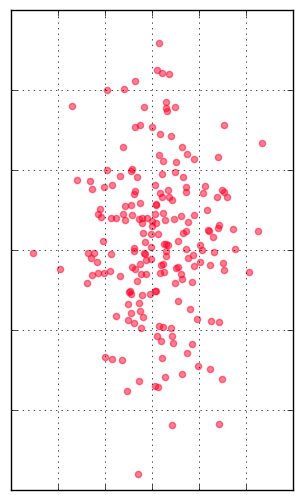}
    \end{subfigure}%
    \begin{subfigure}{0.125\textwidth}
        \includegraphics[width=\textwidth]{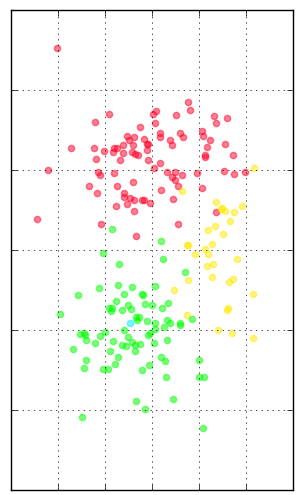}
    \end{subfigure}%
    \begin{subfigure}{0.125\textwidth}
        \includegraphics[width=\textwidth]{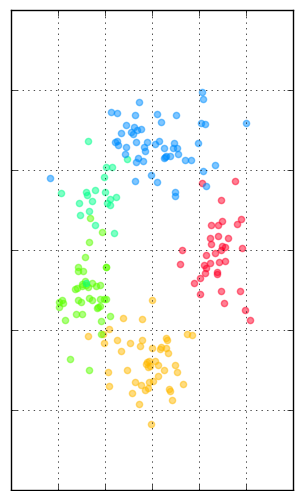}
    \end{subfigure}%
    \begin{subfigure}{0.125\textwidth}
        \includegraphics[width=\textwidth]{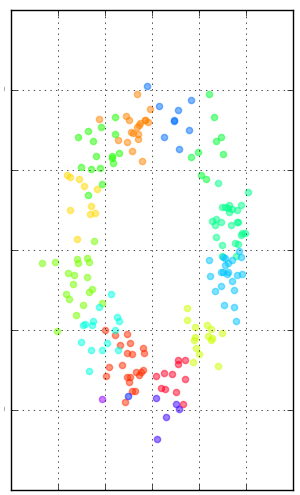}
    \end{subfigure}%
    \begin{subfigure}{0.125\textwidth}
        \includegraphics[width=\textwidth]{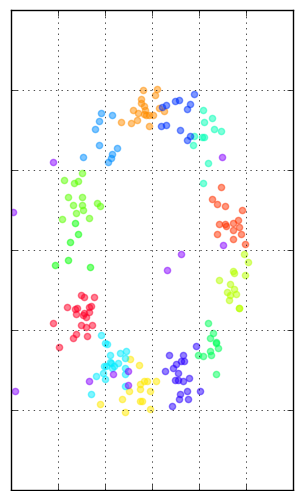}
    \end{subfigure}%
    \begin{subfigure}{0.125\textwidth}
        \includegraphics[width=\textwidth]{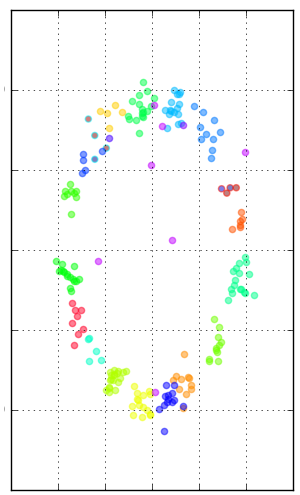}
    \end{subfigure}%
    \begin{subfigure}{0.125\textwidth}
        \includegraphics[width=\textwidth]{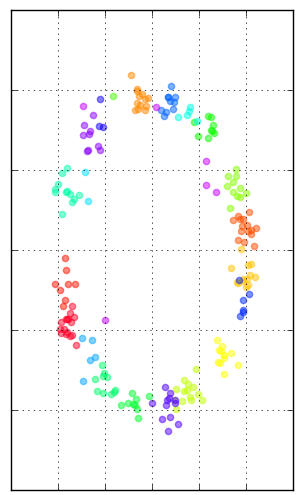}
    \end{subfigure}
    \subcaption{%
    Black dots represent observed samples from a noisy ring with decreasing noise level.
    Colored dots represent samples from CrossCat's posterior predictive after two minutes of analysis.
    The color of a point indicates its latent cluster assignment from CrossCat's inner Dirichlet process mixture.
    This panel illustrates a phenomenon known as the Bayes Occam's razor.
    At higher noise levels (left side plots) there is less evidence for patterns in the data, so the posterior prefers a less complex model with a small number of large clusters.
    At lower noise levels (right side plots) there is more evidence for the functional relationship, so the posterior prefers a more complex model with a large number of small clusters, which is required to emulate the ring.}
    \label{fig:crosscat-emulate-ring-simulate}
\end{subfigure}

\vspace{.25cm}

\begin{subfigure}{\textwidth}
    \begin{subfigure}{0.5\textwidth}
        \includegraphics[width=\textwidth]{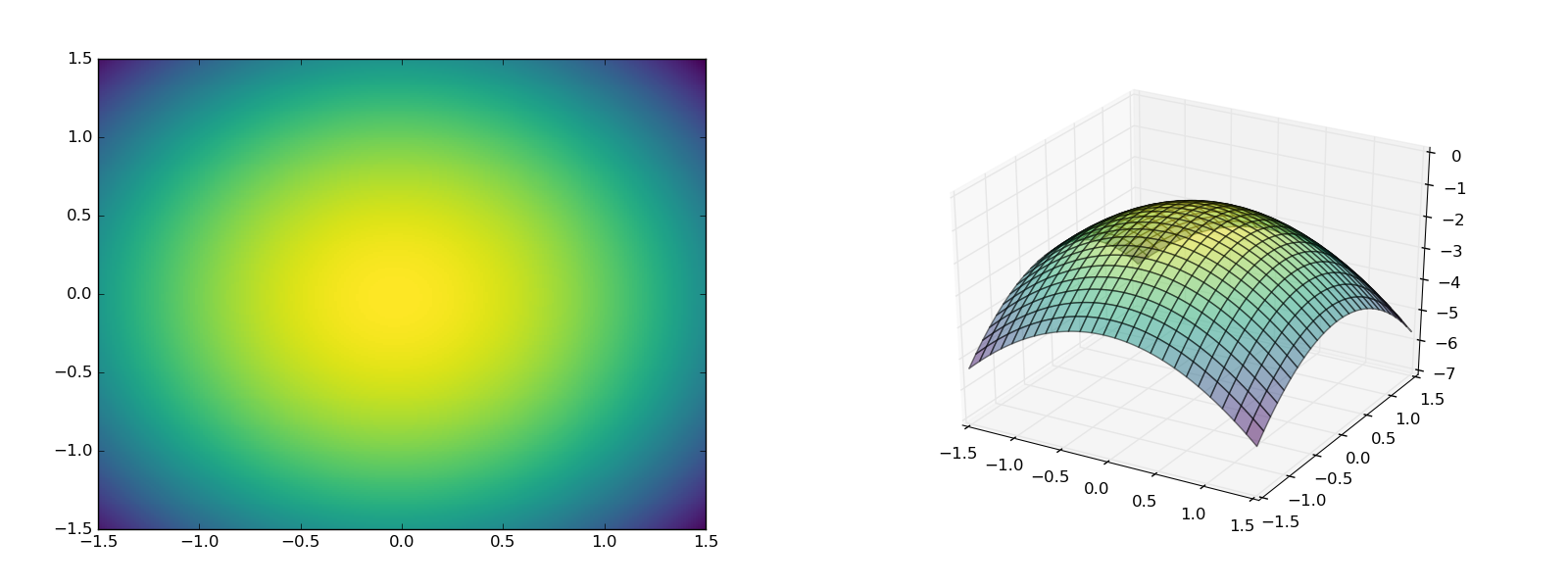}
    \end{subfigure}%
    \begin{subfigure}{0.5\textwidth}
        \includegraphics[width=\textwidth]{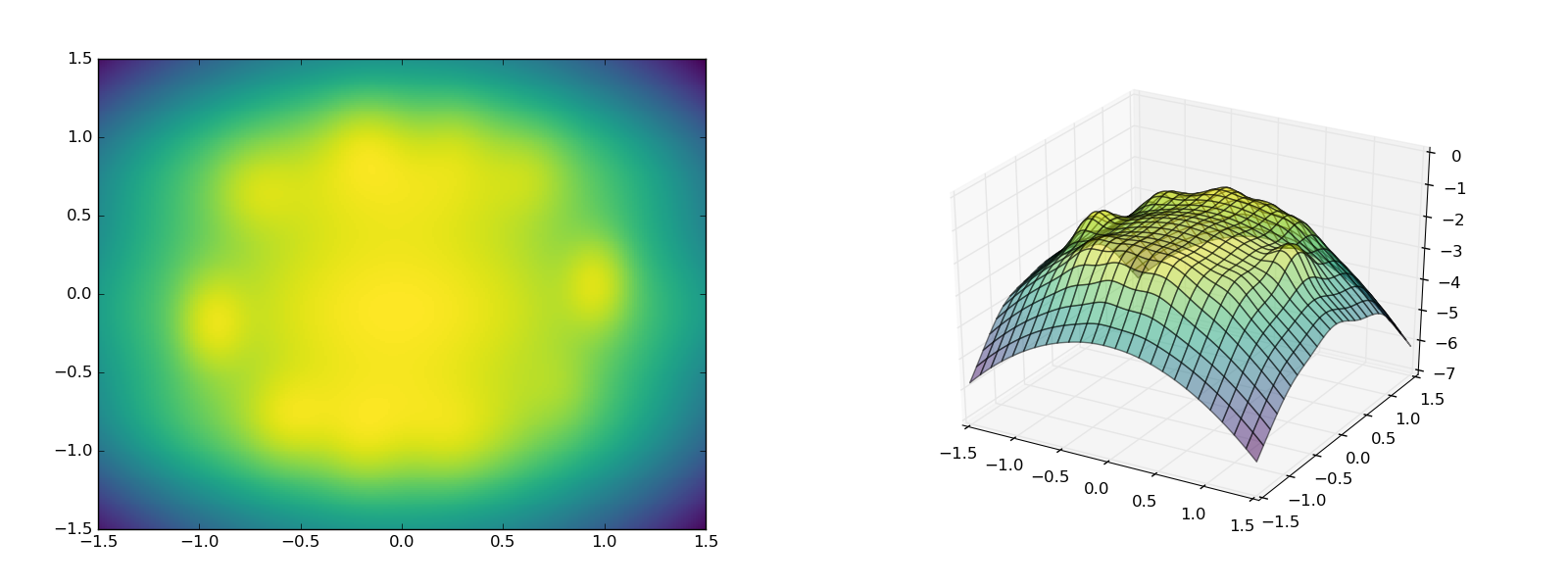}
    \end{subfigure}
    \begin{subfigure}{0.5\textwidth}
        \includegraphics[width=\textwidth]{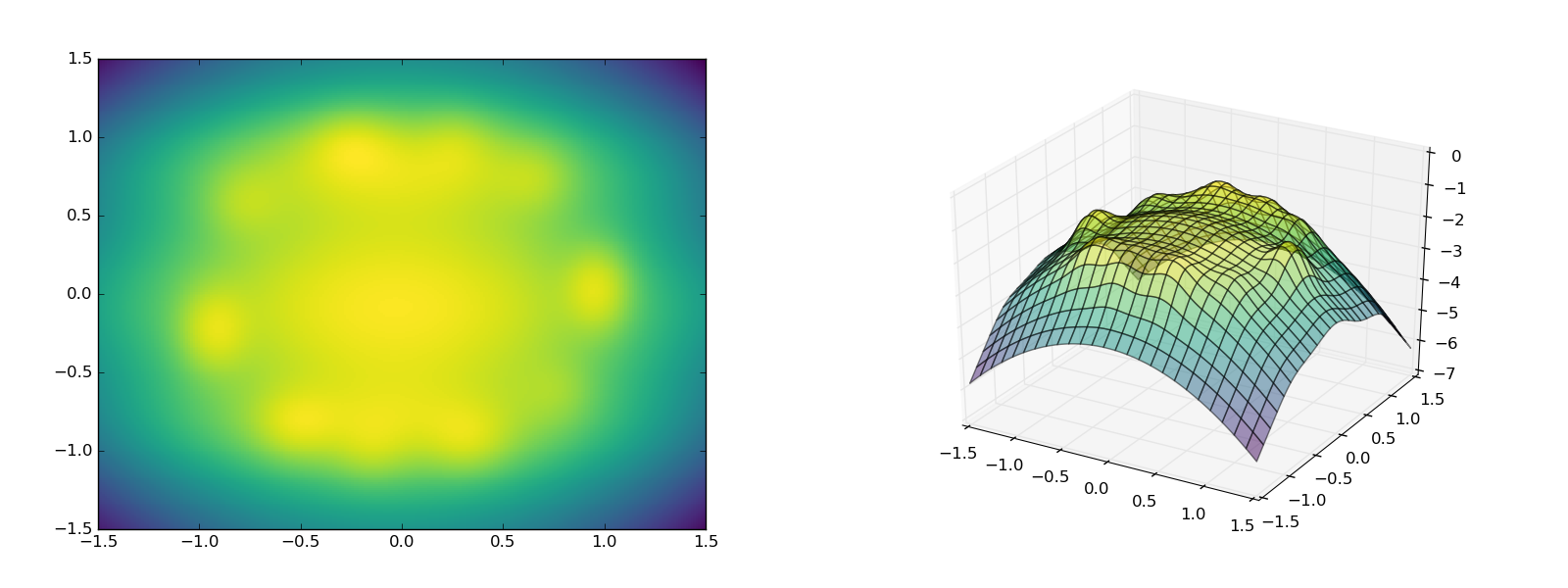}
    \end{subfigure}%
    \begin{subfigure}{0.5\textwidth}
        \includegraphics[width=\textwidth]{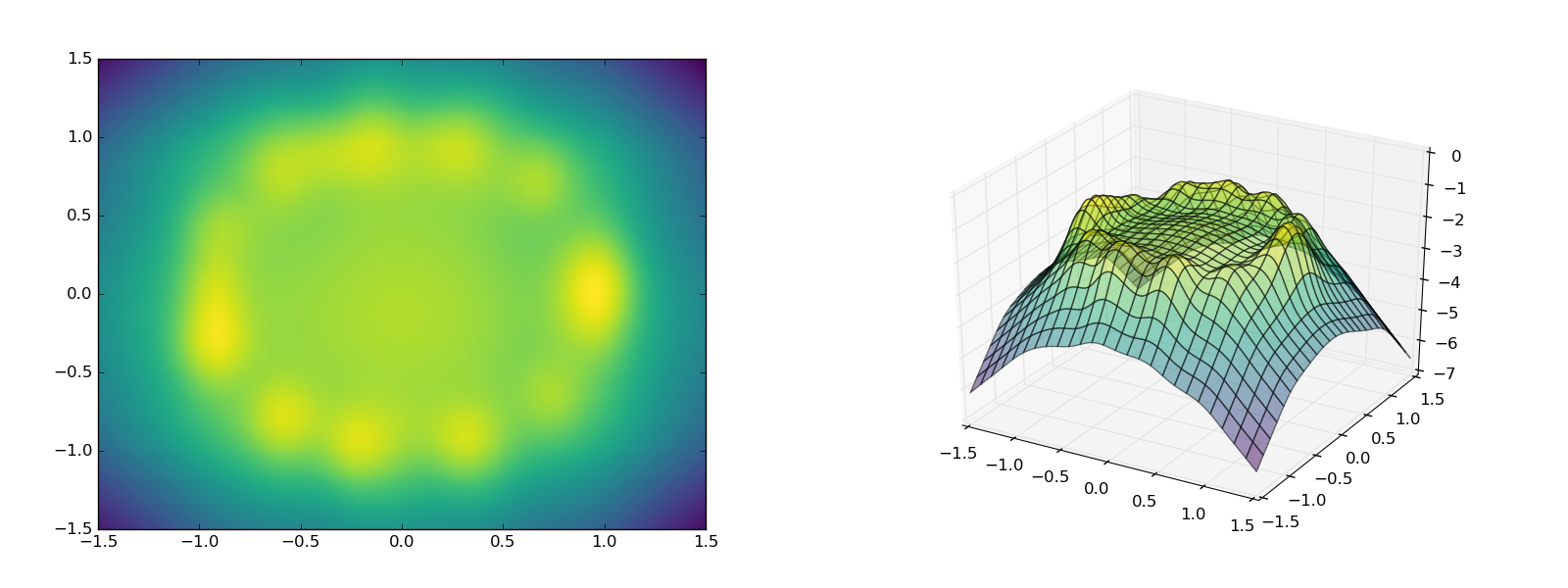}
    \end{subfigure}
    \begin{subfigure}{0.5\textwidth}
        \includegraphics[width=\textwidth]{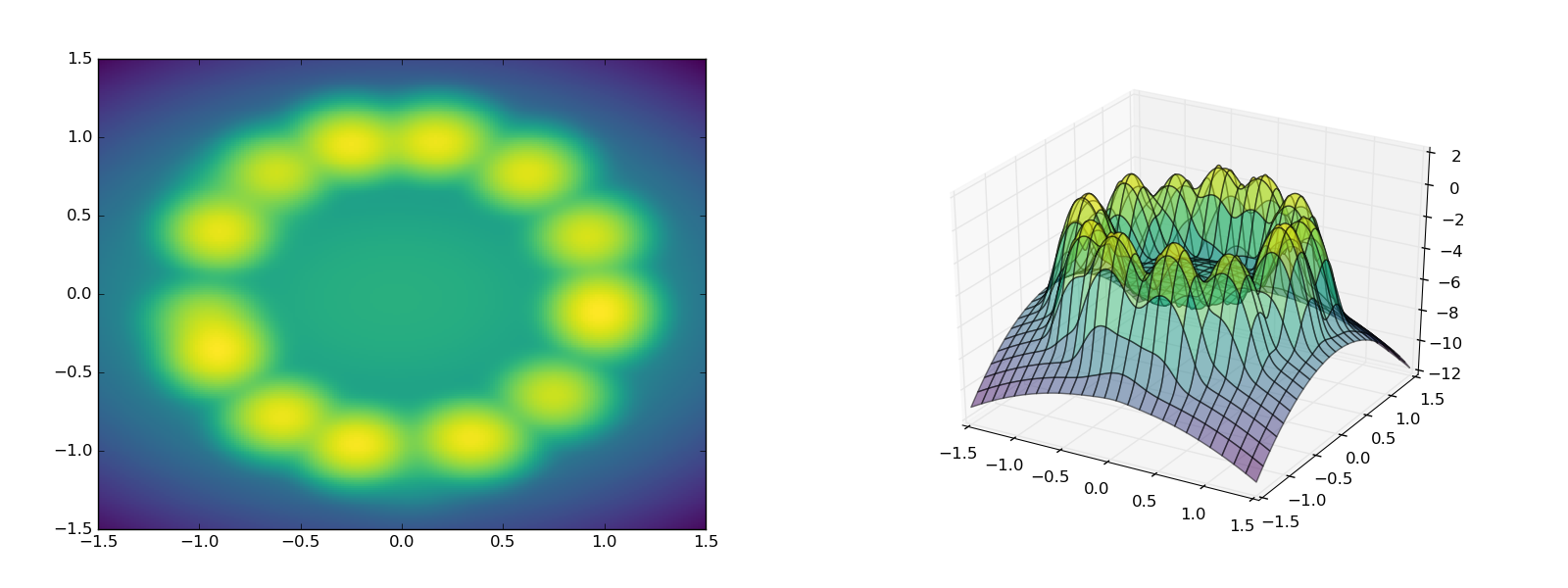}
    \end{subfigure}%
    \begin{subfigure}{0.5\textwidth}
        \includegraphics[width=\textwidth]{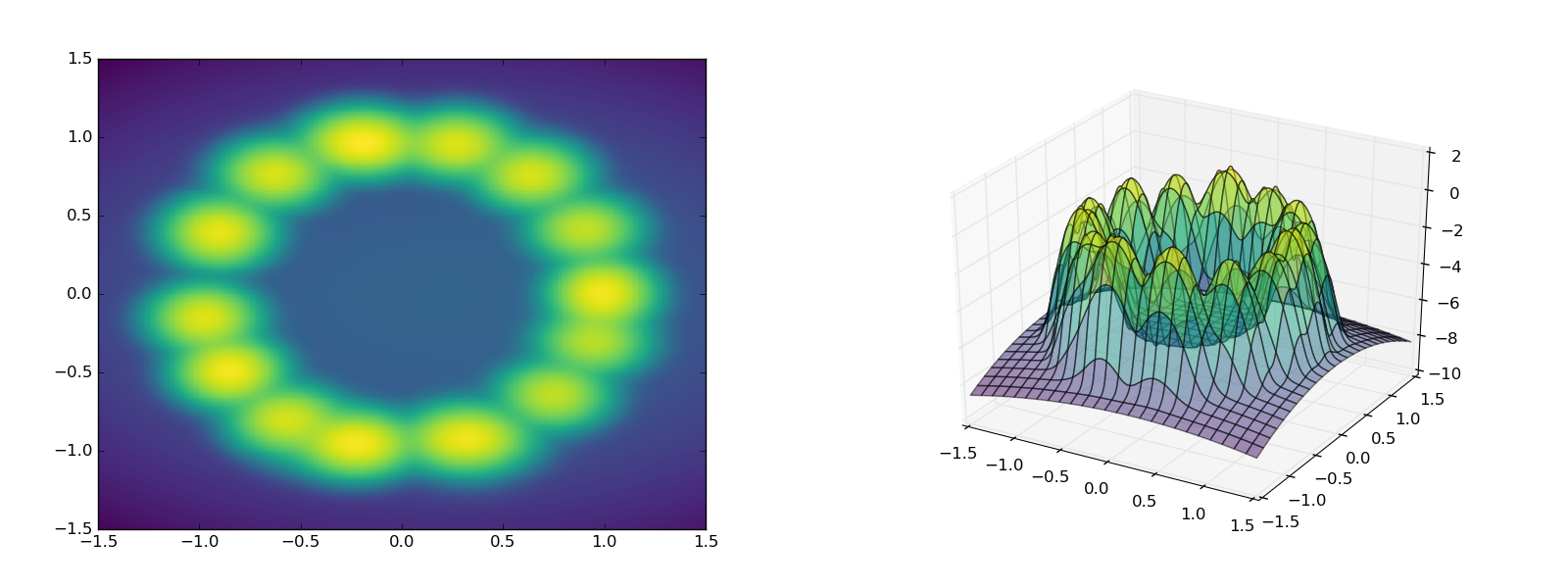}
    \end{subfigure}%
    \subcaption{%
    The heatmaps show the evolution of CrossCat's posterior predictive density with increasing number of inference transitions, given a ring with fixed noise level (sixth ring from right in panel \subref{fig:crosscat-emulate-ring-simulate}).
    Brighter shades of green indicate greater density mass in the region.
    The surface plots to the right of each heatmap show the same density, projected in three dimensions.
    During early stages of inference, the density surface is unimodal and appears as a cloud in the 2D plane.
    Modalities and patterns in the data are captured with increasing inference, as the Markov chain centers on regions of high posterior mass in CrossCat's latent state.}
    \label{fig:crosscat-emulate-ring-logpdf}
\end{subfigure}

\bcaption{%
Using \textnormal{\texttt{simulate}} and \textnormal{\texttt{logpdf}} to study CrossCat's emulation of a noisy ring.}{}
\label{fig:crosscat-emulate}
\end{figure}

\clearpage


\subsection{Ensemble classifiers and regressors}
\label{subsec:implementation-ensemble}

In this section, we describe how to construct CGPMs for a class of ensemble- based classifiers and regressors that are common in machine learning.
These CGPMs are not typically described by a graphical model (Section~\ref{subsec:cgpms-statistical}) yet are still able to satisfy the CGPM interface by implementing \texttt{simulate} and \texttt{logpdf}.
For each member $r$, we assume the CGPM $\G$ generates a single output variable $x_r$, and requires as input a feature vector $\y_r = (y_{[r,1]},\dots,y_{[r,I]})$.
In an ensemble method, $\G$ carries a set of learners $\set{L_1,\dots,L_K}$, where each learner $L_k$ returns a point prediction of $x_r$ given $\y_r$ denoted $L_k(\y_r)$.
As a simple example, $\G$ may represent a random forest, and each learner $L_i$ a constituent decision tree.
For \texttt{infer}, $\G$ may construct the ensemble of learners given measurements $\D$ using any meta-learning algorithm such Boosting \citep{freund1995}, Bagging \citep{breiman1996} or others.

\subsubsection{Classification}
\label{subsubsec:implementation-ensemble-classification}

Let $\set{1,\dots,S}$ denote the set of possible values for the output variable $x_r$ (this specification is consistent with a \texttt{BINARY} or \texttt{NOMINAL} statistical data type from Table~\ref{tab:stattypes} in Section~\ref{subsec:bayesdb-mml}).
Given an input $\y_r$, the simplest strategy to define a probability for the event $[x_r=s]$ is to compute the proportion of learners in the ensemble who predict $[L_k(\y_r) = s]$.
This baseline strategy guarantees that the discrete probabilities sum to 1; however, it suffers from degeneracy in that the \texttt{simulate} and \texttt{logpdf} are undefined when $D$ is empty.
To address this issue, we introduce a smoothing parameter $\alpha$.
With probability $\alpha$, the output $x_r$ is uniform over the $S$ symbols, and with probability $(1-\alpha)$, it is an aggregate of outputs from the learners,
\begin{align}
\pG(x_r|\y_r,\D) =
    (1-\alpha)\left(
        \frac{1}{K}
        \sum_{s=1}^{S} \left(
            \mathbb{I}[x_r=s]\sum_{k=1}^K\left(\mathbb{I}[L_k(\y_r)=s]\right)
        \right)
    \right) +
    \alpha\left(\frac{1}{S}\right).
\label{eq:ensemble-classification-logpdf}
\end{align}
In practice, a prior is placed on the smoothing parameter $\alpha\sim\textsc{Uniform}([0,1])$, which is transitioned by gridded Gibbs sampling \citep{ritter1992} over the prediction likelihood on the measurement set.
When the distribution of $x_r$ given $\y_r$ is in the hypothesis space of the learners, we expect that $\lim_{n\to\infty}\pG(\alpha|\G,\D_n) =0$.
Both \texttt{simulate} and \texttt{logpdf} can be implemented directly from \eqref{eq:ensemble-classification-logpdf}.

\subsubsection{Regression}
\label{subsubsec:implementation-ensemble-regression}

In the regression setting, the predictions $\set{L_k(\y_r)}$ returned by each learner are real-valued, and so the discrete aggregation strategy from \eqref{eq:ensemble-classification-logpdf} does not lead to a well-defined implementation of \texttt{logpdf}.
Instead, for an input vector $\y_r$ the ensemble-based regression CGPM $\G$ first computes the set of predictions $\set{L_1(\y_r),\dots{L_K(\y_r)}}$, and then $\texttt{incorporate}$s them into a primitive univariate CGPM compatible with the statistical type of the output variable, such as a \texttt{NORMAL} for \texttt{NUMERICAL}, or \texttt{LOGNORMAL} for \texttt{MAGNITUDE}.
This strategy fits a statistical type appropriate noise model based on the variability of responses from the learners, which relates to how noisy the regression is.
implementations of \texttt{logpdf} and \texttt{simulate} are directly inherited from the constructed primitive CGPM.


\subsection{Factor analysis \& probabilistic PCA}
\label{subsec:implementation-pca}

Our development of factor analysis closely follows \citep[Chatper 12]{murphy2012}; we extend the exposition to describe implementations of \texttt{simulate} and \texttt{logpdf} for arbitrary patterns of latent and observable variables. Factor analysis is a continuous latent variable model where the vector of output variables $\x_r = (x_{[r,1]},\dots,x_{[r,D]})$ is a noisy linear combination of a set of $L$ basis vectors $\set{\w_1,\dots,\w_L}$,
\begin{align}
\x_r = \bm\mu + \w_1z_{[r,1]} + \w_2z_{[r,2]} + \dots + \w_Lz_{[r,L]}
        + \bm\epsilon
    && \bm\epsilon \sim^\G
        \textsc{Normal}(\bm{0}, \textrm{diag}(\psi_1,\dots, \psi_D)).
\label{eq:factor-analysis-linear-combination}
\end{align}
Each basis vector $\w_l$ is a $D$-dimensional vector and the dimension of the latent space $L$ (a hyperparameter) is less than $D$.
The member latents $\z_r \in \mathbb{R}^L$ are known as factor scores, and they represent a low-dimensional projection of $\x_r$.
The global latents are the bases $\mathbf{W}=[\w_1\dots\w_L]$, covariance matrix $\bm\Psi$ of the noise $\bm\epsilon$, and mean vector $\bm\mu$ of $\x_r$.
To specify a generative model, the member-specific latent variables are given a prior $\z_r \sim \textsc{Normal}(\bm0, \mathbf{I})$.
Combining this prior with \eqref{eq:factor-analysis-linear-combination} the joint distribution over the latent and observable variables is
\begin{align}
    \s_r = \begin{pmatrix}
        \z_r\\
        \x_r
    \end{pmatrix}
    \sim^\G
    \textsc{Normal} \left(
        \bm{m} = \begin{pmatrix}
            \bm{0}\\
            \bm\mu
        \end{pmatrix},
    \bm\Sigma = \begin{pmatrix}
        \mathbf{I}_{L\times{L}}
            & \mathbf{W}^\top_{L\times{D}}\\
        \mathbf{W}^\top_{D\times{L}}
            & \left(\mathbf{W}\mathbf{W}^\top+\bm\Psi\right)_{D\times{D}}\\
        \end{pmatrix}
    \right),
\label{eq:factor-analysis-joint-distribution}
\end{align}
where we have defined the joint vector $\s_r = (\z_r,\x_r) \in \mathbb{R}^{D+L}$.
The CGPM $\G$ implementing factor analysis exposes the member-specific latent variables as output variables.
The multivariate normal \eqref{eq:factor-analysis-joint-distribution} provides the ingredients for \texttt{simulate} and \texttt{logpdf} on any pattern of latent and observable variables with \texttt{query} $\s_{[r,Q]}$ and \texttt{evidence} $\s_{[r,E]}$.
To arrive at the target distribution, the Bayes theorem for Gaussians \citep{bishop2006} is invoked in a two-step process.%
\begin{align*}
& \textrm{\underline{Marginalize}}
    &  \s_{[r,Q\cup{E}]} \sim^\G
    \textsc{Normal} \left(
        \begin{pmatrix}
            \bm\mu_Q\\
            \bm\mu_E
        \end{pmatrix},
    \begin{pmatrix}
        \bm\Sigma_{Q} & \bm\Sigma_{Q\cup{E}} \\
        \bm\Sigma_{Q\cup{E}}^\top & \bm\Sigma_{E} \\
    \end{pmatrix}
    \right) \\
& \textrm{\underline{Condition}}
    &   \s_{[r,Q]} | \s_{[r,E]} \sim^\G
    \textsc{Normal} \left(
        \bm\mu_Q +
            \bm\Sigma_{Q\cup{E}} \bm\Sigma_{E}^{-1} (\s_{[r,E]}-\bm\mu_E),
        \bm\Sigma_{Q} -
            \bm\Sigma_{Q\cup{E}} \bm\Sigma_{E}^{-1} \bm\Sigma_{Q\cup{E}}^\top
    \right)
\end{align*}

Our implementation of \texttt{infer} uses expectation maximization for factor analysis \citep{ghahramani1997}; an alternative approach is posterior inference in the Bayesian setting \citep{press1997}.
Finally, probabilistic principal component analysis \citep{tipping1999} is recovered when covariance of $\bm\epsilon$ is further constrained to satisfy $\psi_1=\dots=\psi_D$.


\begin{figure}[h]
\begin{subfigure}[b]{.5\textwidth}
\begin{Verbatim}[fontsize=\scriptsize]
%mml CREATE TABLE iris FROM `iris.csv';
%mml CREATE POPULATION p FOR iris (GUESS (*));
%mml CREATE METAMODEL m FOR p (
....    OVERRIDE GENERATIVE MODEL FOR
....        sepal_length, sepal_width,
....        petal_length, petal_width
....    AND EXPOSE
....        flower_pc1 NUMERICAL,
....        flower_pc2 NUMERICAL
....    USING probabilistic_pca(L=2));
%mml INITIALIZE 1 MODEL FOR m;
%mml ANALYZE m FOR 10 ITERATION;
%bql .scatter
....    INFER EXPLICIT
....        PREDICT flower_pc1 USING 10 SAMPLES,
....        PREDICT flower_pc2 USING 10 SAMPLES,
...         flower_name
....    FROM p;
\end{Verbatim}
\end{subfigure}%
\begin{subfigure}[b]{.5\textwidth}
    \centering
    \includegraphics[width=.825\textwidth]{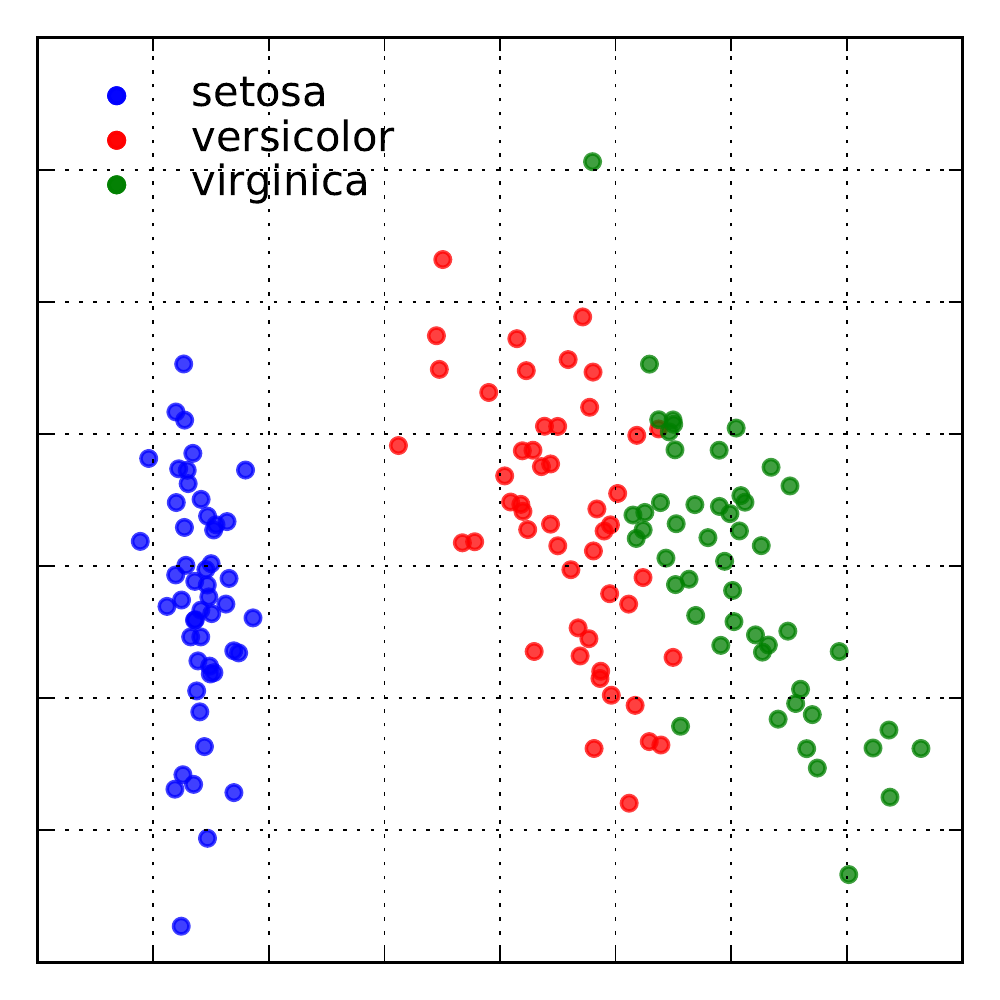}
\end{subfigure}%
\bcaption{%
Low dimensional projection of flowers in the iris dataset using the probabilistic PCA CGPM.}{%
The two latent principal components scores are exposed as queryable outputs in BQL.}
\label{fig:pca-iris}
\end{figure}

\clearpage


\subsection{Parametric mixture of experts}
\label{subsec:implementation-experts}

\begin{figure}[ht]
\centering
    \begin{subfigure}[b]{.5\textwidth}
        \includegraphics[width=\textwidth]{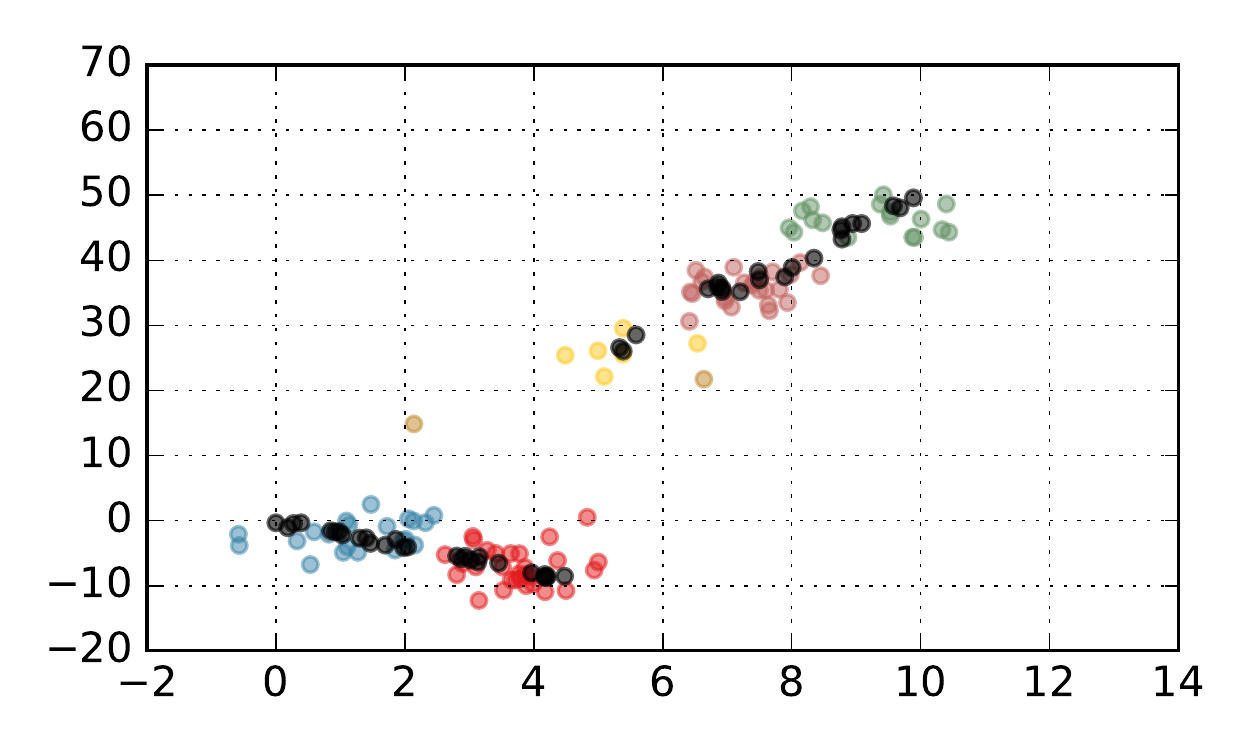}
        \subcaption{CrossCat}
        \label{fig:moe-crosscat}
    \end{subfigure}%
    \begin{subfigure}[b]{.5\textwidth}
        \includegraphics[width=\textwidth]{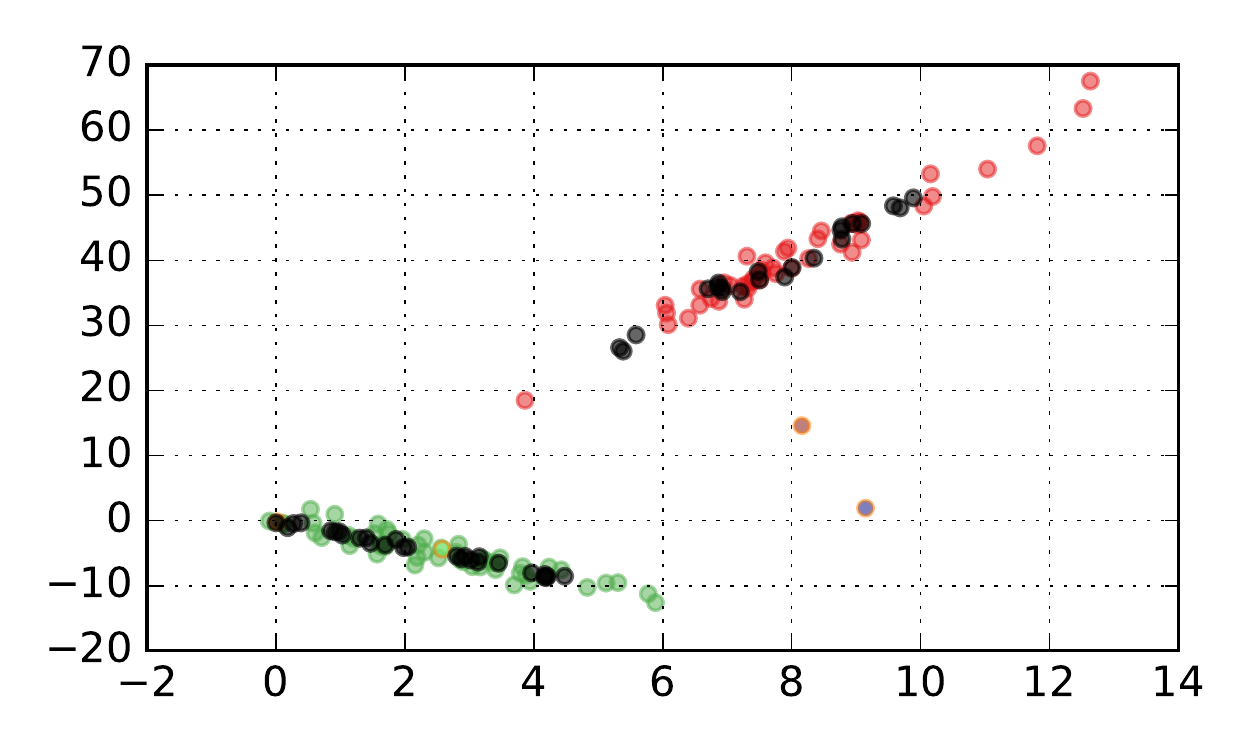}
        \subcaption{Mixture of linear regression experts}
        \label{fig:moe-linear}
    \end{subfigure}
    \bcaption{%
    Posterior samples from CrossCat and mixture of experts given a piecewise continuous linear function.}{%
    Observed data points are shown in black, and posterior samples are shown in color, which represents a latent cluster assignment internal to each CGPM. \textbf{\subref{fig:moe-crosscat}} CrossCat emulates the curve using a mixture of axis-aligned Gaussians, requiring a larger number of small, noisy clusters.
    \textbf{\subref{fig:moe-linear}} Mixture of linear regression experts identifies the two linear regimes and is able to interpolate well (red dots in top curve).
    The two orange datapoints that appear as outliers are samples from a ``singleton'' cluster, since the gating function is implemented using a Dirichlet process mixture.}
    \label{fig:moe-crosscat-linear}
\end{figure}

The mixture of experts \citep{jacobs1991} is a regression model for data which exhibit highly non-linear characteristics, such as heteroskedastic noise and piecewise continuous patterns.
Let $\G$ be a CGPM which generates output variables $\x_r =(x_{[r,1]},\dots,x_{[r,T]})$ given input variables $\y_r$, using mixtures of local parametric mixtures.
The member latent variable $\z_r = (z_r)$ takes values in $[K]$ (possibly unbounded) which induces a Naive Bayes factorization over the outputs
\begin{align}
\pG(\x_{[r,Q]}|\y_r, \btheta) =
    \sum_{k=1}^K
        \left(
            \prod_{t=1}^T\pG(x_{[r,t]}|\y_r,\bgamma_{[q,z_r]})
            \pG(z_r=k|\y_r,\btheta)
        \right),
  \label{eq:moe-naive-bayes}
\end{align}
where $\bgamma_{[q,k]}$ are the regression parameters for variable $x_{[r,t]}$ when $z_r=k$.
While \eqref{eq:moe-naive-bayes} looks similar to the Naive Bayes factorization \eqref{eq:crosscat-naive-bayes} from CrossCat, they differ in important ways.
In CrossCat, the variables $x_{[r,t]}$ are sampled from primitive univariate CGPMs, while in the mixture of experts they are sampled from a discriminative CGPM conditioned on $\y_r$.
The term $\pG(x_{[r,t]}|\y_r,\bgamma_{[q,z_r]})$ may be any generalized linear model for the correct statistical data type (such as a Gaussian linear regression for \texttt{NUMERICAL}, logistic regression for \texttt{NOMINAL}, or Poisson regression for \texttt{COUNTS}).
Second, the mixture of experts has a ``gating function'' for $\pG(z_r=k|\y_r,\btheta)$ which is also conditioned on $\y_r$ and may be a general function such as a softmax or even a Dirichlet process mixture \citep{hannah2011}.
In, CrossCat the member latents $z_{[r,B]}$ are necessarily given a CRP prior in each block.
We leave out implementations of \texttt{simulate} and \texttt{logpdf}, and refer to Figure~\ref{fig:moe-crosscat-linear} for a comparison of posterior samples from CrossCat and mixture of experts given data from a piecewise continuous function.


\clearpage


\subsection{Generative nearest neighbors}
\label{subsec:implementation-knn}

In this section, we present a compositional generative population model which implements \texttt{simulate} and \texttt{logpdf} by building ad-hoc statistical models on a per-query basis.
The method is a simple extension of K Nearest Neighbors to generative modeling.

Let $\G$ be a generative nearest neighbor CGPM, and $\x_{[r,Q]}$ and $\x_{[r,E]}$ denote the query and evidence for a \texttt{simulate} or \texttt{logpdf} query.
The method first finds the $K$ nearest neighbors to $r$ in dataset $\D$, based on the values of the evidence variables $\x_{[r,E]}$.
Let $\mathcal{N}$ denote the top $K$ neighbors, whose generic member is denoted $\x_k \in \mathcal{N}$.
Within $\mathcal{N}$, we assume the query variables $Q$ are independent, and learn a CGPM $\G = \set{\G_{[q]}: q \in Q}$ which is a product of primitive univariate CGPMs $\G_{q}$ (based on the appropriate statistical data type of each variable $q$ from Table~\ref{tab:stattypes}).
The measurements $\set{x_{[k,q]} k \in \mathcal{N}}$ are used to learn the primitive CGPM for $q$ in the neighborhood.
This procedure is summarized in Algorithm~\ref{alg:knn-build-local}.
Implementations of \texttt{simulate} and \texttt{logpdf} follow directly from the product CGPM, as summarized in Algorithms~\ref{alg:knn-simulate} and \ref{alg:knn-logpdf}.
Figure~\ref{fig:knn} illustrates how the behavior of \texttt{simulate} on a synthetic x-cross varies with the neighborhood size parameter K.

It should be noted that building independent models in the neighborhood will result in very poor performance when the query variables remain highly correlated even when conditioned on the evidence.
Our baseline approach can be modified to capture the dependence between the query variables by instead building one independent CGPM around the local neighborhood of each neighbor $k \in \mathcal{N}$, rather than one independent CGPM for the entire neighborhood. These improvements are left for future work.


\begin{figure}[ht]
\centering
\small
\begin{subalgorithms}
    \captionof{algorithm}{%
        \texttt{simulate} for generative nearest neighbors CGPM.}
    \label{alg:knn-simulate}
    \begin{algorithmic}[1]
        \State $\x_{[r,Q]} \gets \varnothing$
            \Comment{initialize empty sample}
        \State $(\G_q: q \in Q) \gets
            \textsc{Build-Local-Cgpms}\: (\x_{[r,E]})$
            \Comment{retrieve the local parametric CGPMs}
        \For{$q\in Q$}
            \Comment{for each query variable $q$}
            \State $x_{[r,q]} \gets $
                \texttt{simulate}($\G_{[j,q]}, r, \set{q}, \varnothing$)
                \Comment{sample from the primitive CGPM}
        \EndFor
        \State \Return $\x_{[r,Q]}$
            \Comment{overall sample of query variables}
    \end{algorithmic}

    \captionof{algorithm}{%
        \texttt{logpdf} for generative nearest neighbors CGPM.}
    \label{alg:knn-logpdf}
    \begin{algorithmic}[1]
        \State $(\G_q: q \in Q) \gets
            \textsc{Build-Local-Cgpms}\: (\x_{[r,E]})$
            \Comment{retrieve the local parametric CGPMs}
            \For{$q\in Q$}
                \Comment{for each query variable $q$}
                \State $\log{w_q} \gets$
                    \texttt{logpdf}($\G_q, r, x_{[r,q]}, \varnothing$)
                    \Comment{compute the density of $q$}
            \EndFor
        \State \Return{$\sum_{q\in Q}\log{w_q}$}
            \Comment{overall density estimate}
    \end{algorithmic}

    \captionof{algorithm}{%
        Building local parametric models in the generative nearest neighbor
        CGPM.}
    \label{alg:knn-build-local}
    \begin{algorithmic}[1]
        \Function{Build-Local-Cgpms}{} $(\x_{[r,E]})$
        \State $\D_{E} \gets \set{\x_{[r',E]}: r' \in \mathcal{D}}$
            \Comment{marginalize by exclusion from neighbor search}
        \State $\mathcal{N} \gets \textsc{Nearest-Neighbors}
            (K, \D_{E}, \x_{[r,E]})$
            \Comment{find neighbors of $r$}
        \For{$q \in Q$}
            \Comment{for each query variable $q$}
            \State $\G_{q} \gets \textsc{Primitive-Univariate-Cgpm}$
                \Comment{initialize a primitive CGPM}
                \For{$k \in \mathcal{N}$}:
                    \Comment{for each neighbor}
                    \State $\G_{q} \gets
                        \texttt{incorporate}(\G_{q}, k, x_{[k,q]})$
                        \Comment{incorporate into primitive CGPM}
                \EndFor
            \State $\G_{q} \gets
                \texttt{infer}(\G_{q}, \mathcal{T}_{\textrm{ML}})$
                \Comment{transition the primitive CGPM}
        \EndFor
        \State \Return $(\G_q: q \in Q)$
            \Comment{collection of primitive CGPMs}
        \EndFunction
    \end{algorithmic}
\end{subalgorithms}
\end{figure}

\begin{figure}[ht]
\begin{subfigure}{\textwidth}
    \begin{subfigure}[b]{0.325\textwidth}
        \includegraphics[width=\textwidth]
            {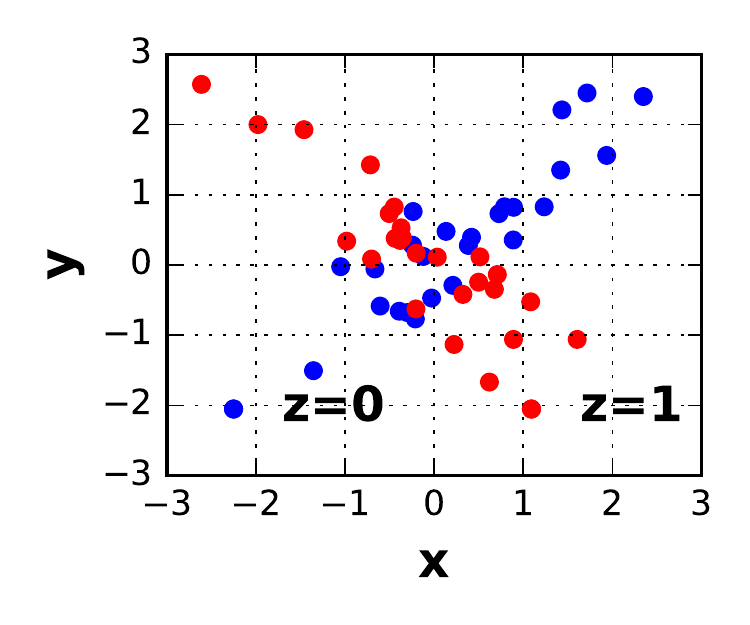}
        \caption{Observed data}
        \label{fig:knn-synthetic}
    \end{subfigure}%
    \begin{subfigure}[b]{0.675\textwidth}
        \includegraphics[width=\textwidth]{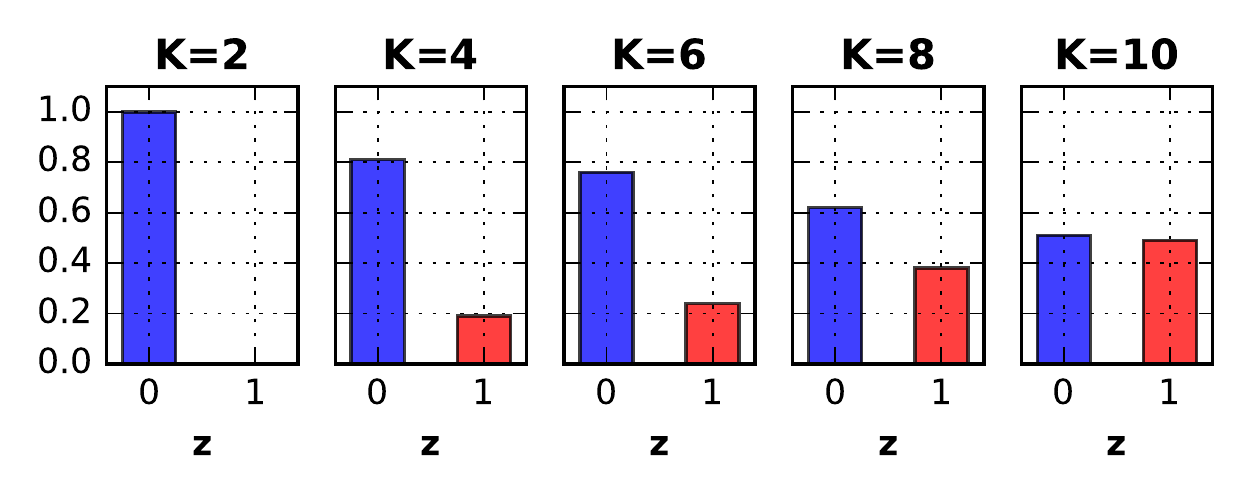}
        \caption{Samples of \texttt{z GIVEN x=0.5, y=0.5} for various neighborhood sizes}
        \label{fig:knn-posterior}
    \end{subfigure}%
    \begin{Verbatim}[gobble=4, frame=lines]
    %mml CREATE METAMDOEL xcross_m WITH BASELINE gknn(K=?) FOR xcross;
    %bql .scatter SIMULATE z FROM xcross_m GIVEN x=0.5, y=0.5 LIMIT 50;
    \end{Verbatim}
\end{subfigure}
\bcaption{%
Posterior samples from the generative nearest neighbors CGPM given an x-cross for varying values of neighbors K.}{%
\textbf{\subref{fig:knn-synthetic}} Samples from the synthetic x-cross data generator.
It produces three variables: \texttt{x} and \texttt{y} are real-valued and are scattered in the 2D plane, and \texttt{z} is a binary variable indicating the functional regime.
\textbf{\subref{fig:knn-posterior}}
For small neighborhoods (\texttt{K}=2, \texttt{K}=4), most members of the neighborhood satisfy \texttt{z=0}, as reflected by the sharp posterior distribution of \texttt{z} at 0.
As the neighborhood size increases (\texttt{K}=8, \texttt{K}=10) they become noisy and include more members with \texttt{z=1}, smoothing out the posterior over \texttt{z} between 0 and 1.}
\label{fig:knn}
\end{figure}


\subsection{Multivariate kernel density estimation}
\label{subsec:implementation-kde}

In this section, we show how to express multivariate kernel density estimation with mixed data types, as developed by \citep{racine2004}, using CGPMs.
Similarly to ensemble methods (Section~\ref{subsec:implementation-ensemble}) this approach implements the CGPM interface without admitting a natural representation in terms of the graphical model in Figure~\ref{fig:cgpm-graphical}.
We extend the exposition of \citep{racine2004} to include algorithms for conditional sampling and density assessment.
Given measurements $\D$, the joint distribution over the variables of $\x_r$ is estimated non-parametrically
\begin{align}
\pG(\x_r|\D) = \frac{1}{|\D|} \sum_{r'\in \D} \mathcal{K}(\x_r|\bgamma)
    & = \frac{1}{|\D|} \sum_{r'\in \D}
            \left(\prod_{i\in[O]}
                \frac{1}{\gamma_i}
                K_i \left( x_{[r,i]},x_{[r',i]}|\gamma_i \right)
            \right).
\label{eq:kde}
\end{align}
$\mathcal{K}(\x_r|\bgamma)$ is a product kernel and $\bgamma$ is a global parameter containing the bandwidths for each kernel $K_i$.
Note that using a product kernel does not imply independence of elements $x_{[r,i]}$ and $x_{[r,j]}$ within a member.
Bandwidths are typically learned by cross-validation or maximum-likelihood.
For a \texttt{NOMINAL} statistical type with $S$ symbols the kernel is
\begin{align*}
K_q(x,x'|\gamma_q) = \left((1-\gamma_q)\mathbb{I}[x=x'] + \gamma_q/(S-1)\mathbb{I}[x\ne x']\right),
\end{align*}
from \citep{aitchison1976}. For a \texttt{NUMERICAL} statistical type the kernel is a standard second order Gaussian
\begin{align*}
K_q(x,x'|\gamma_q) = \left( \exp(-\frac{1}{2}((x-x')/\gamma)^2)/\sqrt{2\pi}\right).
\end{align*}
To implement \texttt{simulate} and \texttt{logpdf}, we first show how the product kernel \eqref{eq:kde} ensures marginalization is tractable,
\begin{flalign}
& \textrm{\underline{Marginalize}} & \notag \\
& \pG(\x_{[r,Q]}|\D)
    = \int_{\x_{[r,\backslash{Q}]}}
        \pG(\x_{[r,Q]},\x_{[r,\backslash{Q}]}) d\x_{[r,\backslash{Q}]}
    = \int_{\x_{[r,\backslash{Q}]}}
        \frac{1}{|\D|} \sum_{r'\in \D}
            \mathcal{K}(\x_r|\bgamma)
        d\x_{[r,\backslash{Q}]} & \notag \\
& = \int_{\x_{[r,\backslash{Q}]}}
    \left[
        \frac{1}{|\D|} \sum_{r'\in \D}
        \left(\prod_{i\in[O]}
            \frac{1}{\gamma_i}
            K_i\left(x_{[r,i]},x_{[r',i]}|\gamma_i\right)
        \right)
        d\x_{[r,\backslash{Q}]}
    \right] & \notag \\
& = \frac{1}{|\D|} \sum_{r'\in \D} \left(
        \int_{\x_{[r,\backslash{Q}]}}
        \left[
            \left(\prod_{q\in{Q}}
                \frac{1}{\gamma_q}
                K_q\left(x_{[r,q]},x_{[r',q]}|\gamma_q\right)
            \right)
            \left(\prod_{j\in{\backslash{Q}}}
                \frac{1}{\gamma_j}
                K_j\left(x_{[r,j]},x_{[r',j]}|\gamma_j\right)
            \right)
            d\x_{[r,\backslash{Q}]}
        \right]
        \right) & \notag \\
& = \frac{1}{|\D|} \sum_{r'\in \D}\left(
        \left(\prod_{q\in{Q}}
            \frac{1}{\gamma_q}
            K_q\left(x_{[r,q]},x_{[r',q]}|\gamma_q\right)
        \right)
        \underbrace{
            \int_{\x_{[r,\backslash{Q}]}}
            \left[
                \left(\prod_{j\in{\backslash{Q}}}
                    \frac{1}{\gamma_j}
                    K_j\left(x_{[r,j]},x_{[r',j]}|\gamma_j\right)
                \right)
                d\x_{[r,\backslash{Q}]}
            \right]
        }_{\textrm{density normalized to 1}}
        \right) & \notag \\
& = \frac{1}{|\D|} \sum_{r'\in \D}
        \left(\prod_{q\in{Q}}
            \frac{1}{\gamma_q}
            K_q\left(x_{[r,q]},x_{[r',q]}|\gamma_q\right)
        \right). & \label{eq:kde-marginal}
\end{flalign}

Conditioning is a direct application of the Bayes Rule, where the numerator and denominator are computed separately using \eqref{eq:kde-marginal}.
\begin{flalign}
& \textrm{\underline{Condition}}
& \pG(\x_{[r,Q]}|\x_{[r,E]}, D) =
    \frac{\pG(\x_{[r,Q]},\x_{[r,E]}|D)}{\pG(\x_{[r,E]}|D)}
\label{eq:kde-bayes}
\end{flalign}

Combining \eqref{eq:kde-marginal} and \eqref{eq:kde-bayes} provides an immediate algorithm for \texttt{logpdf}. To implement \texttt{simulate}, we begin by ignoring the normalizing constant in the denominator of \eqref{eq:kde-bayes} which is unnecessary for sampling. We then express the numerator suggestively,
\begin{align}
\pG(\x_{[r,Q]}|\x_{[r,E]},\D) \propto
     \sum_{r'\in \D}
        \left(
            \prod_{q\in Q}
            \frac{1}{\gamma_q}
            K_q\left(x_{[r,q]},x_{[r',q]}|\gamma_q\right)
        \underbrace
        {\prod_{e\in E} \frac{1}{\gamma_e}
        K_e\left(x_{[r,e]},x_{[r',e]}|\gamma_e\right) }_{\text{weight } w_r'}
    \right), \label{eq:kde-conditional}
\end{align}
In particular, the \texttt{simulate} algorithm first samples a member $r' \sim \textsc{Categorical}(\set{w_r':r\in \D})$, where the weight $w_r'$ is labeled in \eqref{eq:kde-conditional}. Next, it samples the query elements $x_{[r,q]}$ independently from the corresponding kernels curried at $r'$. Intuitively, the CGPM weights each member $r'$ in the population by how well its local kernel explains the evidence $\x_{[r,E]}$ known about $r$.


\subsection{Probabilistic programs in VentureScript}
\label{subsec:implementation-venturescript}

In this section, we show how to construct a composable generative population model directly in terms of its computational and statistical definitions from Section~\ref{sec:cgpms} by expressing it in the VentureScript probabilistic programming language.
For simplicity, this section assumes the CGPM satisfies a more refined conditional independence constraint than \eqref{eq:indep-gpm}, namely
\begin{align}
\exists q, q': (r,c) \ne (r',c') \implies
x_{[r,c]} \indep x_{[r',c']}
  \mid \set{\balpha, \btheta, z_{[r,q]}, z_{[r',q']}, \y_r, \y_r'}.
\label{eq:venturescript-independence}
\end{align}
In words, for every observation element $x_{[r,c]}$, there exists a latent variable $z_{[r,q]}$ that (in addition to $\btheta$) mediates all coupling with other variables in the population.
The member latent variables $\Z$ may still exhibit arbitrary dependencies within and among one another.
While not essential, this requirement simplifies exposition of the inference algorithms.
The approach for \texttt{simulate} and \texttt{logpdf} is based on approximate inference in tagged subparts of the Venture trace.%
\footnote{In Venture, every random choice may be in a \texttt{scope} which is divided into a set of \texttt{block}s. The CGPM places each member $r$ in its own \texttt{scope}, and each observable $x_{[r,i]}$ and latent $z_{[r,i]}$ element in a \texttt{block} within that \texttt{scope}.}
The CGPM carries a set of $K$ independent samples $\set{\btheta_k}_{k=1}^K$ from an approximate posterior $\pG(\btheta|\D)$.
These samples of global latent variables are assigned weights on a per-query basis.
Since VentureScript CGPMs are Bayesian, the target distribution for \texttt{simulate} and \texttt{logpdf} marginalizes over all internal state,
\begin{align}
&\pG(\x_{[r,Q]}|\x_{[r,E]},\D)
  = \int_{\btheta}
    \pG(\x_{[r,Q]}|\x_{[r,E]},\btheta,\D)\pG(\btheta|\x_{[r,E]},\D)d\btheta
    \label{eq:venture-no-weighting}
    \\
&= \int_{\btheta}p(\x_{[r,Q]}|\x_{[r,E]},\btheta,\D)
    \frac{\pG(\x_{[r,E]}|\btheta,\D)p(\btheta|\D)}{\pG(\x_{[r,E]}|\D,\G)}d\btheta
    \notag \\
&\approx \frac{1}{\sum_{k=1}^Kw_k}
    \sum_{k=1}^{K}\pG(\x_{[r,Q]}|\x_{[r,E]},\btheta_k,\D)w_k
    && \btheta_k \sim^\G |\D
    \label{eq:venture-weighting}.
\end{align}
The weight $w_k= \pG(\x_{[r,E]}|\btheta_k,\D)$ is the likelihood of the evidence
under $\btheta_k$. The weighting scheme
\eqref{eq:venture-weighting} is a computational trade-off circumventing the
requirement to run inference on population parameters $\btheta$ on a per-query
basis, i.e. when given new evidence $\x_{[r,E]}$ about $r$.%
\footnote{An alternative strategy is to compute a harmonic mean estimator
based directly on \eqref{eq:venture-no-weighting}.}

It suffices now to consider the target distribution under single sample $\btheta_k$:
\begin{align}
&\pG(\x_{[r,Q]}|\x_{[r,E]},\btheta_k,\D)
    = \int_{\z_r} \pG(\x_{[r,Q]}, \z_r | \x_{[r,E]}, \btheta_k, \D) d\z_r
    \label{eq:joint-local-posterior}  \\
&= \int_{\z_r}
    \left[
        \left( \prod_{q\in Q} \pG(x_{[r,q]} | \z_r, \btheta_k) \right)
        \pG(\z_r | \x_{[r,E]}, \btheta_k, \D) d\z_r
    \right]
    \label{eq:invoke-independence}  \\
&\approx
    \frac{1}{T}{\sum_{t=1}^T}
    \prod_{q\in Q} \pG(x_{[r,q]}|\z_{[t,r]},\btheta_k)
    && \z_{[t,r]} \sim^\G |\set{\x_{[r,E]},\btheta,\D}.
    \label{eq:conditional-posterior}
\end{align}
Eq~\eqref{eq:joint-local-posterior} suggests that \texttt{simulate} for can be implemented by sampling from the joint local posterior $\set{\x_{[r,Q]},\z_r|\x_{[r,E]},\btheta_k,\D}$, and returning only elements $\x_{[r,Q]}$.
Eq~\eqref{eq:conditional-posterior} shows that \texttt{logpdf} can be implemented by first sampling the member latents $\z_r$ from the local posterior.
By invoking conditional independence constraint \eqref{eq:venturescript-independence} in Eq \eqref{eq:invoke-independence}, the query $\x_{[r,Q]}$ factors into a product of density terms for each element $x_{[r,q]}$ which can be evaluated directly.
This description completes the algorithm for \texttt{simulate} and \texttt{logpdf} in trace $\btheta_k$, and is repeated for $\set{\btheta_1,\dots,\btheta_K}$.
The CGPM implements \texttt{simulate} by drawing a trace $j \sim \textsc{Categorical}(\set{w_1,\dots,w_K})$ and returning the sample $\x_{[r,Q]}$ from $\btheta_j$.
Similarly, \texttt{logpdf} is computed using the weighted Monte Carlo estimator \eqref{eq:venture-weighting}.
Algorithms \ref{alg:venturescript-simulate} and \ref{alg:venturescript-logpdf} illustrate implementations in a general probabilistic programming environment.


\begin{figure}[h]
\begin{subfigure}[b]{0.5\textwidth}
\begin{Verbatim}[fontsize=\footnotesize, commandchars=&\[\]]
%sql CREATE TABLE sin_t(x, y REAL);
%mml CREATE POPULATION sin_p FOR t WITH SCHEMA(
....   MODEL x, y AS NUMERICAL);

%mml CREATE METAMODEL sin_m FOR sin_p(
....   OVERRIDE MODEL FOR x USING
....     inline_venturescript(`
....       &fvtc[teal][() ~> {uniform(low: -4.71, high: 4.71)}]
....     ');
....   OVERRIDE MODEL FOR y GIVEN x USING
....     inline_venturescript(`
....      &fvtc[teal][(x) ~> {]
....        &fvtc[teal][if (cos(x) > 0) {]
....          &fvtc[teal][uniform(low: cos(x)-0.5, high: cos(x))}]
....        &fvtc[teal][else {]
....          &fvtc[teal][uniform(low: cos(x), high: cos(x)+0.5)}}]
....     ')
.... );
%mml ANALYZE 1 MODEL for sin_m;
%bql &fvtc[blue][.scatter SIMULATE x, y FROM sin_p LIMIT 100];
%bql &fvtc[red][.scatter SELECT x, 0.5 FROM(]
.... &fvtc[red][  SIMULATE x FROM sin_p GIVEN y=-0.75 LIMIT 50)];
\end{Verbatim}
\caption{}
\label{fig:noisy-sin-code}
\end{subfigure}%
\begin{subfigure}[b]{0.5\textwidth}
    \centering
    \begin{subfigure}{.735\textwidth}
        \includegraphics[width=\textwidth]{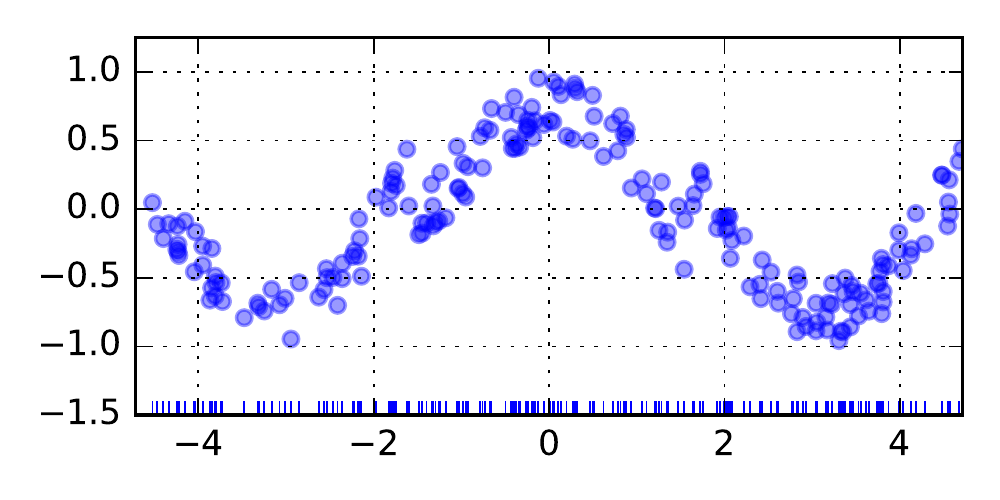}
    \end{subfigure}
    \begin{subfigure}{.735\textwidth}
        \includegraphics[width=\textwidth]{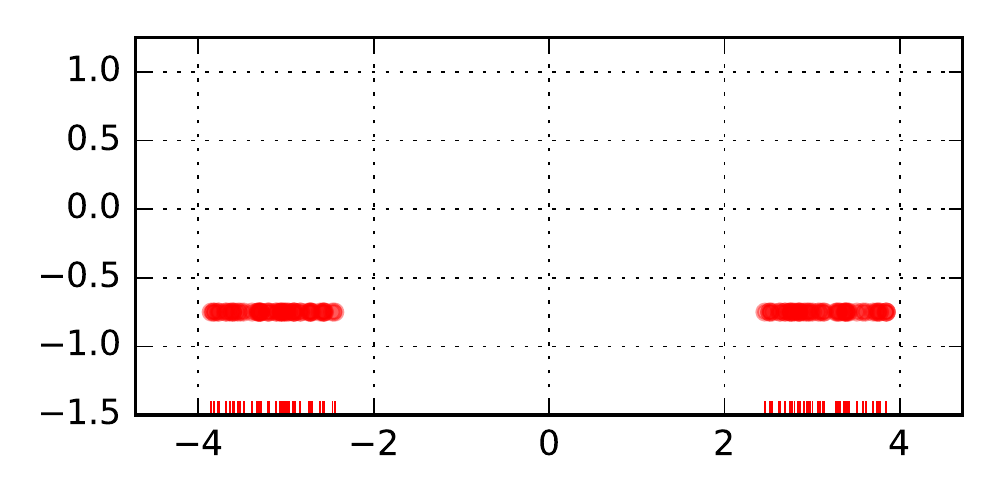}
    \end{subfigure}
    \begin{subfigure}{.735\textwidth}
        \includegraphics[width=\textwidth]{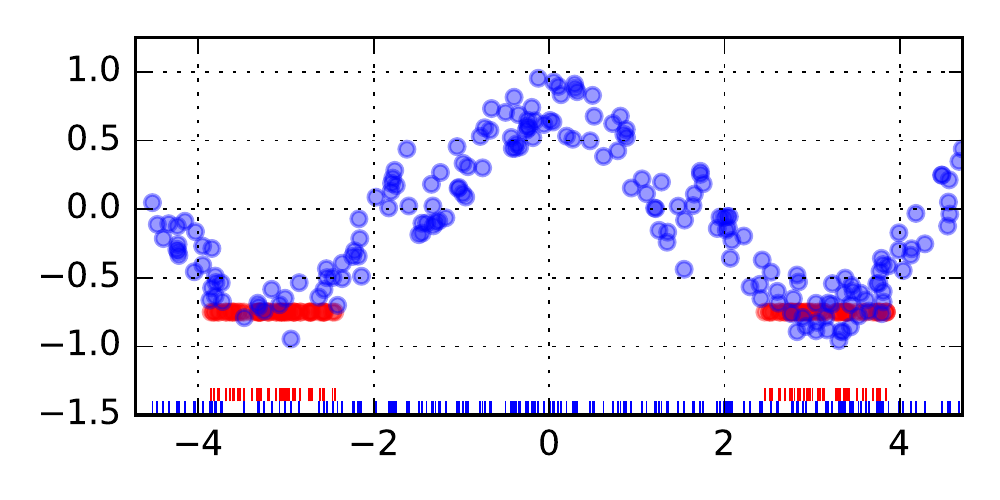}
    \end{subfigure}
\caption{}
\label{fig:noisy-sin-plots}
\end{subfigure}%

\bcaption{%
Composing VentureScript expressions by compiling them into CGPMs.}{%
\textbf{\subref{fig:noisy-sin-code}} Expressions in teal are lambda expressions, or anonymous functions, in VentureScript, which are compiled into CGPMs by the \texttt{inline\_venturescript} adapter.
Both forward simulation (blue query) and inversion (red query) of the joint generative model are achieved by Algorithm~\ref{alg:network-inference-simulate}.
This code is an instance of polyglot probabilistic programming; it includes expressions from two different languages interacting in a single program.
\textbf{\subref{fig:noisy-sin-plots}} The top plot shows samples of forward simulating \texttt{x} and \texttt{y} (blue query); the middle plot shows samples of \texttt{x GIVEN y=-0.75} (red query), successfully capturing the two posterior modes; the bottom plot shows an overlay.}
\label{fig:noisy-sin}
\end{figure}

\clearpage


\begin{table}[h]
\centering
  \begin{subtable}{0.5\textwidth}
      \begin{tabular*}{\textwidth}{@{}l@{\extracolsep{\fill}}l|}
        \toprule
        \textbf{Parameter} & \textbf{Symbol} \\ \midrule
        no. of trace instances & $K$ \\
        global latent variables in trace $k$ & $\btheta_k$ \\
        local latent variables in trace $k$ & $\Z_k$ \\
        observation set in trace $k$ & $\D_k$ \\
        input variable & $\y_r$ \\
        evidence set & $\x_{[r,E]}$ \\
        \bottomrule
      \end{tabular*}
  \end{subtable}%
  \begin{subtable}{0.5\textwidth}
      \begin{tabular*}{\textwidth}{ll}
        \toprule
        \textbf{Parameter} & \textbf{Symbol} \\ \midrule
        weight of trace $k$ & $w_k$ \\
        sample of $\z_{r}$ in trace $k$ & $\z_{[k,r]}$ \\
        sample of $\x_{[r,Q]}$ in trace $k$ & $\x_{[k,r,Q]}$ \\
        no. of internal Monte Carlo samples & $T$ \\
        $t$-th Monte Carlo sample of $\z_{[k,r]}$ & $\z_{[k,t,r]}$ \\
        weighted density estimate in trace $k$ & $q_k$ \\
        \bottomrule
      \end{tabular*}
  \end{subtable}
\bcaption{Parameters and symbols used in
  Algorithms~\ref{alg:venturescript-simulate} and
  \ref{alg:venturescript-logpdf}.}{}
\end{table}

\begin{subalgorithms}
\captionof{algorithm}{%
    \texttt{simulate} for CGPMs in a general probabilistic programming
    environment.}
\label{alg:venturescript-simulate}
\begin{algorithmic}[1]
    \Function{Simulate}{}
    \For{$k = 1,\dots,K$}
        \Comment{for each trace $k$}
        \State $w_k \gets $ \textsc{Compute-Trace-Weight} $(k, \x_{[r,E]})$
        \Comment{retrieve the weight}
    \EndFor
    \Sample{$j$}{$\textsc{Categorical} (\set{w_1,\dots,w_k})$}
      \Comment{importance resample the traces}
    \State $\set{\x_{[j,r,Q]},\z_{[j,r]}} \sim^\G$
      $|\set{\btheta_j,\Z_j,\D_j}$
      \Comment{transition operator leaving target invariant}
    \State \Return $\x_{[j,r,Q]}$
        \Comment{select samples of query set from resampled trace}
    \EndFunction
\end{algorithmic}

\captionof{algorithm}{%
    \texttt{logpdf} for CGPMs in a general probabilistic programming
    environment.}
\label{alg:venturescript-logpdf}
\begin{algorithmic}[1]
    \Function{LogPdf}{}
    \For{$k=1,\dots,K$}
        \Comment{for each trace $k$}
        \State $w_k \gets $ \textsc{Compute-Trace-Weight} $(k, \x_{[r,E]})$
        \Comment{retrieve the weight}
        \For{$t = 1,\dots, T$}
            \Comment{obtain $T$ samples of latents in scope $r$}
            \State $\z_{[k,t,r]} \sim^\G$
              $|\set{\btheta_k,\Z_k,\D_k}$
              \Comment{transition operator leaving target invariant}
            \State $h_{[k,t]} \gets
                \prod_{q\in Q}p(x_{[r,q]}|\btheta_k, \z_{[k,t,r]})$
                \Comment{compute a density estimate}
        \EndFor
        \State $r_k \gets \frac{1}{T}\sum_{t=1}^{T}{h_{[k,t]}}$
          \Comment{aggregate density estimates by simple Monte Carlo}
        \State $q_k \gets r_kw_k$
          \Comment{importance weight the estimate}
      \EndFor
    \State \Return $\log\left(\sum_{k=1}^{K}q_k\right) -
      \log\left(\sum_{k=1}^{K}w_k\right)$
        \Comment{weighted importance sampling estimator}
    \EndFunction
\end{algorithmic}

\captionof{algorithm}{%
    Computing the weight of a trace on a per-query basis.}
\label{alg:venturescript-weight-trace}
\begin{algorithmic}[1]
    \Function{Compute-Trace-Weight}{}
        (\texttt{trace}: $k$, \texttt{evidence}: $\x_{[r,E]}$)
    \Let{$\D_k$}{$\D_k \cup \y_r$}
        \Comment{observe the input variable}
    \If{$\z_{[k,r]} \not\in \Z_k$}
        \Comment{if member $r$ has unknown local latents}
      \State $\z_{[k,r]} \sim^\G$
        $|\set{\btheta_k, \Z_k, \D_k}$
        \Comment{sample from the prior}
    \EndIf
    \Let{$\D_k$}{$\D_k\cup \x_{[r,E]}$}
      \Comment{observe new evidence variables}
    \State $w_k \gets
        \prod\limits_{e\in E}\pG(x_{[r,e]}|\btheta_k,\z_{[k,r]})$
      \Comment{weight by likelihood of $\x_{[r,E]}$}
    \State \Return $w_k$
    \EndFunction
\end{algorithmic}

\end{subalgorithms}


\section{Integrating Conditional Generative Population Models into BayesDB}
\label{sec:bayesdb}

Without probabilistic programming systems and languages that treat data analysis computationally, it is difficult to both utilize the expressive power of CGPMs and use general-purpose inference machinery to develop and query them.
In this section, we show how CGPMs have been integrated into BayesDB, a probabilistic programming platform with two languages:
the Bayesian Query Language (BQL) for model-independent querying, and the Metamodeling Language (MML) for model discovery and building.
We first describe how simple BQL queries map directly to invocations of the CGPM interface.
We then show how to compose CGPMs into networks, and outline new expressions in MML used to construct populations and networks of CGPMs.
The experiments in Section \ref{sec:applications} illustrate how extending BayesDB with CGPMs can be used for non-trivial data analysis tasks.


\begin{figure}[ht]
\begin{subfigure}[b]{0.38\textwidth}
\begin{Verbatim}[fontsize=\footnotesize]
%mml CREATE TABLE t FROM "customers.csv"
%mml CREATE POPULATION p FOR t(
....   GUESS STATTYPES FOR (*);
....   MODEL age AS MAGNITUDE
.... );

%mml CREATE METAMODEL m FOR p
....  WITH BASELINE crosscat(
....   SET CATEGORY MODEL
....     FOR age TO lognormal;
....   OVERRIDE GENERATIVE MODEL
....     FOR income GIVEN age, state
....     USING linear_regression
.... );

%mml INITIALIZE 4 MODELS FOR m;
%mml ANALYZE m FOR 1 MINUTE;

%bql SIMULATE age, state
....   GIVEN income = 145000
....   FROM p LIMIT 100;
\end{Verbatim}
\begin{tabular}{l|l|l}
{\tt\scriptsize age} & {\tt\scriptsize state} & {\tt\scriptsize income} \\
\hline
{\tt\scriptsize 29}  & {\tt\scriptsize CA}    & {\tt\scriptsize 145000} \\
{\tt\scriptsize 61}  & {\tt\scriptsize TX}    & {\tt\scriptsize 145000} \\
{\tt\scriptsize 48}  & {\tt\scriptsize MA}    & {\tt\scriptsize 145000} \\
\end{tabular}
\end{subfigure}%
\begin{subfigure}[b]{0.62\textwidth}
\usetikzlibrary{positioning}
\tikzstyle{block} = [draw, rectangle, minimum height=2em]
\begin{tikzpicture}[node distance=1cm,>=stealth]

    \node[block, dashed] (populations) {\small\bf Populations};
    \node[block, dashed, left=of populations, xshift=.75cm] (tables)
        {\small\bf Data Tables};
    \node[block, dashed, right =of populations, xshift=-.75cm] (metamodels)
        {\small\bf Metamodels};

    \node[block, minimum width = 12em, ultra thick, below=1.5 of populations]
        (mml-interpreter)
        {\small\bf \begin{tabular}{c}
            Metamodeling Language \\ (MML) Interpreter \\
            \end{tabular}
        };

    \node[block, minimum width=5em, left=of mml-interpreter, xshift=.75cm] (mml)
        {\small\tt \begin{tabular}{c}
            MML \\ Script \\
            \end{tabular}
        };

    \node[block, right=of mml-interpreter, xshift=-.6cm] (cgpm-library)
        {\small\bf \begin{tabular}{c}
            CGPM \\ Library \\
            \end{tabular}
        };

    \node[block, dashed, minimum width = 12em, below=of mml-interpreter]
        (cgpm)
        {\small\bf \begin{tabular}{c}
            Composable Generative \\ Population Models (CGPMs) \\
            \end{tabular}
        };

    \node[block, minimum width = 12em, ultra thick, below=of cgpm] (bql-interpreter)
        {\small\bf \begin{tabular}{c}
            Bayesian Query Language \\ (BQL) Interpreter
            \end{tabular}
        };

    \node[block, minimum width=5em, left=of bql-interpreter, xshift=.75cm] (bql)
        {\small\tt \begin{tabular}{c}
            BQL \\ Query \\
            \end{tabular}
        };

    \node[block, below=of bql-interpreter] (result)
        {\small\tt Query Results};

    \draw[thick, dashed, <->] (tables) -- (mml-interpreter);
    \draw[thick, dashed, <->] (populations) -- (mml-interpreter);
    \draw[thick, dashed, <->] (metamodels) -- (mml-interpreter);
    \draw[thick, <->] (cgpm-library) -- (mml-interpreter);
    \draw[thick, ->] (mml) -- (mml-interpreter);
    \draw[thick, ->] (mml-interpreter) -- (cgpm);
    \draw[thick, ->] (cgpm) -- (bql-interpreter);
    \draw[thick, ->] (bql) -- (bql-interpreter);
    \draw[thick, ->] (bql-interpreter) -- (result);

\end{tikzpicture}
\end{subfigure}%
\bcaption{%
System architecture and modules that comprise BayesDB.}{%
The Metamodeling Language interpreter reads (i) population schemas to define variables and statistical types, (ii) metamodel definitions to apply automatic and custom modeling strategies for groups of variables in the population, and (iii) commands such as \texttt{INITIALIZE}, which instantiates an ensemble of CGPM networks, and \texttt{ANALYZE}, which applies inference operators to CGPMs to learn from observed data.
The Bayesian Query Language is a model-independent probabilistic query language that allows users to (i) \texttt{ESTIMATE} properties of CGPMs such strength and existence of dependence relationships between variables, similarity between members, and conditional density queries, and (ii) \texttt{SIMULATE} missing or hypothetical observations subject to user-provided constraints.
Together, these components allow users to build population models and query the probable implications of their data.}
\label{fig:architecture}
\end{figure}

\clearpage


\subsection{Querying composable generative population models using the Bayesian Query Language}

The BQL interpreter allows users to ask probabilistic questions about populations using a structured query language.
Figure~\ref{fig:bql-interpreter} shows how the BQL queries \texttt{SIMULATE} and \texttt{ESTIMATE PROBABILITY OF} translate into invocations of \texttt{simulate} and \texttt{logpdf} for an illustrative population and CGPM.

BQL defines a large collection of row-wise and column-wise estimators for CGPMs \citep[Sec. 3.2.2]{mansinghka2015}, such as \texttt{MUTUAL INFORMATION}, \texttt{DEPENDENCE PROBABILITY} and \texttt{SIMILIARITY WITH RESPECT TO}.
These quantities admit default implementations in terms of Monte Carlo estimators formed by \texttt{simulate} and \texttt{logpdf}, and any CGPM may override the BQL interpreter's generic implementations with a custom, optimized implementation.
A full description of implementing BQL in terms of the CGPM interface is beyond the scope of this work.


\begin{figure}[ht]
\begin{subtable}{\textwidth}
\small
\centering
\begin{tabular}{@{}lllll@{}} \toprule
  \texttt{rowid}
    & \texttt{a}
    & \texttt{b}
    & \texttt{c}
    & \texttt{d} \\
  \midrule
  1
    & 57
    & 2.5
    & Male
    & 15 \\
  2
    & 15
    & 0.8
    & Female
    & 10 \\
  3
    & NA
    & 1.4
    & NA
    & NA \\
  \vdots
    & \vdots
    & \vdots
    & \vdots
    & \vdots \\
  $r$
    & $x_{[r,a]}$
    & $x_{[r,b]}$
    & $x_{[r,c]}$
    & $y_{[r,d]}$ \\
  \vdots
    & \vdots
    & \vdots
    & \vdots
    & \vdots \\
  \bottomrule
\end{tabular}
\caption{A population represented as an infinite table in BayesDB, modeled by a CGPM $\G$ which generates variables \texttt{a}, \texttt{b}, and \texttt{c} as outputs, and requires variable \texttt{d} as input.}
\label{fig:bql-interpreter-1}
\end{subtable}
\bigskip

\begin{subtable}{\textwidth}
\small
\begin{tabular*}{\textwidth}{@{}l|l@{}}
  \toprule
  \textbf{BQL}
    & \texttt{SIMULATE a FROM G GIVEN d=12 WHERE rowid=3 LIMIT 2}\\
  \textbf{CGPM}
    & \texttt{simulate}
      ($\G$,
        \texttt{member}: $3$,
        \texttt{query}: $\set{\texttt{a}}$,
        \texttt{evidence}: $\set{(\texttt{d},12)}$)\\
  \textbf{Quantity}
    & $s_i \sim^\G
        x_{[3,a]}|\set{y_{[3,d]}=3, x_{[3,b]}=1.4, \D}$
        for $i=1,2$ \\[.3cm]
  \textbf{Result}
    & \begin{tabular}{|l|l|l|}
      \hline
      \texttt{rowid} & \texttt{a} & \texttt{d} \\ \hline
      3 & 51 & 12  \\ \hline
      3 & 59 & 12  \\ \hline
      \end{tabular}\\
\bottomrule
\end{tabular*}
\caption{Mapping a \texttt{SIMULATE} query to the CGPM interface invocation of \texttt{simulate}. The sampled quantity $s_i$ also includes $\set{x_{[3,b]}=1.4}$ as a conditioning value, which was extracted from the dataset $\D$. The CGPM must condition on every observed value in $\D$, as well as additional per-query constraints specified by the user, such as $\set{y_{[3,d]}=3}$. The result is a table with two rows corresponding to the two requested samples.}
\label{fig:bql-interpreter-2}
\end{subtable}
\bigskip

\begin{subtable}{\textwidth}
\small
\begin{tabular*}{\textwidth}{@{}l|l@{}}
  \toprule
  \textbf{BQL}
    & \texttt{ESTIMATE PROBABILITY OF a=49, c=`MALE'
      GIVEN d=12 FROM G WHERE rowid=3}\\
  \textbf{CGPM}
    & \texttt{logpdf}(
      $\G$,
      \texttt{member}: $3$,
      \texttt{query} : $\set{(\texttt{a}, 49),
        (\texttt{c},\texttt{`MALE'})}$
      \texttt{evidence} : $\set{(\texttt{d},12)}$) \\
  \textbf{Quantity}
    & $\pG(x_{[3,a]} = 49, x_{[3,c]} = \texttt{`MALE'}
      | y_{[3,d]} = 12, x_{[3,b]} = 1.4, \D)$ \\ [.3cm]
  \textbf{Result}
    & \begin{tabular}{@{}|l|l|l|l|l|@{}} \hline
        \texttt{rowid} & \texttt{a} & \texttt{c} & \texttt{d}
          & \texttt{bql\_pdf((a,c),(d))}\\ \hline
          3 & 49 & `Male' & 12 & 0.117 \\ \hline
      \end{tabular}\\
\bottomrule
\end{tabular*}
\caption{Mapping an \texttt{ESTIMATE PROBABILITY OF} query to the CGPM interface invocation of \texttt{logpdf}. The output is a table with a single row that contains the value of the queried joint density.}
\label{fig:bql-interpreter-3}
\end{subtable}

\bcaption{%
Translating BQL queries into invocations of the CGPM interface.}{}
\label{fig:bql-interpreter}
\end{figure}

\clearpage


\subsection{Compositional networks of composable generative population models}
\label{subsec:composition}

Our development of CGPMs has until now focused on their computational interface and their internal probabilistic structures.
In this section, we outline the mathematical formalism which justifies closure of CGPMs under input/output composition.
For a collection of CGPMs $\set{\G_k:k\in[K]}$ operating on the same population $\mathcal{P}$, we will show how they be organized into a generalized directed graph which itself is a CGPM $\G_{[K]}$, and provide a Monte Carlo strategy for performing joint inference over the outputs and inputs to the internal CGPMs.
This composition allows complex probabilistic models to be built from simpler CGPMs.
They communicate with one another using the \texttt{simulate} and \texttt{logpdf} interface to answer queries against the overall network.
In the next section, we describe the surface syntaxes in MML to construct networks of CGPMs in BayesDB.

Let $\bm{v}=(v_1,\dots,v_T)$ be the variables of $\mathcal{P}$, and $\G_a$ be a CGPM which generates outputs $\bm{v}^{out}_a=(v^{out}_{[a,1]},\dots,v^{out}_{[a,O_a]})$, accepts inputs $\bm{v}^{in}_a=(v^{in}_{[a,1]},\dots,v^{in}_{[a,I_a]})$, and satisfies $(\bm{v}^{out}_a\cup\bm{v}^{in}_a)\subset\bm{v}$.
Similarly, consider another CGPM $\G_b$ on the same population with outputs $\bm{v}^{out}_b$ and inputs $\bm{v}^{in}_b$.
The composition $(\G_{[b, \mathcal{B}]} \circ \G_{[a,\mathcal{A}]})$ applies the subset of outputs $\bm{v}^{out}_{[a,\mathcal{A}]}$ of $\G_a$ to the subset of inputs $\bm{v}^{in}_{[b,\mathcal{B}]}$ of $\G_b$, resulting in a new CGPM $\G^c$ with output $(\bm{v}^{out}_a\cup\bm{v}^{out}_b)$ and input $(\bm{v}^{in}_a\cup\bm{v}^{out}_{[b,\backslash\mathcal{B}]})$.
The rules of composition require that $(\bm{v}^{out}_a \cap \bm{v}^{out}_b) = \varnothing$ i.e. $\G_a$ and $\G_b$ do not share any output, and that $\bm{v}^{out}_{[a,\mathcal{A}]}$ and $\bm{v}^{in}_{[b,\mathcal{B}]}$ correspond to the same subset of variables in the original population $\mathcal{P}$.
Generalizing this idea further, a collection of CGPMs $\set{\G_k:k\in[K]}$ can thus be organized as a graph where node $k$ represents internal CGPM $\G_k$, and the labeled edge $a_{\mathcal{A}} \to b_{\mathcal{B}}$ denotes the composition $(\G_{[b,\mathcal{B}]} \circ \G_{[a,\mathcal{A}]})$.
These labeled edges between different CGPMs in the network must form a directed acyclic graph.
However, elements $x_{[k,r,i]}$ and $x_{[k,r,j]}$ of the same member $r$ within any particular $\G_k$ are only required to satisfy constraint \eqref{eq:indep-gpm} which may in general follow directed and/or undirected dependencies.
The topology of the overall CGPM network $\G_{[K]}$ can be summarized by its generalized adjacency matrix $\pi_{[K]}:=\set{\pi_k: k\in[K]}$, where $\pi_k=\set{(p,t): v^{out}_{[p,t]}\in\bm{v}^{in}_k}$ is the set of all output elements from upstream CGPMs connected to the inputs of $\G_k$.

To illustrate that the class of CGPMs is closed under composition, we need to show how the network $\G_{[K]}$ implements the interface.
First note that $\G_{[K]}$ produces as \texttt{outputs} the union of all output variables of its constituent CGPMs, and takes as \texttt{inputs} the collection of variables in the population which are not the output of any CGPM in the network.
The latter collection of variables are ``exogenous'' to the network, and must be provided for queries that require them.

The implementations of \texttt{simulate} and \texttt{logpdf} against $\G_{[K]}$ are shown in Algorithms~\ref{alg:network-inference-simulate} and \ref{alg:network-inference-logpdf}.
Both algorithms use an importance sampling scheme which combines the methods provided by each individual node $\G_k$, and a shared forward-sampling subroutine in Algorithm~\ref{alg:network-inference-forward}.
The estimator for \texttt{logpdf} uses ratio likelihood weighting; both estimators derived from lines \ref{alg-line:network-inference-logpdf-joint} and \ref{alg-line:network-inference-logpdf-marginal} of Algorithm~\ref{alg:network-inference-logpdf} are computed using unnormalized importance sampling, so the ratio estimator on line \ref{alg-line:network-inference-logpdf-ratio} is exact in the infinite limit of importance samples $J$ and $J'$.
The algorithms explicitly pass the member id $r$ between each CGPM so that they agree about which member-specific latent variables are relevant for the query, while preserving abstraction boundaries.
The importance sampling strategy used for compositional \texttt{simulate} and \texttt{logpdf} may only be feasible when the networks are shallow and the primitive CGPMs are fairly noisy; better Monte Carlo strategies or perhaps even variational strategies may be needed for deeper networks, and are left to future work.

The network's \texttt{infer} method can be implemented by invoking \texttt{infer} separately on each internal CGPM node.
In general, several improvements on this baseline strategy are possible and are also interesting areas for further research (Section~\ref{sec:discussion}).

\clearpage

\begin{table}[ht]
\centering
  \begin{tabular*}{\textwidth}{l|l}
    \toprule
    \textbf{Parameter} & \textbf{Symbol} \\ \midrule
    number of importance samples & $J, J'$ \\
    identifier of the population & $r$ \\
    indices of CGPM nodes in the network & $k=1,2,\dots,K$ \\
    CGPM representing node $k$ & $\G_k$ \\
    parents of node $k$ & $\pi_k$ \\
    input variables exogenous to network for node $k$ & $\y_{[k,r]}$ \\
    query set for node $k$ & $\x_{[k,r,Q_k]}$ \\
    evidence set for node $k$ & $\x_{[k,r,E_k]}$ \\
    query/evidence sets aggregated over all nodes in network
      & $\x_{[r,A]} = \underset{k\in[K]}{\cup}{\x_{[k,r,A_k]}}$ \\
    \bottomrule
  \end{tabular*}
  \bcaption{%
  Parameters and symbols used in Algorithms~\ref{alg:network-inference-simulate}, \ref{alg:network-inference-logpdf}, and \ref{alg:network-inference-forward}.}{%
  All parameters provided to the functions in which they appear.
  \textsc{Weighted-Sample} ignores \texttt{query} and \texttt{evidence} from the global environment, and is provided with an explicit set of constrained nodes by \textsc{Simulate} and \textsc{LogPdf}.}
\end{table}

\begin{subalgorithms}
\captionof{algorithm}{\texttt{simulate} in a directed acyclic network of CGPMs.}
\label{alg:network-inference-simulate}
\begin{algorithmic}[1]
  \Function{Simulate}{}
  \For{$j=1,\dots,J$}
      \Comment{generate $J$ importance samples}
    \Let{$(\s_j,w_j)$}{\textsc{Weighted-Sample} ($\x_{[r,E]}$)}
    \Comment{retrieve sample weighted by evidence}
  \EndFor
  \Let{$m$}{$\textsc{Categorical}(\set{w_1,\dots,w_J})$}
      \Comment{resample importance sample}
  \State \Return{$\underset{k\in[K]}{\cup}{\x_{[k,r,Q_k]} \in \s_m}$}
      \Comment{overall sample of query variables}
  \EndFunction
\end{algorithmic}

\captionof{algorithm}{\texttt{logpdf} in a directed acyclic network of CGPMs.}
\label{alg:network-inference-logpdf}
  \begin{algorithmic}[1]
    \Function{LogPdf}{}
    \For{$j=1,\dots,J$} \label{alg-line:network-inference-logpdf-joint}
      \Comment{generate $J$ importance samples}
      \Let{$(\s_j,w_j)$}{
        \textsc{Weighted-Sample} ($\x_{[r,E]} \cup \x_{[r,Q]}$)}
              \Comment{joint density of query/evidence}
    \EndFor
    \For{$j=1,\dots,J'$} \label{alg-line:network-inference-logpdf-marginal}
      \Comment{generate $J'$ importance samples}
      \Let{$(\s'_j,w'_j)$}{
        \textsc{Weighted-Sample} ($\x_{[r,E_k]}$)}
            \Comment{marginal density of evidence}
    \EndFor
    \State
      \Return{$\log\left(\sum_{[J]}w_j/\sum_{[J']}w_j\right) - \log(J/J')$}
      \Comment{likelihood ratio importance estimator}
      \label{alg-line:network-inference-logpdf-ratio}
    \EndFunction
  \end{algorithmic}

\captionof{algorithm}
  {Weighted forward sampling in a directed acyclic network of CGPMs.}
\label{alg:network-inference-forward}
\begin{algorithmic}[1]
  \Function{Weighted-Sample}{} (\texttt{constraints}: $\x_{[r,C_k]}$)
  \Let{$(\s, \log{w})$}{$(\varnothing, 0)$}
      \Comment{initialize empty sample with zero weight}
  \For{$k\in$ \textsc{TopoSort} $(\set{\pi_1\dots\pi_K})$}
      \Comment{topologically sort the adjacency matrix}
    \Let
        {$\tilde{\y}_{[k,r]}$}
        {$\y_{[k,r]}\cup\set{x_{[p,r,t]}\in\s:(p,t)\in\pi_k}$}
      \Comment{retrieve required inputs at node $k$}
    \Let{$\log{w}$}{$\log{w} +
        \texttt{logpdf}(\G_k, r,\x_{[k,r,C_k]},\tilde{\y}_{[k,r]})$}
        \Comment{{\small{update weight by constraint likelihood}}}
    \Let{$\x_{[k,r,\backslash C_k]}$}{$\texttt{simulate}(
      \G_k, r,\backslash C_k, \x_{[k,r,C_k]} \cup \tilde{\y}_{[k,r]})$}
      \Comment{simulate unconstrained nodes}
    \Let{$\s$}{$\s\cup\x_{[k,r,C_k\cup\backslash C_k]}$}
      \Comment{append to sample}
  \EndFor
  \State \Return{$(\s,w)$}
      \Comment{overall sample and its weight}
  \EndFunction
\end{algorithmic}

\end{subalgorithms}

\clearpage


\subsection{Building populations and networks of composable generative population models with the Metamodeling Language}
\label{subsec:bayesdb-mml}

As shown in Figure~\ref{fig:architecture}, the MML interpreter in BayesDB interacts with data tables and populations, metamodels, and a library of CGPMs.
Population schemas are MML programs which are used to declare a list of variables and their statistical types.
Every population is backed by a base table in BayesDB, which stores the measurements. Metamodel definitions are MML programs which are used to declare a composite network of CGPMs for a given population.
The internal CGPMs nodes in this network come from the CGPM library available to BayesDB.
After declaring a population and a metamodel for it, further MML commands are used to instantiate stochastic ensembles of CGPM networks (\texttt{INITIALIZE}), and apply inference operators to them (\texttt{ANALYZE}).

In this section, we describe the surface level syntaxes in the Metamodeling Language for population schemas, metamodel definitions, and other MML commands.
We also describe how to use the Bayesian Query Language to query ensembles of CGPMs at varying levels of granularity.
A formal semantics for MML that precisely describes the relationship between the compositional surface syntax and a network of CGPMs is left for future work.

\subsubsection{Population Schemas}

A population schema declares a collection of variables and their statistical types.

\begin{itemize}
\item[] \texttt{%
        \oliveg{CREATE POPULATION} <\blue{p}> \oliveg{FOR} <\blue{table}>
            \oliveg{WITH SCHEMA} (<\red{schemum}>[; ...]);}

    Declares a new population \texttt{p} in BayesDB.
    The token \texttt{table} references a database table, which stores the measurements and is known as the base table for \texttt{p}.

\item[] \texttt{%
        \red{schemum} :=
            \oliveg{MODEL} <\blue{var-names}> \oliveg{AS} <\blue{stat-type}>}

    Uses \texttt{stat-type} as the statistical data type for all the variables named in \texttt{var-names}.

\item[] \texttt{%
        \red{schemum} := \oliveg{IGNORE} <\blue{var-names}>}

    Excludes \texttt{var-names} from the population.
    This command is typically applied for columns in the base table representing unique names, timestamps, and other metadata.

\item[] \texttt{%
        \red{schemum} := \oliveg{GUESS STATTYPES FOR} (* | <\blue{var-names}>)}

    Uses existing measurements in the base table to guess the statistical data types of columns in the table.
    When the argument is (\texttt{*}), the target columns are all those which do not appear in \texttt{MODEL} or \texttt{IGNORE}.
    When the argument is (\texttt{var-names}), only those subset of columns are guessed.
\end{itemize}

Every column in the base table must have a derivable policy (guess, ignore, or explicitly model with a user-provided statistical data type) from the schema.
The statistical data types available in MML are shown in Table~\ref{tab:stattypes}.
The \oliveg{\textbf{\texttt{GUESS}}} command is implemented using various heuristics on the measurements (such as the number of unique values, sparsity of observations, and SQL \texttt{TEXT} columns) and only assigns a variable to either \texttt{NOMINAL} or \texttt{NUMERICAL}.
Using a more refined statistical type for a variable is achieved with an explicit \textbf{\texttt{\oliveg{MODEL}...\oliveg{AS}}} command.
Finally, two populations identical same base tables and variables, but different statistical type assignments, are considered distinct populations.

\clearpage

\subsubsection{Metamodel Definitions}
\label{subsubsec:bayesdb-mml-metamodel-definitions}

After creating a population $\mathcal{P}$ in BayesDB, we use metamodel definitions to declare CGPMs for the population.
This MML program specifies both the topology and internal CGPM nodes of the network (Section~\ref{subsec:composition}).
Starting with a baseline CGPM at the ``root''of the graph, nodes and edges are constructed by a sequence overrides that extract variables from the root node and place them into newly created CGPM nodes.
The syntax for a metamodel definition is:

\begin{itemize}
\item[] \texttt{%
        \oliveg{CREATE METAMODEL} <\blue{m}> \oliveg{FOR} <\blue{population}>
        \oliveg{WITH BASELINE} <\red{baseline-cgpm}>\newline
        [(<\red{schemum}>[; ...])];}

    Declares a new metamodel \texttt{m}.
    The token \texttt{population} references a BayesDB population, which contains a set of variable names and their statistical types and is known as the base population for \texttt{m}.

\item[] \texttt{%
    \red{baseline-cgpm} ::= (crosscat | multivariate\_kde | generative\_knn)}

    Identifies the automatic model discovery engine, which learns the full joint distribution of all variables in the \texttt{population} of \texttt{m}.
    Baselines include
    Cross-Categorization
        (Section~\ref{subsec:implementation-crosscat}),
    Multivariate Kernel Density Estimation
        (Section~\ref{subsec:implementation-kde}), or
    Generative K-Nearest-Neighbors
        (Section~\ref{subsec:implementation-knn}).

\item[] \texttt{%
        \red{schemum} :=
            \oliveg{OVERRIDE GENERATIVE MODEL FOR} <\blue{output-vars}>\newline
            [\oliveg{GIVEN} <\blue{input-vars}>]
            [\oliveg{AND EXPOSE}
                (<\blue{exposed-var}> <\blue{stat-type}>)[, ...]]\newline
            \oliveg{USING} <\blue{cgpm-name}>}

    Overrides \texttt{baseline-cgpm} by creating a new node in the CGPM network. The node generates \texttt{output-vars}, possibly requires the specified \texttt{input-vars}.
    Additionally, the CGPM may expose some of its latent variable as queryable outputs.
    The token \texttt{cgpm-name} refers to the name of the CGPM which is overriding \texttt{baseline-cgpm} on the specified subpart of the joint distribution.

\item[] \texttt{%
        \red{schemum} :=
            \oliveg{SET CATEGORY MODEL FOR} <\blue{output-var}>
            \oliveg{TO} <\blue{primitive-cgpm-name}>}

    (This command is only available when \texttt{baseline-cgpm} is \texttt{crosscat}.)

    Replaces the default category model used by \texttt{crosscat} for
    \texttt{output-var}, based on its statistical type, with an alternative
    \texttt{primitive-cgpm} that is also applicable to that statistical type
    (last column of Table~\ref{tab:stattypes}).
\end{itemize}

To answer arbitrary BQL queries about a population, BayesDB requires each CGPM to carry a full joint model over all the population variables.
Thus, each metamodel is declared with a baseline CGPM, such as CrossCat, a non-parametric Bayesian structure learner for high-dimensional and heterogeneous data tables \citep{mansinghka2015-2}, among others outlined in Section~\ref{sec:implementation}.
It is important to note that the \texttt{input-vars} in the \texttt{\oliveg{OVERRIDE MODEL}} command may be the outputs of not only the baseline but any collection of upstream CGPMs.
It is also possible to completely override the baseline by overriding all the variables in the population.

\subsubsection{Homogeneous Ensembles of CGPM Networks}
\label{subsubsec:bayesdb-mml-homogenous}

In BayesDB, a metamodel $\M$ is formally defined as an ensemble of CGPM networks $\set{(\G_k,w_k)}_{i=1}^{N}$, where $w_k$ is the weight of network $\G_k$ \citep[Section 3.1.2]{mansinghka2015}.
The CGPMs in $\M$ are homogeneous in that (from the perspective of MML) they have the same metamodel definition, and (from the perspective of the CGPM interface) they are all \texttt{create}d with the same \texttt{population}, \texttt{inputs}, \texttt{outputs}, and \texttt{binary}.
The ensemble $\M$ is populated with $K$ instances of CGPMs using the following MML command:

\begin{itemize}
\item[] \texttt{%
        \oliveg{INITIALIZE}  <\blue{K}> \oliveg{MODELS FOR} <\blue{metamodel}>;
        }

    Creates $K$ independent replicas of the composable generative population
    model network contained in the MML definition of \texttt{metamodel}.
\end{itemize}

CGPM instances in the ensemble are different in that BayesDB provides each $\G_k$ a unique \texttt{seed} during \texttt{create}.
This means that invoking \texttt{infer}($\G_k$, \texttt{program}: $\mathcal{T}$) causes each network's internal state to evolve differently over the course of inference (when $\mathcal{T}$ contains non-deterministic execution).
In MML surface syntax, \texttt{infer} is invoked using the following command:

\begin{itemize}
\item[] \texttt{%
        \oliveg{ANALYZE} <\blue{metamodel}> \oliveg{FOR} <\blue{K}>
            (\oliveg{ITERATIONS} | \oliveg{SECONDS}) [(<\red{plan}>)];
        }

        Runs analysis (in parallel) on all the initialized CGPM networks in the ensemble, according to an optional inference \texttt{plan}.

\item[] \texttt{%
        \red{plan} := (\oliveg{VARIABLES} | \oliveg{SKIP}) <\blue{var-names}>
        }

        If \texttt{VARIABLES}, then runs analysis on all the CGPM nodes which have at least one output variable in \texttt{var-names}.
        If \texttt{SKIP}, then then transitions all the CGPM nodes except those which have a an output variable in \texttt{var-names}.
        As outlined at the end of Section~\ref{subsec:composition}, each CGPM node is learned independently at present time.
\end{itemize}

Weighted ensembling of homogeneous CGPMs can be interpreted based on the modeling and inference tactics internal to a CGPM.
For example, in Bayesian CGPM network where \texttt{\oliveg{ANALYZE}} invokes MCMC transitions, each $\G_k$ may represent a different posterior sample; for variational inference, each $\G_k$ may converge to a different set of latent parameters due to different random initializations.
More extensive syntaxes for inference plans in MML are left for future work.

\subsubsection{Heterogeneous Ensembles of CGPM Networks}
\label{subsubsec:bayesdb-bql-heterogenous}

Section~\ref{subsubsec:bayesdb-mml-homogenous} defined a metamodel $\M$ as an ensemble of homogeneous CGPM networks with the same metamodel definition.
It is also possible construct a heterogeneous ensemble of CGPM networks by defining a set of metamodels $\set{\M_1,\dots,M_K}$ for the same population $\mathcal{P}$ but with different metamodel definitions.
Let $\G_{[k,t]}$ be the $t\textsuperscript{th}$ CGPM network in the metamodel $\M_k$.
The Bayesian Query Language is able to query CGPM networks at three levels of granularity, starting from the most coarse to the most granular.

\begin{itemize}
\item[] \texttt{%
        (\oliveg{ESTIMATE} |  \oliveg{SIMULATE} | \oliveg{INFER})
        <\blue{bql-expression}> \oliveg{FROM} <\blue{population}>;}

        Executes the BQL query by aggregating responses from all metamodels $\set{\M_1,\dots,\M_k}$ defined for \texttt{<population>}.

\item[] \texttt{%
        (\oliveg{ESTIMATE} |  \oliveg{SIMULATE} | \oliveg{INFER})
        <\blue{bql-expression}> \oliveg{FROM} <\blue{population}>\newline
            \oliveg{MODELED BY} <\blue{metamodel-k}>;}

        Executes the BQL query by aggregating responses from all the CGPM networks $\set{\G_{[k,t]}}$  that have been initialized with the MML definition for \texttt{<metamodel-k>}.

\item[] \texttt{%
        (\oliveg{ESTIMATE} |  \oliveg{SIMULATE} | \oliveg{INFER})
        <\blue{bql-expression}> \oliveg{FROM} <\blue{population}>\newline
            \oliveg{MODELED BY} <\blue{metamodel-k}>
            \oliveg{USING MODEL} <\blue{t}>;}

        Executes the BQL query by returning the single response from $\G_{[k,t]}$ in \texttt{<metamodel-k>}.
\end{itemize}

Monte Carlo estimators obtained by \texttt{simulate} and \texttt{logpdf} remain well-defined even when the ensemble contains heterogeneous CGPMs.
All CGPMs across different metamodels are defined for the same population, which determines the statistical types of the variables. This guarantees that the associated supports and (product of) base measures (from Table~\ref{tab:stattypes}) for \texttt{simulate} and \texttt{logpdf} queries are all type-matched.


\subsection{Composable generative population models generalize and extend generative population models in BayesDB}
\label{subsec:bayesdb-compare}

It is informative to compare both the conceptual and technical differences between generative population models (GPMs) in BayesDB \citep{mansinghka2015} with composable generative population models (CGPMs).
In its original presentation, the GPM interface served the purpose of being the primary vehicle for motivating BQL as a model-independent query language \citep[Sec.3.2]{mansinghka2015}.
Moreover, GPMs were based around CrossCat as the baseline model-discovery engine \citep[Sec. 4.5.1]{mansinghka2015}, which provided good solutions for several data analysis tasks. However, by not accepting inputs, GPMs offered no means of composition; non-CrossCat objects, known as ``foreign predictors'', were discriminative models embedded directly into the CrossCat joint density \citep[Sec. 4.4.2]{mansinghka2015}.
By contrast, the main purpose of the CGPM interface is to motivate more expressive MML syntaxes for building hybrid models, comprised of arbitrary generative and discriminative components.
Since CGPMs natively accept inputs, they admit a natural form of composition (Section~\ref{subsec:composition}) which does violate the internal representation of any particular CGPM.

The computational interface and probabilistic structure of GPMs and CGPMs are different in several respects.
Because GPMs were presented as Bayesian models with Markov Chain Monte Carlo inference \citep[Sec. 4.2]{mansinghka2015}, both \texttt{simulate} and \texttt{logpdf} were explicitly conditioned on a particular set of latent variables extracted from some state in the posterior inference chain \citep[Sec. 3.1.1]{mansinghka2015}.
On the other hand, CGPMs capture a much broader set of model classes, and \texttt{simulate} and \texttt{logpdf} do not impose any conditioning constraints internal to the model besides conditioning on input variables and the entire dataset $\D$.
Internally, GPMs enforced much stronger assumptions regulating inter-row independences; all the elements in a row are conditionally independent give a latent variable \cite[Sec.3.1]{mansinghka2015}, effectively restricting the internal structure to a directed graphical model.
CGPMs allow for arbitrary coupling between elements within a row from Eq~\eqref{eq:indep-gpm}, which uniformly expresses both directed and undirected probabilistic models, as well approaches which are not naturally probabilistic that implement the interface.
Finally, unlike GPMs, CGPMs may expose some of member-specific latent variables as queryable outputs. This features trades-off the model independence of BQL with the ability to learn and query the details of the internal probabilistic process encapsulated by the CGPM.

\clearpage


\section{Applications of Composable Generative Population Models}
\label{sec:applications}

The first part of this section outlines a case study applying compositional generative population models in BayesDB to a population of satellites maintained by the Union of Concerned Scientists.
The dataset contains 1163 entries, and each satellites has 23 numerical and categorical features such as its material, functional, physical, orbital and economic characteristics.
We construct a hybrid CGPM using an MML metamodel definition which combines (i) a classical physics model written as a probabilistic program in VentureScript, (ii) a random forest to classify a a nominal variable, (iii) an ordinary least squares regressor to predict a numerical variable, and (iv) principal component analysis on the real-valued features of the satellites.
These CGPMs allow us to identify satellites that probably violate their orbital mechanics, accurately infer missing values of anticipated lifetime, and visualize the dataset by projecting the satellite features into two dimensions.

The second part of this section explores the efficacy of hybrid compositional generative population models on a collection of common tasks in probabilistic data analysis by reporting lines of code and accuracy measurements against standard baseline solutions.
Large savings in lines of code and improved accuracy are demonstrated in several important regimes.
Most of the analysis of experimental results is contained in the figure gallery at the end of the section.

\subsection{Analyzing satellites using a composite CGPM built from causal probabilistic programs, discriminative machine learning, and Bayesian non-parametrics}
\label{subsec:experiment-satellites}

The left panel in Figure~\ref{fig:mml-hybrid} illustrates a session in MML which
declares the population schema for the satellites data, as well as the metamodel
definition for building the hybrid CGPM network that models various
relationships of interest between variables.%
\footnote{This program is executed in iVenture, an experimental interactive
probabilistic programming environment that supports running \texttt{\%bql},
\texttt{\%mml} and \texttt{\%venturescript} code cells, all of which operate on
a common underlying BayesDB instance and Venture interpreter.}
The \texttt{CREATE POPULATION} block shows the high-dimensional features of each satellite and their heterogeneous statistical types.
For simplicity, several variables such as \texttt{perigee\_km}, \texttt{launch\_mass\_kg} and \texttt{anticipated\_lifetime} have been modeled as \texttt{NUMERICAL} rather than a more refined type such as \texttt{MAGNITUDE}.
In the remainder of this section, we explain the CGPMs declared in the MML metamodel definition under the \texttt{CREATE METAMODEL} block, and refer to figures for results of BQL queries executed against them.

The PCA CGPM on line 34 of the metamodel definition generates as output five real-valued variables, and exposes the first two principal component scores to BayesDB.
This low-dimensional projection allows us to both visualize a clustering of the dataset in latent space, and discover oddities in the distribution of latent scores for satellites whose \texttt{class\_of\_orbit} is \texttt{elliptical}.
It also identifies a single satellite, in cyan at grid point $(1, 1.2)$, as a candidate for further investigation.
Figure~\ref{fig:pca-satellites} shows the result and further commentary on this experiment.

Four variables in the population relate to the orbital characteristics of
each satellite: \texttt{apogee\_km} $A$, \texttt{perigee\_km} $P$,
\texttt{period\_minutes} $T$, and \texttt{eccentricity} $e$.
These variables are constrained by the theoretical Keplerian relationships $e = \frac{A-P}{A+P}$ and $T=2\pi\sqrt{\frac{((A+P)/2)^3}{GM}}$, where $GM$ is a physical constant.
In reality, satellites deviate from their theoretical orbits for a variety of reasons, such orbital and measurement noise, having engines, or even data-entry errors.
The right panel of Figure~\ref{fig:mml-hybrid} shows a CGPM in pure VentureScript which accepts as input $\y_r=(A_r, P_r)$ (apogee and perigee), and generates as output $x_r = T_r$ (period).
The prior is a Dirichlet process mixture model on the error, based on a stochastic variant of Kepler's Law,
\begin{align*} & G \sim DP(\alpha, \textsc{Normal-Inverse-Gamma}(m,V,a,b))\\ & (\mu_r,\sigma_r^2) | G \sim G\\ & \epsilon_r | \y_r \sim \textsc{Normal}(\cdot|\mu_r,\sigma_r^2) && \textrm{where } \epsilon_r := T_r - \textsc{Kepler}(A_r,P_r). \end{align*}
While the internal details, external interface, and adapter which compiles the VentureScript source into a CGPM are beyond the scope of this paper, note that its MML declaration uses the \texttt{EXPOSE} command on line 45.
This command makes the inferred cluster identity and noise latent variables (lines 17 and 22 of the VentureScript program) available to BQL.
Figure~\ref{fig:kepler-overall} shows a posterior sample of the cluster assignments and error distribution, which identifies three distinct classes of anomalous satellites based on the magnitude of error.
For instance, satellite \texttt{Orion6} in the right panel of Figure~\ref{fig:kepler-overall}, belongs to a cluster with ``extreme'' deviation.
Further investigation reveals that \texttt{Orion6} has a period 23.94 minutes, a data-entry error for the true period of 24 hours (1440 minutes).

Figure~\ref{fig:confusion} shows the improvement in prediction accuracy achieved by the hybrid CGPM over the purely generative CrossCat baseline, for a challenging multiclass classification task.
As shown in lines 57-62 of the metamodel definition in Figure~\ref{fig:mml-hybrid}, the hybrid CGPM uses a random forest CGPM for the target variable \texttt{type\_of\_orbit} given five numerical and categorical predictors.
Figures~\ref{fig:confusion-rf} and \ref{fig:confusion-cc} shows the confusion matrices on the test set for both the composite and baseline CGPMs.
While both methods systematically confuse sun-synchronous with intermediate orbits, the use of a random forest classifier results in 11 less classification errors, or an improvement of 11 percentage points.
Using a purely discriminative model for this task, i.e. a random forest without a generative model over the features (not shown), would require additional logic and heuristic imputation on feature vectors in the test set, which general contained missing entries.

The final experiment in Figure~\ref{fig:cc-vs-kde} compares the posterior distribution of the vanilla CrossCat baseline and multivariate KDE for a two-dimensional density estimation task with nominal data types.
The task is to jointly simulate the \texttt{country\_of\_operator} and \texttt{purpose} for a hypothetical satellite, given that its \texttt{type\_of\_orbit} is geosynchronous.
The empirical conditional distribution from the dataset is shown in red.
Both CrossCat and multivariate KDE capture the posterior modes, although the distribution form KDE has a fatter tail, as indicated by the high number of samples classified as ``Other''.
The figure caption contains additional discussion.

There dozens of additional BQL queries that can be posed about the satellites population and, based on the analysis task of interest, answered using both the existing CGPMs in the hybrid metamodel as well as more customized CGPMs.
The empirical studies in this section has shown it is possible and practical to apply CGPMs in BayesDB to challenging data analysis tasks in a real-world dataset, and use BQL queries to compare their performance characteristics.

\subsection{Comparing code length and accuracy on representative data analysis tasks}
\label{subsec:loc-accuracy}

One of the most sparsely observed variables in the satellites dataset is the \texttt{anticipated\_lifetime}, with roughly one in four missing entries.
The analysis task in Figure~\ref{fig:loc-regression} is to infer the anticipated lifetime $x_*$ of a new satellite, given the subset of its numerical and nominal features $\y_*$ shown in the codeblock above the plot.
To quantify performance, the predictions of the CGPM were evaluated on a held-out set of satellites with known lifetimes.
Many satellites in both the training set and test set contained missing entries in their covariates, requiring the CGPM to additionally impute missing values in the predictors before forward simulating the regression.
Unlike the purely generative and purely discriminative baselines (shown in the legend), the hybrid CGPM learns both a joint distribution over the predictors and a discriminative model for the response, leading to significantly improved predictive performance.

The improvement in lines of code over the baseline methods in Figure~\ref{fig:loc-regression} is due to using combinations of (i) SQL for data processing, (ii) MML for model building, and (iii) BQL for predictive querying, in BayesDB.
All the baselines required custom logic for (i) manual data preprocessing such as reading csv files, (ii) Euclidean embedding of large categorical values, and (iii) heuristic imputation of missing features during train and test time (i.e. either imputing the response from its mean value, or imputing missing predictors from their mean values).
The left panel from Figure~\ref{fig:loc-regression-cgpm} shows and end-to-end session in BayesDB which preprocesses the data, builds the hybrid CGPM, runs analysis on the training set and computes predictions on the test set.
The right panel from Figure~\ref{fig:loc-regression-baseline} shows a single ad-hoc routine used by the Python baselines, which dummy codes a data frame with missing entries and nominal data types.
For nominal variables taking values in a large set, dummy coding with zeros may cause the solvers to fail when the system is under-determined.
The workaround in the code for baselines is to drop such problematic dimensions from the feature vector.
The regression in the hybrid CGPM does not suffer from this problem because, the default linear regressor in the CGPM library gives all parameters a Bayesian prior \citep{banerjee2008}, which smooths irregularities.

Figures~\ref{fig:independence-testing}, \ref{fig:dependence-strength}, \ref{fig:bivariate-classification} and \ref{fig:anomaly-detection} extend the lines of code and accuracy comparisons for CGPMs and baseline methods to several more tasks using diverse statistical methodologies. These figures further illustrate coverage and conciseness of CGPMs -- the captions detail the setup and commentary of each experiment in greater detail.

\clearpage

\begin{figure}[ht]
\bcaption{Building a hybrid CGPM in Venturescript and MML for the satellites
population.}{}
\label{fig:mml-hybrid}
\begin{subfigure}[t]{.5\textwidth}
\begin{Verbatim}[fontsize=\scriptsize, frame=single, numbers=left, label=\%mml, commandchars=&\[\]]
&fbvtcg[CREATE TABLE] &fbvtcb[satellites_ucs] &fbvtcg[FROM] 'satellites.csv'

.nullify &fbvtcb[satellites_ucs] 'NaN'

&fbvtcg[CREATE POPULATION] &fbvtcb[satellites] &fbvtcg[FOR] &fbvtcb[satellites_ucs]
  &fbvtcg[WITH SCHEMA] (
    &fbvtcg[IGNORE] Name;

    &fbvtcg[MODEL]
      country_of_operator, operator_owner,
      purpose, class_of_orbit, type_of_orbit
      users, contractor, launch_vehicle,
      country_of_contractor, launch_site,
      source_used_for_orbital_data
    &fbvtcg[AS NOMINAL];

    &fbvtcg[MODEL]
      perigee_km, apogee_km, eccentricity,
      period_minutes launch_mass_kg,
      dry_mass_kg, power_watts,
      date_of_launch, anticipated_lifetime
    &fbvtcg[AS NUMERICAL];

    &fbvtcg[MODEL]
        longitude_radians_of_geo,
        inclination_radians
    &fbvtcg[AS CYCLIC]
);

&fbvtcg[CREATE METAMODEL] &fbvtcb[sat_hybrid] &fbvtcg[FOR] &fbvtcb[satellites]
  &fbvtcg[WITH BASELINE] &fbvtcr[crosscat](
    &fbvtcg[SET CATEGORY MODEL FOR] eccentricity &fbvtcg[TO] beta;

    &fbvtcg[OVERRIDE GENERATIVE MODEL FOR]
      launch_mass_kg, dry_mass_kg, power_watts,
      perigee_km, apogee_km
    &fbvtcg[AND EXPOSE]
      pc1 &fbvtcg[NUMERICAL], pc2 &fbvtcg[NUMERICAL]
    &fbvtcg[USING] &fbvtcr[factor_analysis](L=2);

    &fbvtcg[OVERRIDE GENERATIVE MODEL FOR]
      period_minutes
    &fbvtcg[GIVEN]
      apogee_km, perigee_km
    &fbvtcg[AND EXPOSE]
      kepler_cluster &fbvtcg[CATEGORICAL],
      kepler_noise &fbvtcg[NUMERICAL]
    &fbvtcg[USING] &fbvtcr[venturescript](sp=kepler);

    &fbvtcg[OVERRIDE GENERATIVE MODEL FOR]
      anticipated_lifetime
    &fbvtcg[GIVEN]
      date_of_launch, power_watts, apogee_km,
      perigee_km, dry_mass_kg, class_of_orbit
    &fbvtcg[USING] &fbvtcr[linear_regression];

    &fbvtcg[OVERRIDE GENERATIVE MODEL FOR]
      type_of_orbit
    &fbvtcg[GIVEN]
      apogee_km, perigee_km, period_minutes,
      users, class_of_orbit
    &fbvtcg[USING] &fbvtcr[random_forest](k=7);
);
\end{Verbatim}
\end{subfigure}
\begin{subfigure}[t]{.5\textwidth}
\begin{Verbatim}[fontsize=\scriptsize, frame=single, label={\%venturescript}]
// Kepler CGPM.
define kepler = () -> {
  // Kepler's law.
  assume keplers_law = (apogee, perigee) -> {
    let GM = 398600.4418;
    let earth_radius = 6378;
    let a = (abs(apogee) + abs(perigee)) *
        0.5 + earth_radius;
    2 * 3.1415 * sqrt(a**3 / GM) / 60
  };
  // Internal samplers.
  assume crp_alpha = .5;
  assume cluster_sampler = make_crp(crp_alpha);
  assume error_sampler = mem((cluster) ->
        make_nig_normal(1, 1, 1, 1));
  // Output simulators.
  assume sim_cluster_id =
    mem((rowid, apogee, perigee) ~> {
      tag(atom(rowid), atom(1), cluster_sampler())
  });
  assume sim_error =
    mem((rowid, apogee, perigee) ~> {
      let cluster_id = sim_cluster_id(
        rowid, apogee, perigee);
      tag(atom(rowid), atom(2),
        error_sampler(cluster_id)())
  });
  assume sim_period =
    mem((rowid, apogee, perigee) ~> {
      keplers_law(apogee, perigee) +
        sim_error(rowid, apogee, perigee)
  });
  // List of simulators.
  assume simulators = [
    sim_period, sim_cluster_id, sim_error];
};

// Output observers.
define obs_cluster_id =
  (rowid, apogee, perigee, value, label) -> {
    $label: observe sim_cluster_id(
      $rowid, $apogee, $perigee) = atom(value);
};
define obs_error =
  (rowid, apogee, perigee, value, label) -> {
    $label: observe sim_error(
      $rowid, $apogee, $perigee) = value;
};
define obs_period =
  (rowid, apogee, perigee, value, label) -> {
    let theoretical_period = run(
      sample keplers_law($apogee, $perigee));
    obs_error(
      rowid, apogee, perigee,
      value - theoretical_period, label);
};
// List of observers.
define observers = [
  obs_period, obs_cluster_id, obs_error];
// List of inputs.
define inputs = ["apogee", "perigee"];
// Transition operator.
define transition = (N) -> {mh(default, one, N)};
\end{Verbatim}
\end{subfigure}%
\end{figure}

\begin{figure}[ht]
  \begin{subfigure}[t]{.5\textwidth}
    \centering
    \includegraphics[width=\textwidth]{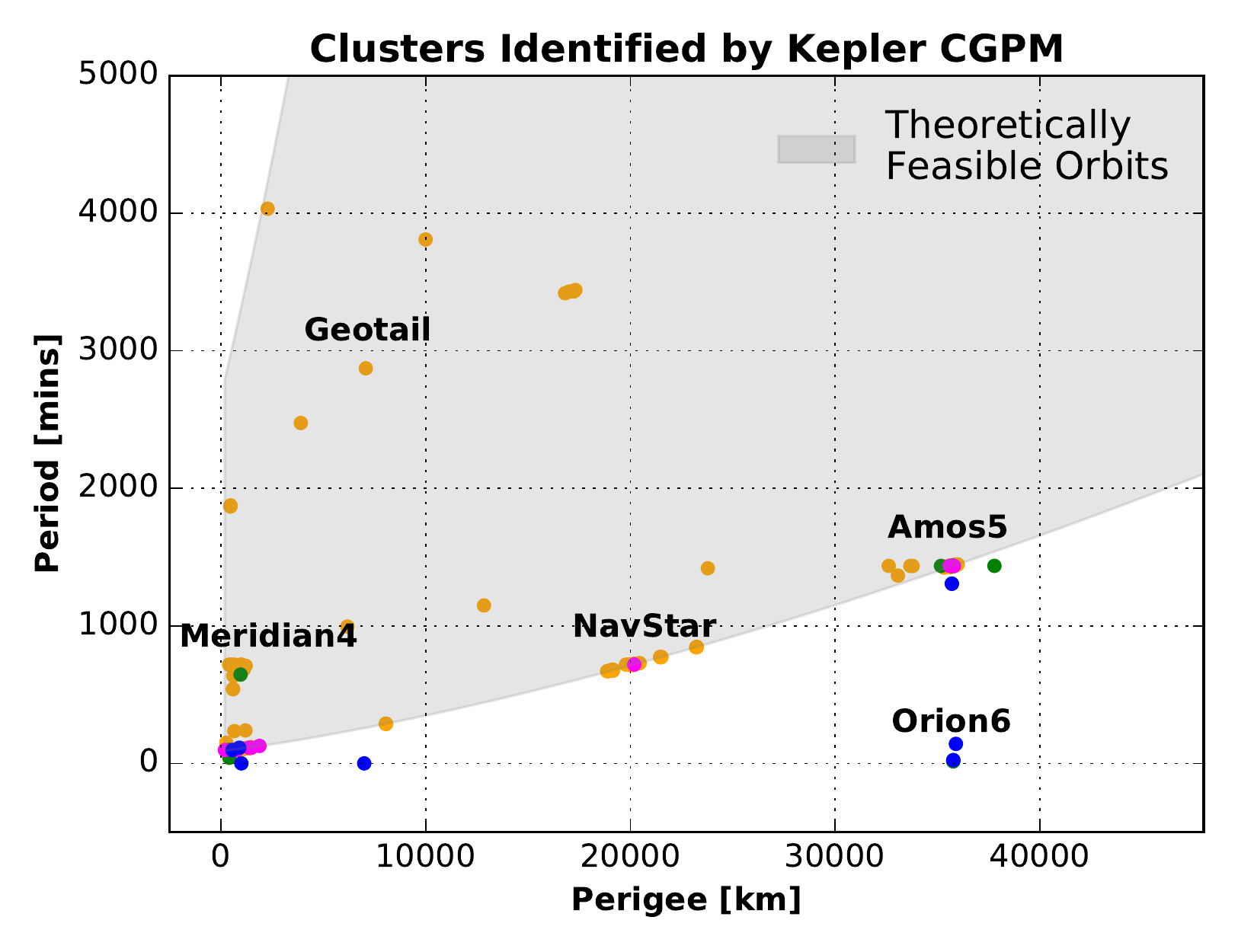}
  \end{subfigure}%
  \begin{subfigure}[t]{.5\textwidth}
    \includegraphics[width=\textwidth]{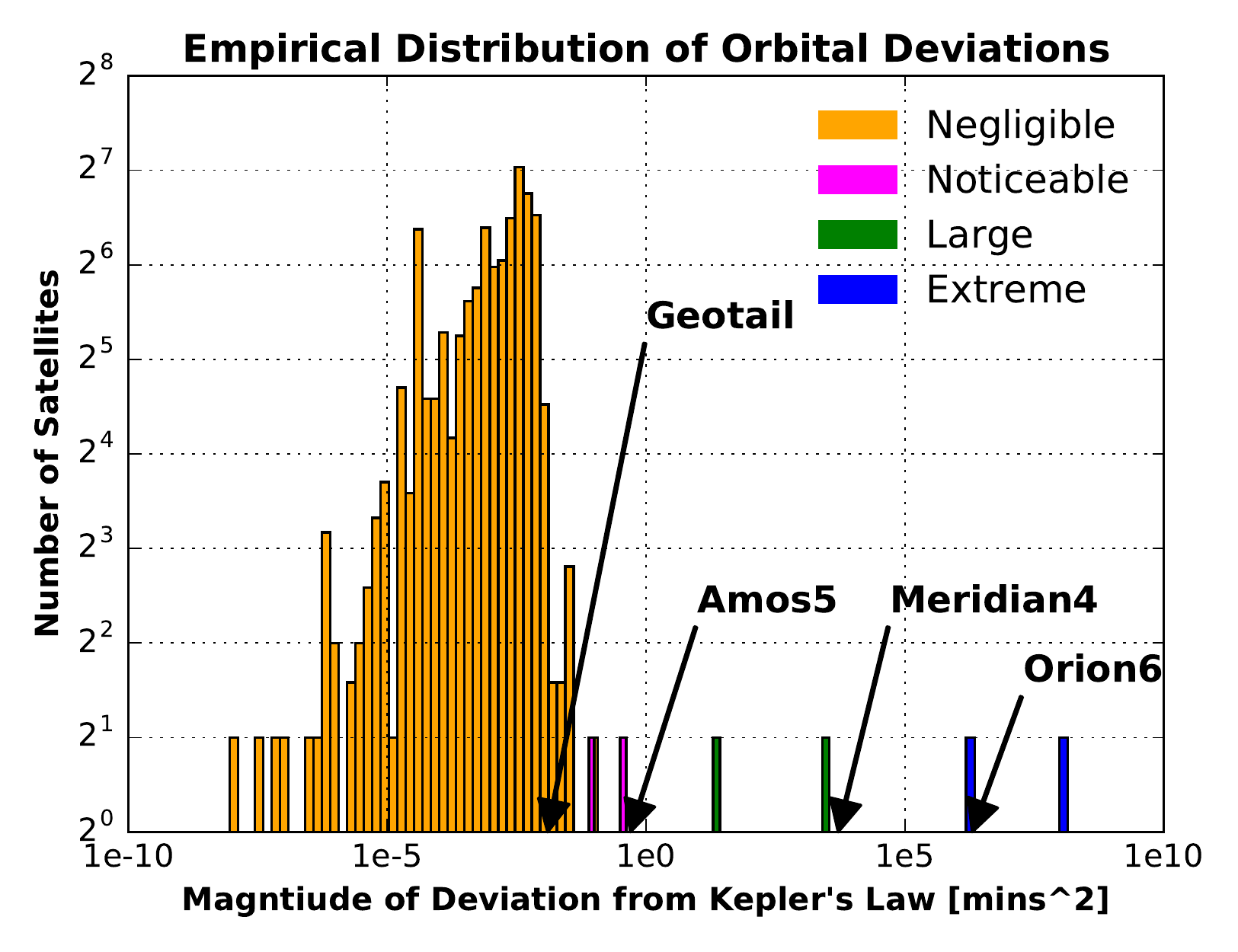}
  \end{subfigure}\newline
    \texttt{%
        \%bql INFER kepler\_cluster, kepler\_noise FROM satellites;}
    \bcaption{%
        Finding satellites whose orbits are likely violations of Kepler's Third Law using a causal CGPM in Venturescript, which learns a Dirichlet process mixture on the residuals.}{%
        Each dot in the scatter plot (left) is a satellite in the dataset, and its color represents the latent cluster assignment learned by the causal CGPM.
        Both the cluster identity and inferred noise are exposed latent variables.
        The histogram (right) shows that each of the four distinct clusters roughly translates to a qualitative description for the magnitude of a satellite's deviation from its theoretical period: yellow (negligible), magenta (noticeable), green (large), and blue (extreme).
        These clusters were learned non-parametrically.}
        \label{fig:kepler-overall}
\end{figure}

\clearpage

\begin{figure}
    \begin{subfigure}{\textwidth}
        \centering
        \includegraphics[width=.5\textwidth]{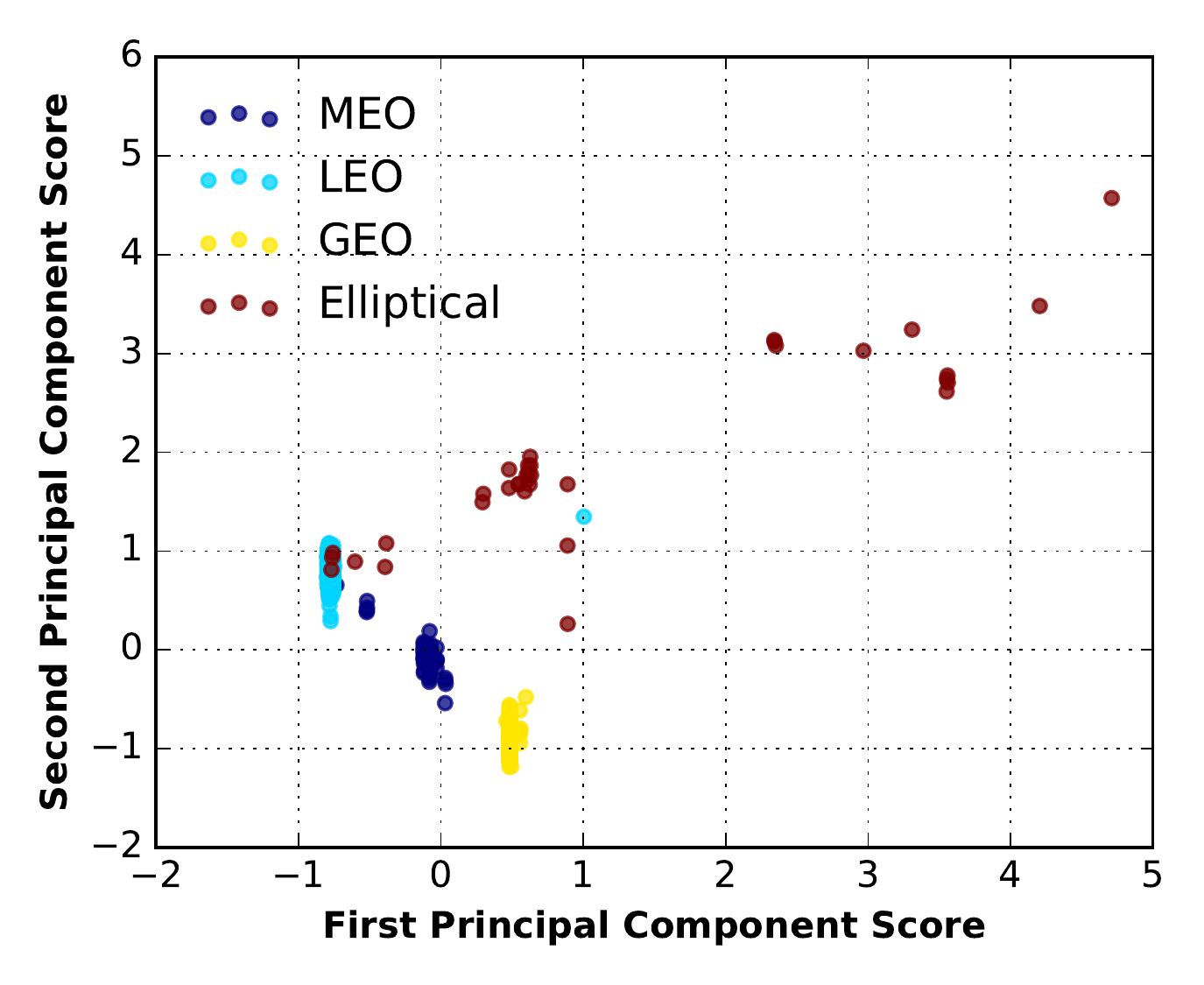}
    \end{subfigure}\newline
    \texttt{\small%
        \%bql INFER EXPLICIT PREDICT pc1, PREDICT pc2,
            class\_of\_orbit FROM satellites;}
    \bcaption{%
        Low dimensional projection of the satellites using the PCA CGPM
        reveals clusterings in latent space and suggests candidate
        outliers.}{%
        The principal component scores are based on the numerical features of a satellite, and the color is the \texttt{class\_of\_orbit}.
        Satellites in low earth, medium earth, and geosynchronous orbit form tight clusters in latent space along PC1, and exhibit most within-cluster variance along PC2.
        The distribution on factor scores for elliptical satellites has much higher variability along both dimensions, indicating a collection of weak local modes depending on the regime of the satellite's \texttt{eccentricity} (not shown), and/or many statistical outliers.}
    \label{fig:pca-satellites}
\end{figure}

\clearpage

\begin{figure}
  \begin{subfigure}[b]{.5\textwidth}
    \includegraphics[width=\textwidth]{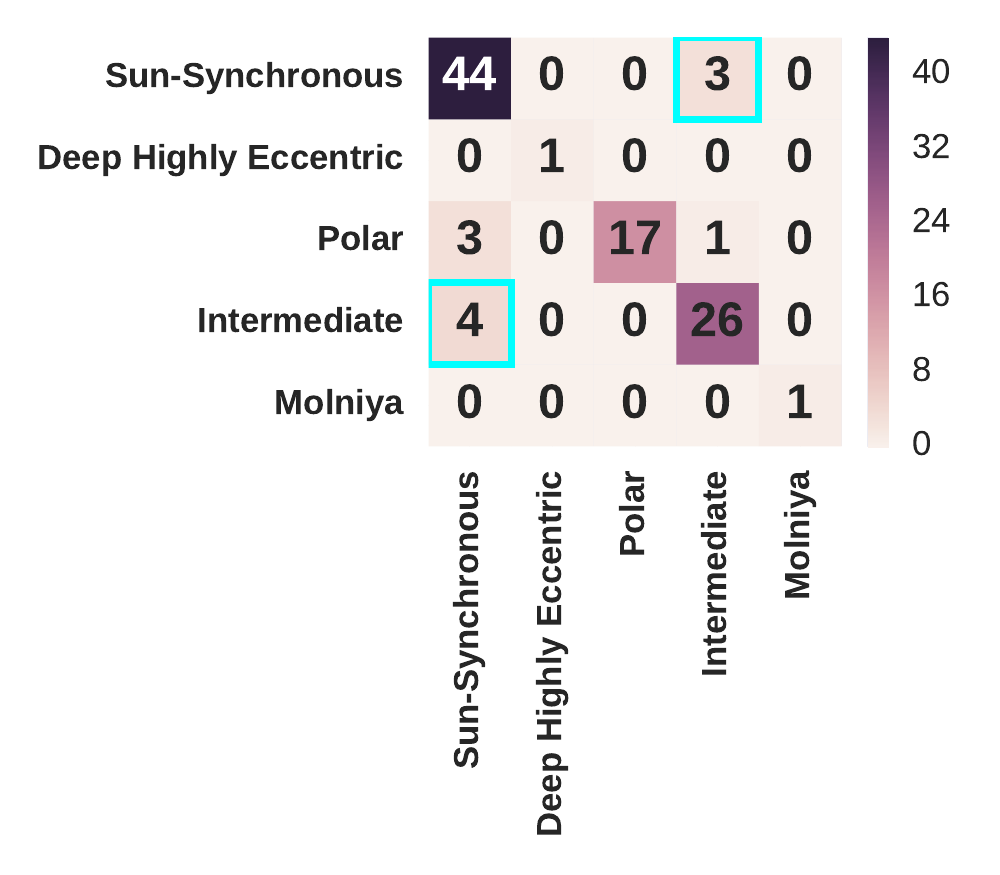}
    \captionsetup{margin=10pt}
    \caption{Crosscat/Random Forest hybrid CGPM.}
    \label{fig:confusion-rf}
  \end{subfigure}%
  \begin{subfigure}[b]{.5\textwidth}
    \includegraphics[width=\textwidth]{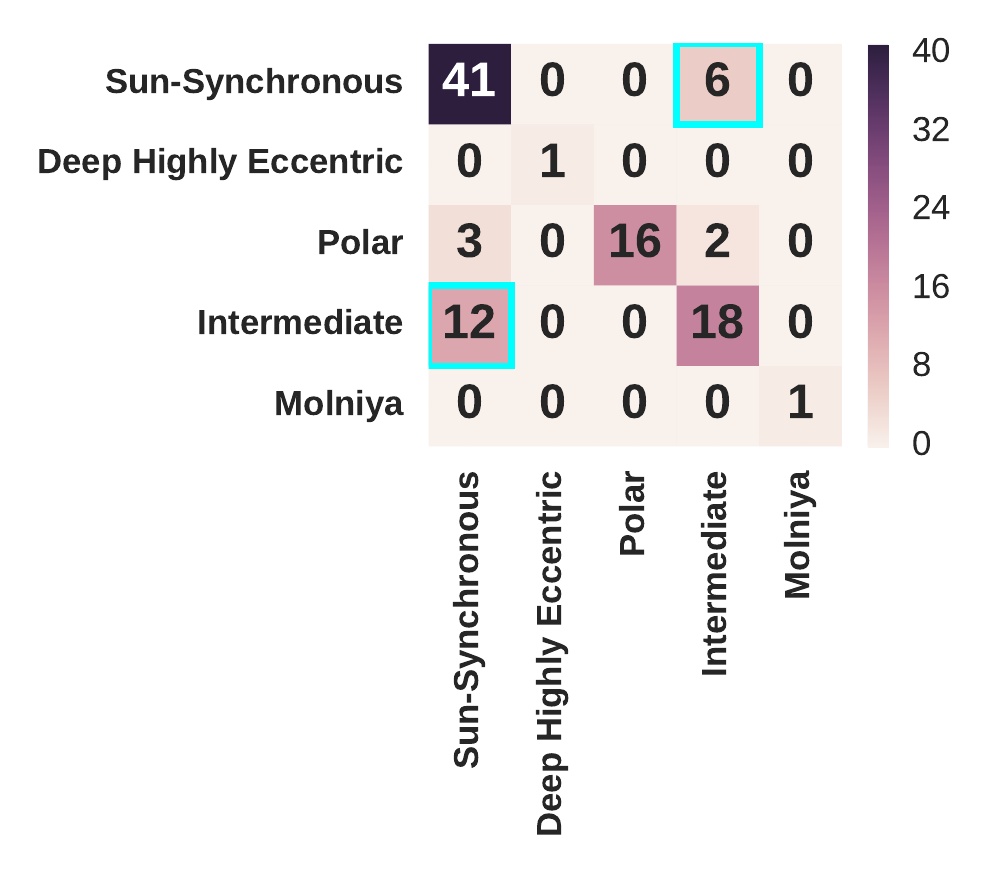}
    \captionsetup{margin=10pt}
    \caption{CrossCat baseline CGPM.}
    \label{fig:confusion-cc}
  \end{subfigure}\newline\newline
    \texttt{%
        \%bql INFER type\_of\_orbit FROM held\_out\_satellites;}
    \bcaption{%
        Confusion matrices for a multiclass classification task show improved prediction accuracy by the hybrid CGPM over the CrossCat baseline.}{%
        The y-axis shows the true label for ``type of orbit'' of 100 held-out satellites, and the x-axis shows the predicted label by each CGPM.
        The feature vectors are five dimensional and consist of numerical and categorical variables (lines 57-62 of Figure~\ref{fig:mml-hybrid}), and both test and training sets contained missing data.
        While both CrossCat and Crosscat + Random Forest systematically confuse ``sun-synchronous''and ``intermediate'' orbits (entries in cyan), the overall error rate is reduced by 11\% in the hybrid CGPM.}
    \label{fig:confusion}
\end{figure}

\clearpage

\begin{figure}
    \includegraphics[width=\textwidth]{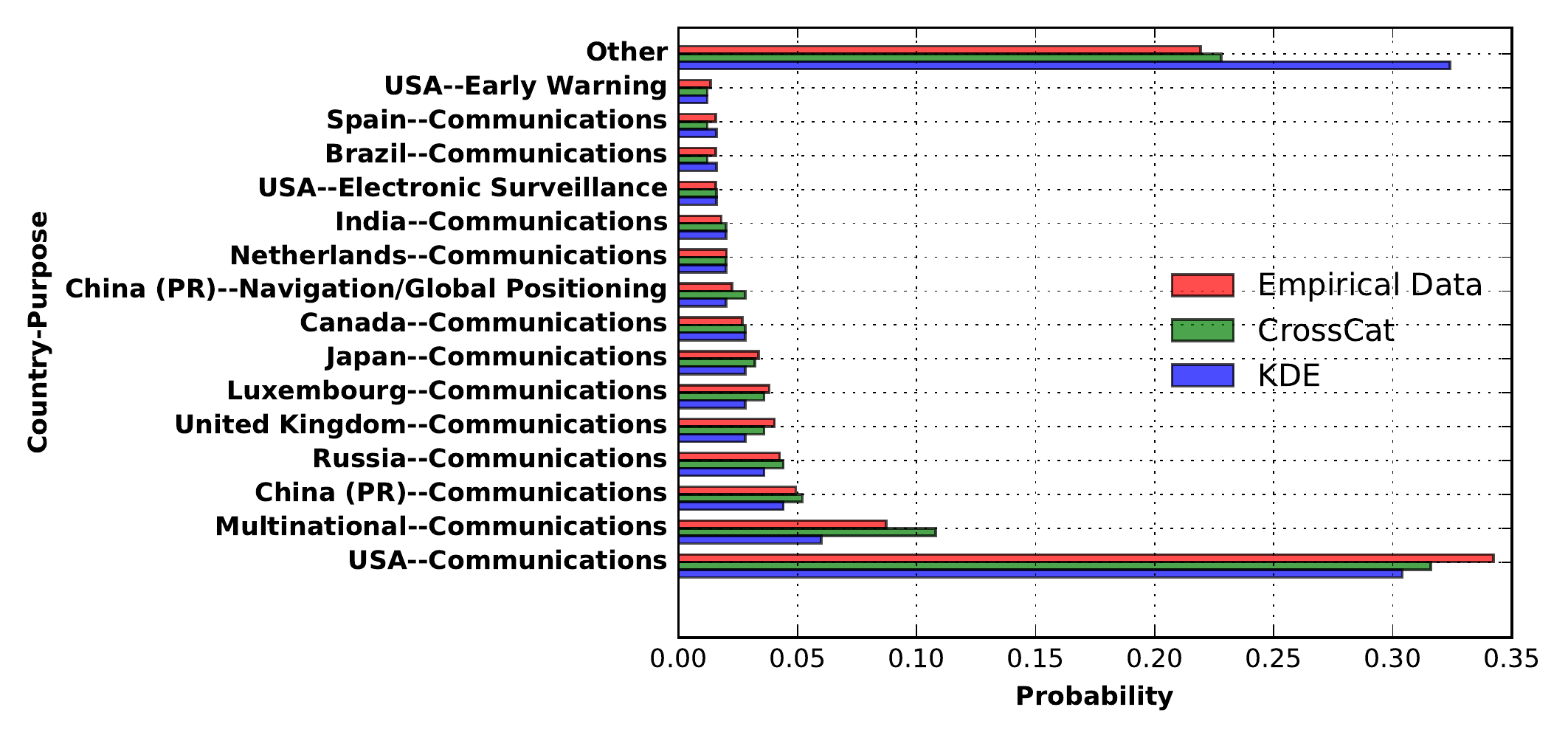}
    \texttt{\small%
        \%bql SIMULATE country\_of\_operator, purpose GIVEN class\_of\_orbit = `GEO';}
        \bcaption{%
        Simulating from the joint distribution of the country and purpose of a hypothetical satellite, given its orbit type.
        }{%
        The y-axis shows the simulated country-purpose pairs, and the x-axis shows the frequency of simulations, compared to the true frequency in the dataset.
        500 samples were obtained from CrossCat and multivariate KDE to estimate the posterior probabilities.
        The posteriors of both CrossCat and KDE are smooth versions of the empirical data -- the smoothing for CrossCat is induced by the inner Dirichlet process mixture over category models, and for KDE is induced by the bandwidth parameters of the Aitchison and Aitken kernels.
        The plot shows that CrossCat's samples provide a tighter fit to the dataset.
        The distribution from KDE has a fatter tail, as indicated by the high number of samples classified in the ``Other'' category.
        }
    \label{fig:cc-vs-kde}
\end{figure}

\clearpage


\begin{figure}[ht]
    \begin{subfigure}{.5\textwidth}
        \begin{Verbatim}[gobble=8, fontsize=\scriptsize, frame=single, numbers=left, commandchars=&\[\]]
        &fbvtcg[CREATE TABLE] &fbvtcb[data_train] &fbvtcg[FROM] satellites_train.csv;
        .nullify &fbvtcb[data_train] 'NaN';

        &fbvtcg[CREATE POPULATION] &fbvtcb[satellites] &fbvtcg[FOR] &fbvtcb[data_train]
          &fbvtcg[WITH SCHEMA](
            &fbvtcg[GUESS STATTYPES FOR] (*)
        );

        &fbvtcg[CREATE METAMODEL] &fbvtcb[cc_ols] &fbvtcg[FOR] &fbvtcb[satellites]
          &fbvtcg[WITH BASELINE] &fbvtcr[crosscat](
            &fbvtcg[OVERRIDE GENERATIVE MODEL FOR]
                anticipated_lifetime
            &fbvtcg[GIVEN]
                type_of_orbit, perigee_km, apogee_km,
                period_minutes, date_of_launch,
                launch_mass_kg
            &fbvtcg[USING] &fbvtcr[linear_regression]
        );

        &fbvtcg[INITIALIZE] 4 &fbvtcg[MODELS FOR] &fbvtcb[cc_ols];
        &fbvtcg[ANALYZE] &fbvtcb[cc_ols] &fbvtcg[FOR] 100 &fbvtcg[ITERATION WAIT];

        &fbvtcg[CREATE TABLE] &fbvtcb[data_test] &fbvtcg[FROM] satellites_test.csv;
        .nullify &fbvtcb[data_test] 'NaN';
        .sql &fbvtcg[INSERT INTO] &fbvtcb[data_train]
            &fbvtcg[SELECT] * &fbvtcg[FROM] &fbvtcb[data_test];

        &fbvtcg[CREATE TABLE] predicted_lifetime &fbvtcg[AS]
            &fbvtcg[INFER EXPLICIT]
                &fbvtcg[PREDICT] anticipated_lifetime
                &fbvtcg[CONFIDENCE] pred_conf
            &fbvtcg[FROM] &fbvtcb[satellites] &fbvtcg[WHERE] _rowid_ > 1000;
        \end{Verbatim}
    \captionsetup{skip=-10pt, margin=10pt}
    \caption{Full session in BayesDB which loads the training and test sets, creates a hybrid CGPM, and runs the regression.}
    \label{fig:loc-regression-cgpm}
    \end{subfigure}%
    \begin{subfigure}{.5\textwidth}
        \begin{Verbatim}[gobble=8, fontsize=\scriptsize, frame=single]
        def dummy_code_categoricals(frame, maximum=10):

            def dummy_code_categoricals(series):
                categories = pd.get_dummies(
                    series, dummy_na=1)
                if len(categories.columns) > maximum - 1:
                    return None
                if sum(categories[np.nan]) == 0:
                    del categories[np.nan]
                categories.drop(
                    categories.columns[-1], axis=1,
                    inplace=1)
                return categories

        def append_frames(base, right):
            for col in right.columns:
                base[col] = pd.DataFrame(right[col])

        numerical = frame.select_dtypes(include=[float])
        categorical = frame.select_dtypes(
            include=['object'])
        categorical_coded = filter(
            lambda s: s is not None,
            [dummy_code_categoricals(categorical[c])
                for c in categorical.columns])

        joined = numerical

        for sub_frame in categorical_coded:
            append_frames(joined, sub_frame)

        return joined
        \end{Verbatim}
    \captionsetup{skip=-10pt, margin=10pt}
    \caption{Ad-hoc Python routine (used by baselines) for dummy coding nominal predictors in a dataframe with missing values and heterogeneous types.}
    \label{fig:loc-regression-baseline}
    \end{subfigure}
    \begin{subfigure}{\textwidth}
        \center
        \includegraphics[width=\textwidth]{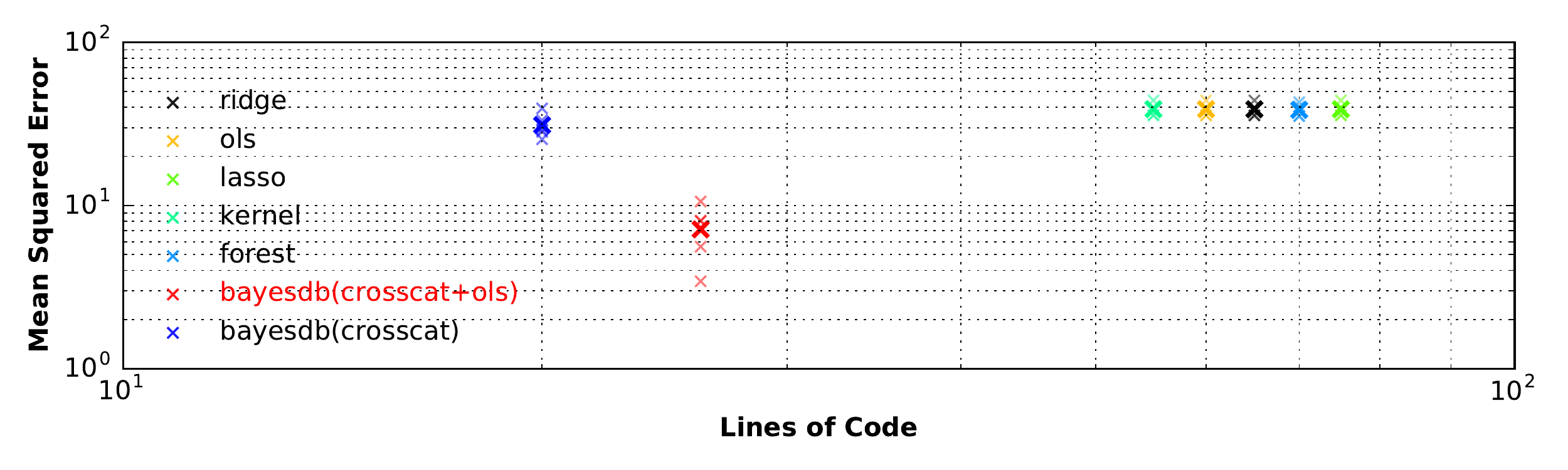}
    \end{subfigure}
    \captionsetup{skip=0pt}
    \bcaption{%
        In a high-dimensional regression problem with mixed data types and missing data, the composite CGPM shows improvement in prediction accuracy over purely generative and purely discriminative baselines.}{%
        The task is to infer the anticipated lifetime of a held-out satellite given categorical and numerical features such as type of orbit, launch mass, and orbital period.
        Some feature vectors in the test set have missing entries, leading purely discriminative models (ridge, lasso, OLS) to either heuristically impute missing features, or to ignore the features and predict the mean lifetime from its marginal distribution in the training set.
        The purely generative model (CrossCat) is able to impute missing data from their full joint distribution, but only indirectly mediates dependencies between the predictors and response through latent variables.
        The composite CGPM (CrossCat+OLS) combines advantages of both approaches; statistically rigorous imputation followed by direct regression on the features leads to improved predictive accuracy.}
    \label{fig:loc-regression}
\end{figure}

\clearpage

\begin{figure}[ht]
    \begin{subfigure}{.45\textwidth}
        \includegraphics[width=\textwidth]{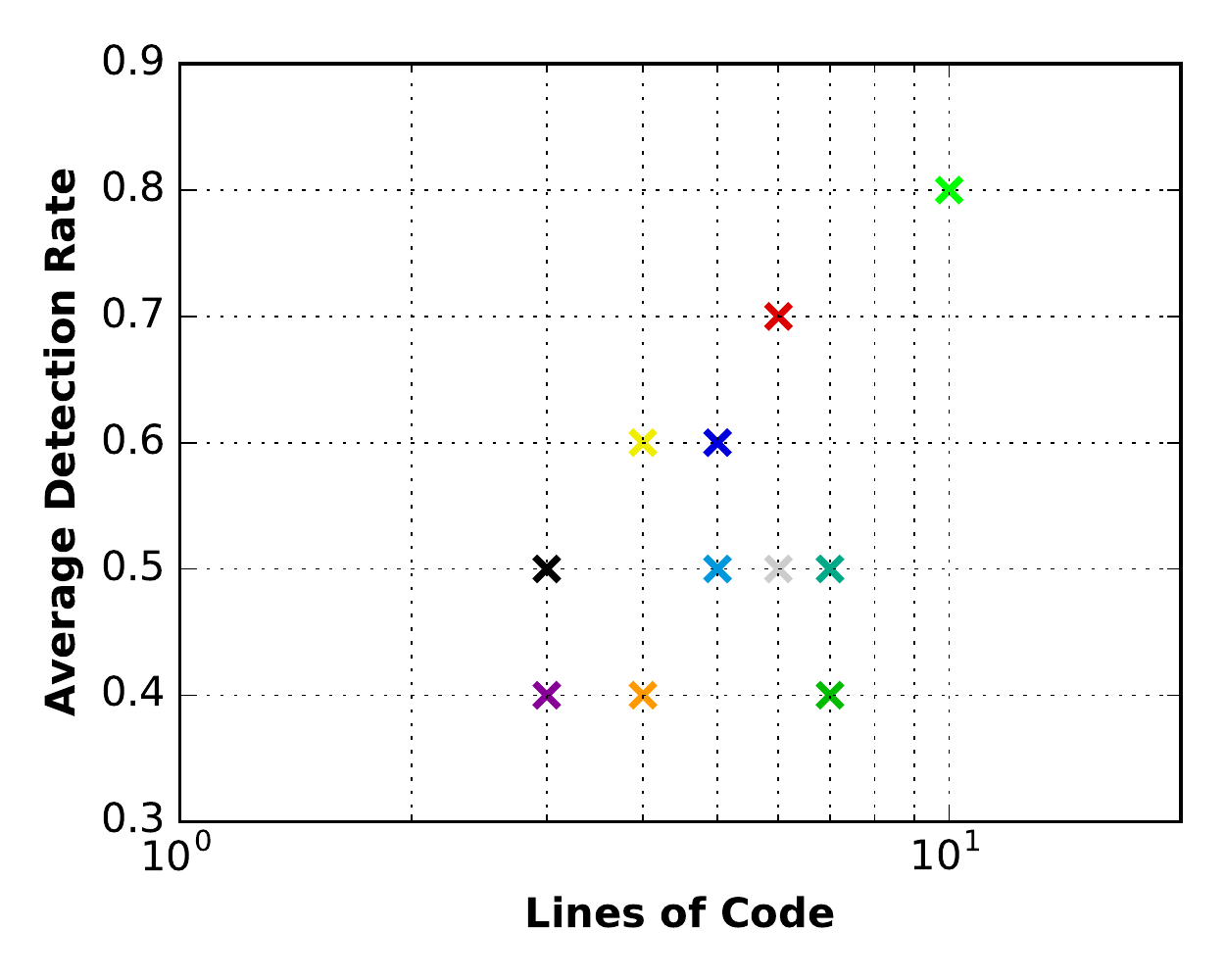}
    \end{subfigure}
    \begin{subfigure}{.5\textwidth}
        \includegraphics[width=\textwidth]{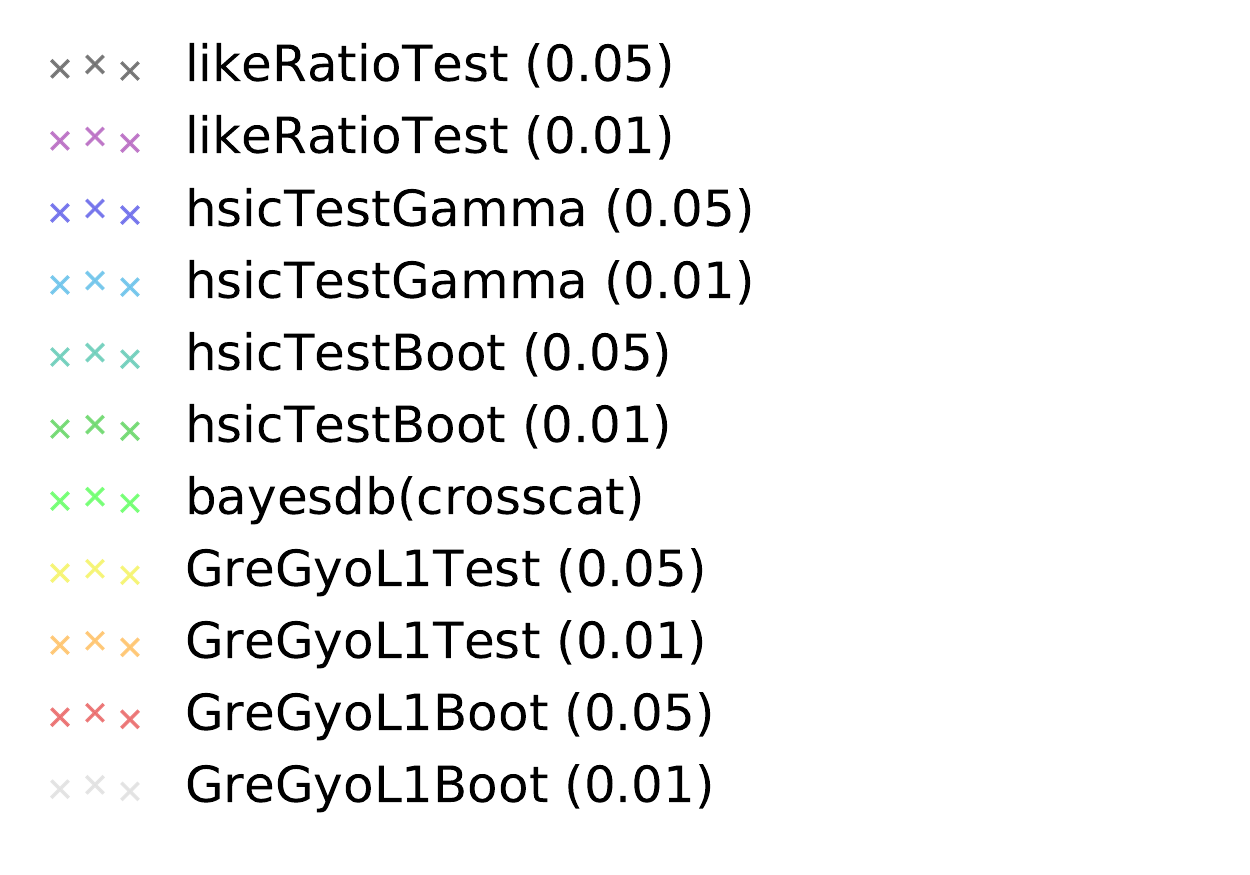}
    \end{subfigure}
    \texttt{\small%
        \%bql ESTIMATE DEPENDENCE PROBABILITY OF x WITH y;}
    \bcaption{%
        Dependence discovery.}{%
        Binary hypothesis tests of independence for synthetic two-dimensional data drawn from five noisy zero-correlation datasets: sin wave, parabola, x-cross, diamond, and ring.
        For all datasets the two dimensions are dependent.
        The y-axis shows the fraction of correct hypotheses achieved by each method, averaged over all datasets.
        The decision rule for kernel-based tests \citep{gretton2007,gretton2008,gretton2010}, is based on a frequentist significance level of 5\% and 1\%.
        The decision rule for CrossCat is based on a dependence probability threshold of 50\%.}
        \label{fig:independence-testing}
\end{figure}

\begin{figure}[ht]
    \begin{subfigure}{.45\textwidth}
        \includegraphics[width=\textwidth]{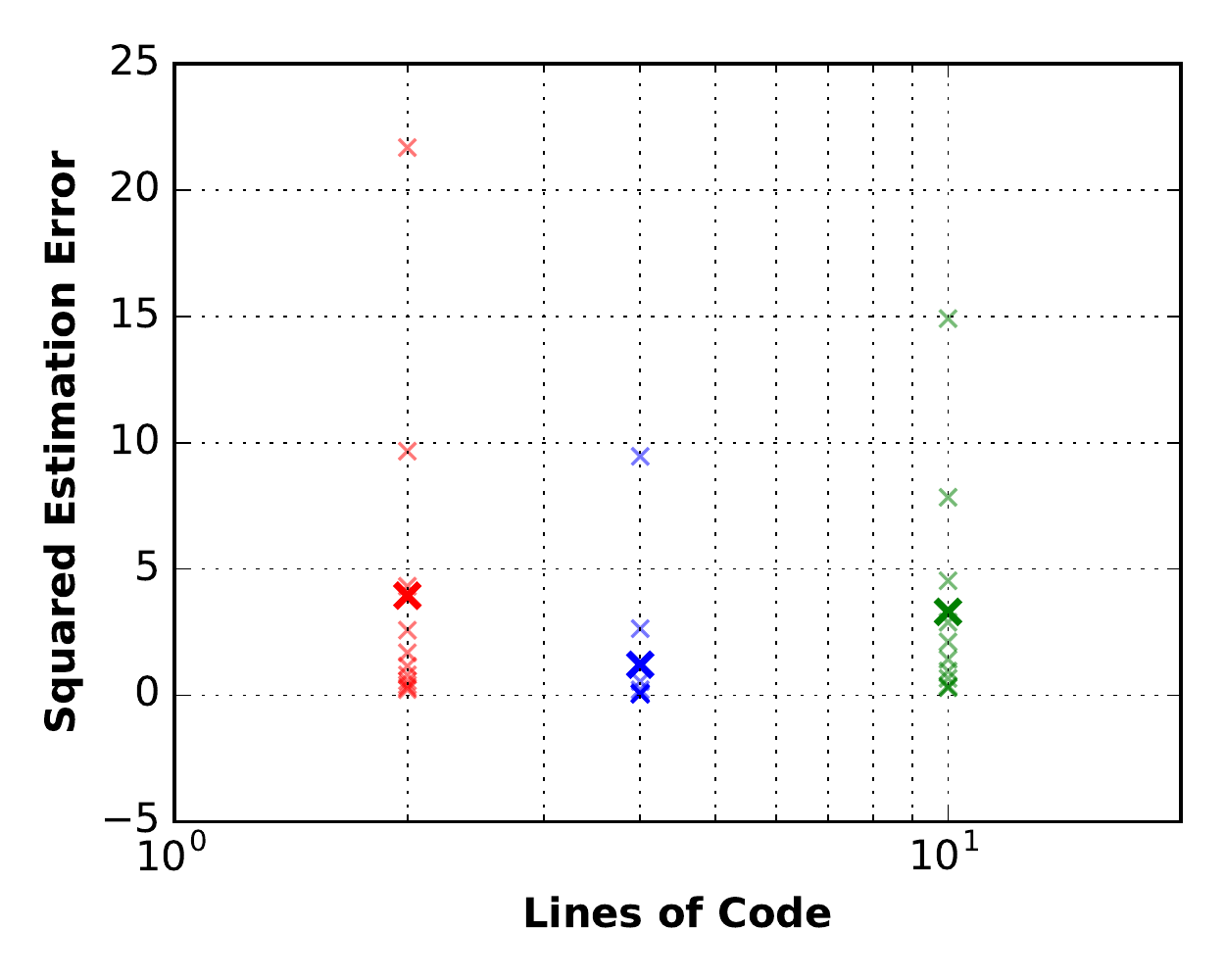}
    \end{subfigure}
    \begin{subfigure}{.5\textwidth}
        \includegraphics[width=\textwidth]{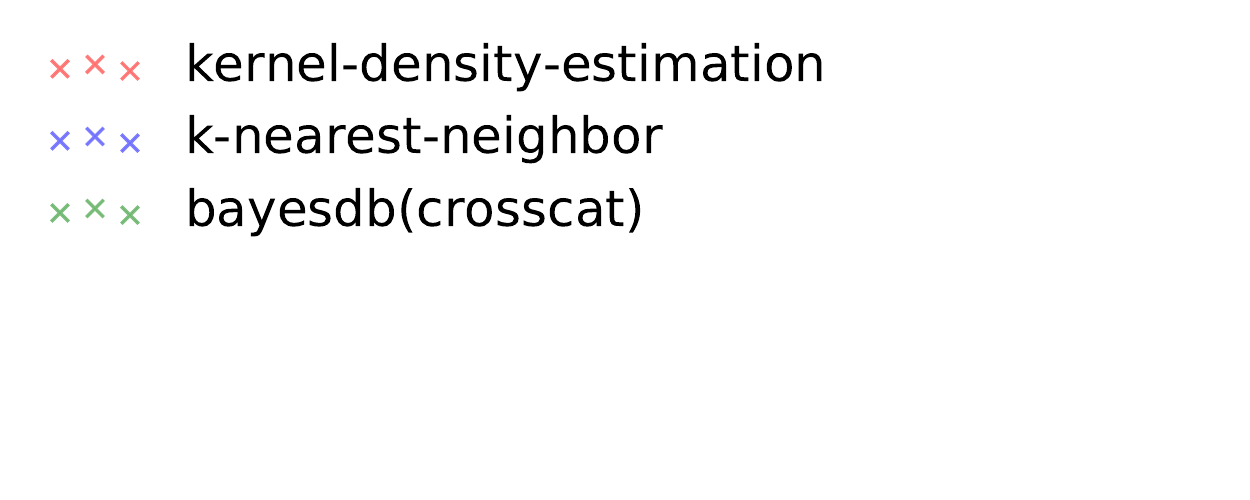}
    \end{subfigure}
    \texttt{\small%
        \%bql ESTIMATE MUTUAL INFORMATION OF x WITH y;}
    \bcaption{%
        Dependence strength}{%
        Estimating the mutual information of a noisy sin wave.
        The y-axis shows the squared estimation error, randomized over observed datasets.
        The ``ground truth'' mutual information was derived analytically, and the integral computed by quadrature.
        Baseline methods estimate mutual information using K nearest neighbors \citep{kraskov2004} and kernel density estimation \citep{moon1995}.
        CrossCat estimates the mutual information first by learning a Dirichlet process mixture of Gaussians, and using Monte Carlo estimation by generating samples from the posterior predictive distribution and assessing their density.}
    \label{fig:dependence-strength}
\end{figure}

\clearpage

\begin{figure}[ht]
    \begin{subfigure}{.45\textwidth}
        \includegraphics[width=\textwidth]{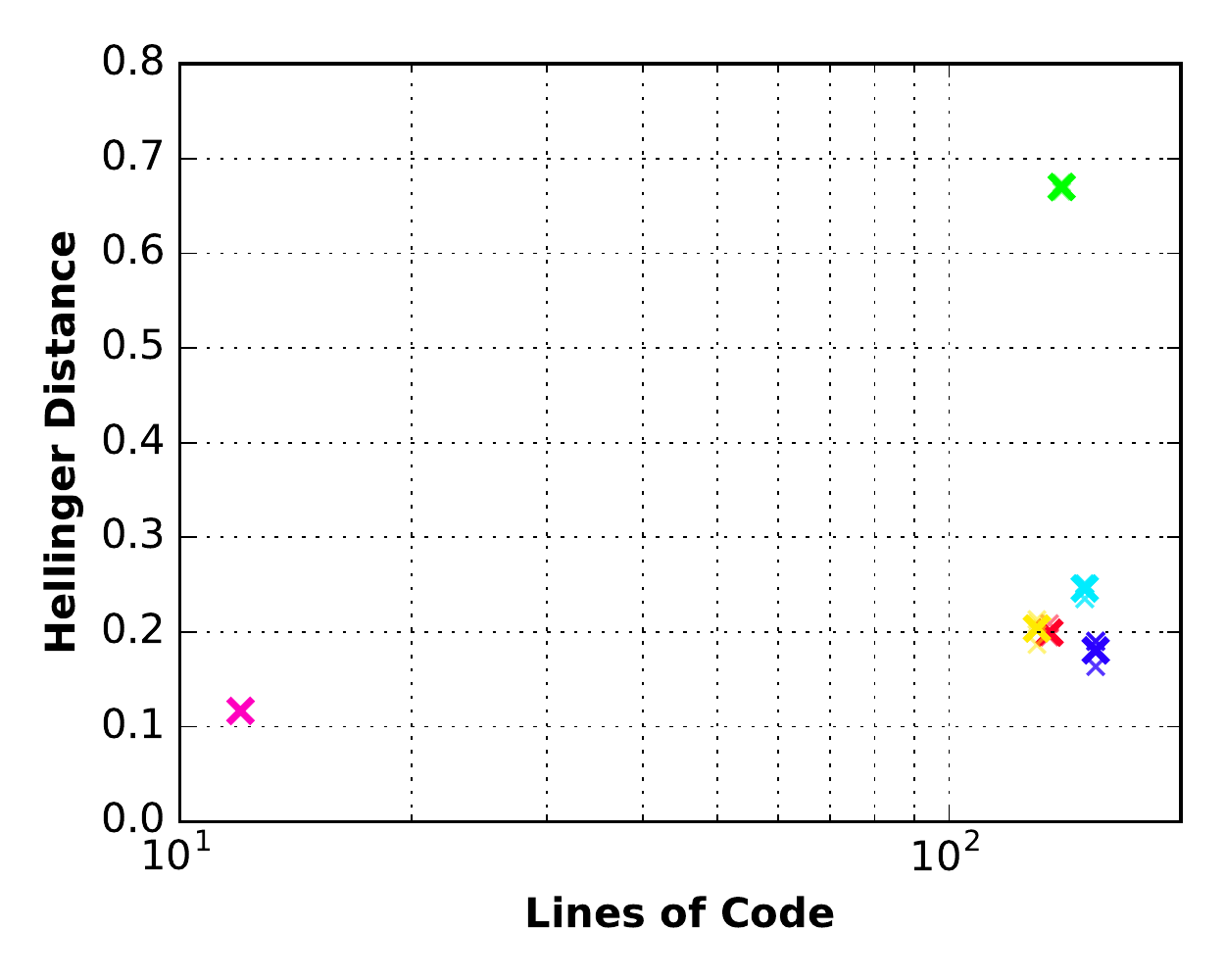}
    \end{subfigure}
    \begin{subfigure}{.5\textwidth}
        \includegraphics[width=\textwidth]{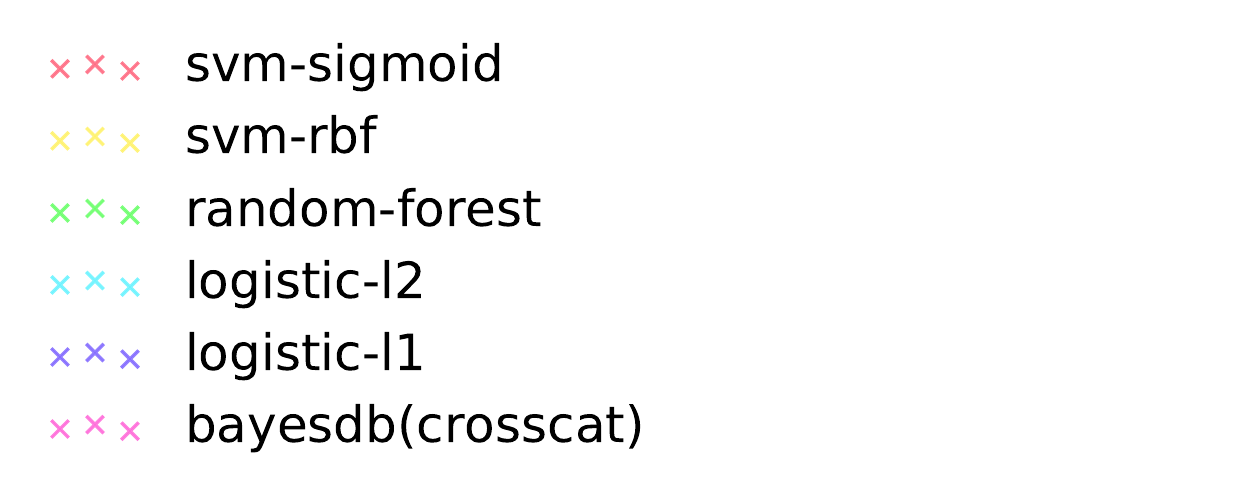}
    \end{subfigure}
    \texttt{\small%
        \%bql SIMULATE country\_of\_operator, purpose GIVEN class\_of\_orbit = `GEO';}
        \bcaption{%
        Bivariate categorical density estimation.}{%
        Simulating from the posterior joint distribution of the country and purpose of a hypothetical satellite, given its orbit type.
        500 samples were obtained from each method to estimate the posterior probabilities.
        The y-axis shows the Hellinger distance between posterior samples from each method and the empirical conditional distribution from the dataset, used as ``ground truth''.
        Standard discriminative baselines struggle to learn the distribution of a two-dimensional discrete outcome based on a discrete input, where both the predictor and response variables take values in large categorical sets.}
        \label{fig:bivariate-classification}
\end{figure}

\begin{figure}[ht]
    \begin{subfigure}{.45\textwidth}
        \includegraphics[width=\textwidth]{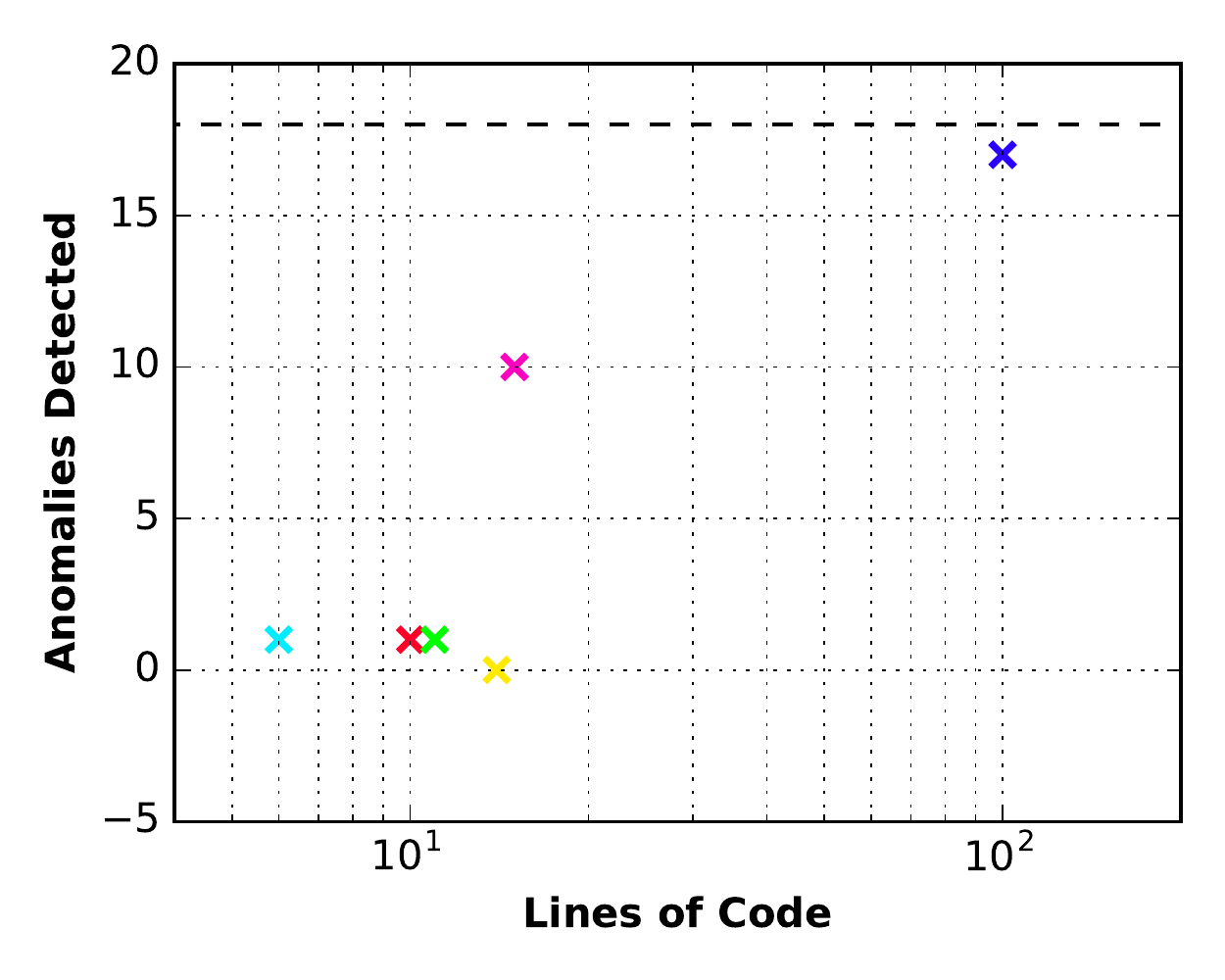}
    \end{subfigure}
    \begin{subfigure}{.5\textwidth}
        \includegraphics[width=\textwidth]{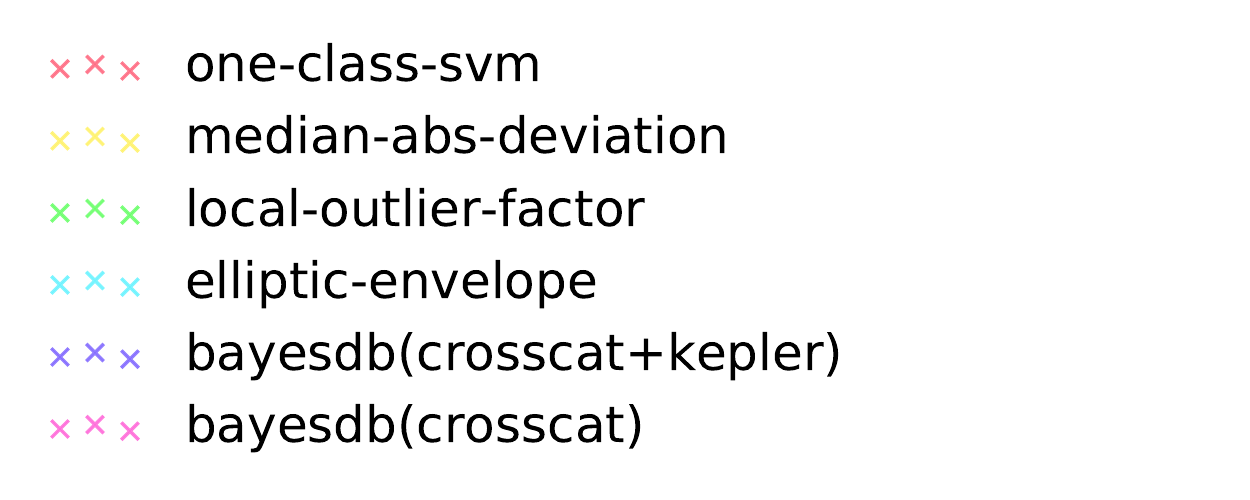}
    \end{subfigure}
    \texttt{\small%
        \%bql ESTIMATE PREDICTIVE PROBABILITY OF period\_minutes;}
    \bcaption{%
        Anomaly detection.}{%
        Detecting satellites with anomalous orbital periods.
        18 satellites from the dataset demonstrated a non-trivial deviation (greater than five minutes) from their theoretical period, used as ``ground truth'' anomalies.
        For each method, the top 20 satellites ranked by ``outlyingness'' score were used as the predicted anomalies.
        Hybrid CGPMs learn multivariate and multimodal distributions over all variables in the dataset, leading to higher detection rates than baseline methods which use univariate and/or unimodal statistics.
        The Kepler CGPM identifies most anomalies at the expense of a highly complex program in comparison to baselines.}
        \label{fig:anomaly-detection}
\end{figure}


\section{Discussion and Future Work}
\label{sec:discussion}

This paper has shown that it is possible to use a computational formalism in probabilistic programming to apply, combine, and compare a broad class of probabilistic data analysis techniques.
CGPMs extend the core provided by directed graphical models, which express elaborate probabilistic models in terms of smaller univariate pieces, by specifying a computational interface that allows these pieces to be multivariate, more black-box, and defined directly as software.
A key feature of this framework is that it enables statistical modelers to compose discriminative, generative and hybrid models from different philosophies in machine learning and statistics using probabilistic programming.
Moreover, the compositional abstraction is neutral to a CGPM's internal choices of (i) modeling assumptions, which may be i.e. hierarchical or flat, or Bayesian or non-Bayesian, and (ii) inference tactics, which may be i.e. optimization- or sampling-based.

Several models from statistics admit natural implementations in terms of the current CGPM interface, such as non-linear mixed effect models \citep{davidian1995}, where each member represents a potentially repeated measurement with latent variables grouping the members into observation units; or Gaussian processes \citep{rasmussen2006}, where the input variables are time indexes from another CGPM, and the outputs are noisy observations of the (latent) function values \citep{tresp2001, rasmussen2002}.
Computational representations of these models as CGPMs allows them to be composable as hybrid models, reusable as software, and queryable in interesting ways using the Bayesian Query Language.

Both \texttt{simulate} and \texttt{logpdf} in Listing~\ref{lst:cgpm-interface} are executed against a single member of the population i.e. variables within a single row.
Queries that target multiple members in the population are currently supported by an explicit sequence of \texttt{incorporate}, \texttt{infer}, and then \texttt{simulate} or \texttt{logpdf}.
It is interesting to consider extending the CGPM interface to natively handle arbitrary multi-row cases -- this idea was originally presented in the GPM interface \cite[Section 3.1.1]{mansinghka2015} although concrete algorithms for implementing multi-row queries, or surface-level syntax in the Bayesian Query Language for invoking them, were left as open questions.
Rather than support multi-row queries directly in the CGPM interface, it is instead possible to extend the BQL interpreter with a probabilistic query planner.
Given given a cross-row query, the BQL interpreter automatically determines a candidate set of invocation sequences of the CGPM interface to answer it, and then selects among them based on time/accuracy requirements.

A worthy direction for future work is extending the set of statistical data types (Section~\ref{subsec:implement-primitive}), and possibly CGPM interface, to support analysis tasks beyond traditional multivariate statistics.
Some possible new data types and associated CGPMs are
\begin{itemize}
\item \texttt{GRAPH} data type, using a relational data CGPM based on the stochastic block model \citep{nowicki2001} or infinite relational model \citep{kemp2006},

\item \texttt{TEXT} data type, using a topic model CGPM such as latent Dirichlet allocation \citep{blei2003} or probabilistic latent semantic analysis \citep{hofmann1999},

\item \texttt{IMAGE} data type, using a CGPM based on neural networks.
\end{itemize}
Composing CGPMs with these data types leads to interesting tasks over their induced joint distributions.
Consider an \texttt{IMAGE} variable with an associated \texttt{TEXT} annotation; a generative CGPM for the image and discriminative CGPM for the text (given the image) leads to image classification; a generative CGPM for the text and a discriminative CGPM for the image (given the text) allows simulating unstructured text followed by their associated images.

It is also interesting to consider introducing additional structure to our current formalism of populations from Section~\ref{subsec:cgpms-populations} to support richer notions of population modeling.
For instance, populations may be hierarchical in that the variables of population A correspond to outputs produced by a CGPM for population B -- the simplest case being summary statistics such as means, medians, and inter-quartile ranges.
Such hierarchical populations are common in census data, which contain raw measurements of variables for individual households, as well as row-wise and column-wise summaries based on geography, income level, ethnicity, educational background, and so on.
Populations can also be extended to support ``merge'' operations in MML, which are analogous to the \texttt{JOIN} operations in SQL, where the CGPM on the joined population allows for transfer learning.

Our presentation of the algorithm for \texttt{infer} in a composite network of CGPMs (Section~\ref{subsec:composition}) left open improvements to the baseline strategy of learning each CGPM node separately.
One way to achieve joint learning, without violating the abstraction boundaries of the CGPM interface, is: after running \texttt{infer} individually for each CGPM, run a ``refine'' phase, where (i) missing measurements in the population are imputed using one forward pass of \texttt{simulate} throughout the network, then (ii) each CGPM updates its parameters based on the imputed measurements.
This strategy can be repeated to generate several such imputed networks, which are then organized into an ensemble of CGPMs in a BayesDB metamodel (Section~\ref{subsubsec:bayesdb-mml-homogenous}) where each CGPM in the metamodel corresponds to a different set of imputations.
The weighted-averaging of these CGPMs by BayesDB would thus correspond to integration over different imputations, as well as their induced parameters.

Extending BQL, or developing new probabilistic programming languages, to assess the inference quality of CGPMs built in MML will be an important step toward broader application of these probabilistic programming tools for real-world analysis tasks.
For instance, it is possible to develop a command in BQL such as
\begin{center}
\small\tt ESTIMATE KL DIVERGENCE BETWEEN <cgpm-1> AND <cgpm-2>\\
FOR VARIABLES <var-names-a> GIVEN <var-names-b>;
\end{center}
which takes two CGPMs (and an overlapping subset of their output variables) and returns an estimate of the KL divergence between their conditional predictive distributions, based on a Monte Carlo estimator using \texttt{simulate} and \texttt{logpdf}.
Such model-independent estimators of inference quality, backed by the CGPM interface, provide a proposal for unifying the testing and profiling infrastructure among a range of candidate solutions for a given data analysis task.

This paper has shown that it is possible to unify and formalize a broad class of probabilistic data analysis techniques by integrating them into a probabilistic programming platform, which is itself integrated with a traditional database.
We have focused on a class of probabilistic models that can be tightly integrated with flat database tables.
Population schemas define the variables of interest along with their types, but unlike traditional database schemas, they can additionally include variables whose values are never directly observed.
Concrete probabilistic models for populations are built via automated inference mechanisms, according to a baseline meta-modeling strategy which can also be customized.
This idea is similar to concrete indexes for tables in traditional databases which are built by automated mechanisms, according to an indexing strategy which can be customized via its own schema.
While we are encouraged by the early successes of this approach, there is a vast literature of richer ``data modeling'' formalisms from both databases and statistics.
Integrating these ideas could yield further conceptual insight and practical benefits.
We hope this paper encourages others to develop these connections, along with a new generation of intelligent tools for machine-assisted probabilistic data analysis.

\bibliography{thesis}

\end{document}